%% file: example_paper.tex
\pdfoutput=1

\documentclass{article}

\input{math_commands.tex}

\usepackage{xcolor}
\usepackage{colortbl}
\usepackage{hyperref}
\usepackage{url}
\usepackage{amsmath}
\usepackage{amsthm}
\usepackage{amssymb}
\usepackage{dsfont}
\usepackage[capitalize,nameinlink]{cleveref}[0.19]
\usepackage{enumerate}
\usepackage[shortlabels]{enumitem}
\setlist[itemize]{topsep=0em, itemsep=0em, partopsep=0em, parsep=0.5em}
\allowdisplaybreaks
\usepackage{mdframed}
\usepackage[normalem]{ulem}
\usepackage[most]{tcolorbox}   

\usepackage{tikz}
\usetikzlibrary{backgrounds}
\usetikzlibrary{arrows,shapes}
\usetikzlibrary{tikzmark}
\usetikzlibrary{calc}

\usepackage{graphicx}
\usepackage{svg}
\usepackage{microtype}
\usepackage{booktabs} % for professional tables

\usepackage{caption}
\usepackage{subcaption}

\usepackage{booktabs, multicol, multirow, diagbox}

\usepackage{fontawesome5}

\newtheorem{theorem}{Theorem}
\newtheorem{lemma}[theorem]{Lemma}
\newtheorem{proposition}[theorem]{Proposition}
\newtheorem{corollary}[theorem]{Corollary}
\newtheorem{definition}{Definition}

\crefname{definition}{Definition}{Definitions}
\crefname{assumption}{Assumption}{Assumptions}
\crefname{theorem}{Theorem}{Theorems}
\crefname{remark}{Remark}{Remarks}
\crefname{lemma}{Lemma}{Lemmas}
\crefname{corollary}{Corollary}{Corollaries}
\crefname{proposition}{Proposition}{Propositions}
\crefname{section}{Section}{Sections}
\crefname{subsection}{Subsection}{Subsections}
\crefname{example}{Example}{Examples}
\crefname{table}{Table}{Tables}
\crefname{problem}{Problem}{Problems}
\crefname{algorithm}{Algorithm}{Algorithms}
\crefname{figure}{Figure}{Figures}
\crefname{property}{Property}{Properties}

\crefformat{equation}{#2Equation~(#1)#3}
\crefformat{inequality}{#2Inequality~{}(#1){}#3}
\crefrangeformat{inequality}{#3Inequality~(#1)#4--#5(#2)#6}
\Crefformat{section}{#2Section~#1#3}
\Crefformat{page}{#2Page~#1#3}

\newcommand{\bw}{\mathbf{w}}
\newcommand{\bW}{\mathbf{W}}

\newcommand{\bp}{\mathbf{P}}
\newcommand{\bI}{\mathbf{I}}

\newcommand{\xcal}{\mathcal{X}}
\newcommand{\ycal}{\mathcal{Y}}

\newcommand{\dcal}{\mathcal{D}}

\newcommand{\ebb}{\mathbb{E}}
\newcommand{\rbb}{\mathbb{R}}

\colorlet{LightBlue}{blue!39!white} 
\colorlet{DarkBlue}{blue!70!black} 
\colorlet{VeryLightBlue}{blue!30!white} 
\colorlet{LightRed}{red!35!white} 
\newcommand{\unaryminus}{\scalebox{0.75}[1.0]{\( - \)}}
\newcommand{\tinyminus}{\scalebox{0.5}[1.0]{\( - \)}}
\newcommand{\unaryplus}{\scalebox{0.75}[1.0]{\( + \)}}
\newcommand{\unaryequal}{\scalebox{0.75}[1.0]{\( = \)}}

\newcommand{\unarycdot}{\scalebox{0.75}[1.0]{\( \cdot \)}}

\newcommand{\unaryrightarrow}{\scalebox{0.75}[1.0]{\( \rightarrow \)}}
\newcommand{\unaryapprox}{\scalebox{0.75}[1.0]{\( \approx \)}}

\newcommand{\hlmath}[2]{\colorbox{#1!17}{$\displaystyle #2$}}
\newcommand{\hltext}[2]{\colorbox{#1!17}{#2}}

\makeatletter
\newcommand{\biggg}{\bBigg@{3}}
\newcommand{\vast}{\bBigg@{4}}
\makeatother

\usepackage[accepted]{icml2023}

\icmltitlerunning{Decentralized SGD and Average-direction SAM are Asymptotically Equivalent}

\begin{document}

\twocolumn[
\icmltitle{Decentralized SGD and Average-direction SAM are Asymptotically Equivalent}



\icmlsetsymbol{equal}{*}

\begin{icmlauthorlist}
\icmlauthor{Tongtian Zhu}{zju}
\icmlauthor{Fengxiang He}{jde,ed}
\icmlauthor{Kaixuan Chen}{zju}
\icmlauthor{Mingli Song}{zju}
\icmlauthor{Dacheng Tao}{usyd}
\end{icmlauthorlist}

\icmlaffiliation{zju}{College of Computer Science and Technology, Zhejiang University}
% \icmlaffiliation{sias}{Shanghai Institute for Advanced Study of Zhejiang University}
\icmlaffiliation{jde}{JD Explore Academy, JD.com, Inc.}
\icmlaffiliation{ed}{Artificial Intelligence and its Applications Institute, School of Informatics, University of Edinburgh}
\icmlaffiliation{usyd}{The University of Sydney}
% \icmlaffiliation{zju}{Zhejiang University City College} 

\icmlcorrespondingauthor{Fengxiang He}{F.He@ed.ac.uk}

\icmlkeywords{Generalization, Decentralized Learning}

\vskip 0.3in
]



\printAffiliationsAndNotice{}  % leave blank if no need to mention equal contribution

\input{section/0-abstract}
\input{section/1-introduction}
\input{section/2-related_work}

\input{section/3-preliminaries}
\input{section/4-theoretical_results}

\input{section/5-empirical_results}

\input{section/7-conclusion}

\input{section/acknowledgement}
\bibliography{example_paper}
\bibliographystyle{icml2022}
\input{section/appendix}

\end{document}

%% file: math_commands.tex
\usepackage{amsmath,amsfonts,bm}

\def\eqref#1{equation~\ref{#1}}
\def\1{\bm{1}}

\def\mH{{\bm{H}}}

\def\mL{{\bm{L}}}
\def\mM{{\bm{M}}}

\def\mT{{\bm{T}}}

\DeclareMathAlphabet{\mathsfit}{\encodingdefault}{\sfdefault}{m}{sl}
\SetMathAlphabet{\mathsfit}{bold}{\encodingdefault}{\sfdefault}{bx}{n}

\DeclareMathOperator{\Tr}{Tr}

%% file: section/0-abstract.tex
\begin{abstract}
   Decentralized stochastic gradient descent (D-SGD) allows collaborative learning on massive devices simultaneously without the control of a central server. However, existing theories claim that decentralization invariably undermines generalization. In this paper, we challenge this conventional belief and present a completely new perspective for understanding decentralized learning. We prove that D-SGD implicitly minimizes the loss function of an average-direction sharpness-aware minimization (SAM) algorithm under general non-convex non-$\beta$-smooth settings. This surprising asymptotic equivalence reveals an intrinsic regularization-optimization trade-off and three advantages of decentralization: (1) there exists a free uncertainty evaluation mechanism in D-SGD to improve posterior estimation; (2) D-SGD exhibits a gradient smoothing effect;  and (3) the sharpness regularization effect of D-SGD does not decrease as total batch size increases, which justifies the potential generalization benefit of D-SGD over centralized SGD (C-SGD) in large-batch scenarios. Experiments support our theory and the code is available at
   \href{https://github.com/Raiden-Zhu/ICML-2023-DSGD-and-SAM}{\faGithub~D-SGD and SAM}.
\end{abstract}

%% file: section/1-introduction.tex
\section{Introduction}
\label{introduction}

Decentralized training harnesses the power of locally connected computing resources while preserving privacy \citep{warnat2021swarm, borzunov2022petals, yuan2022decentralized, beltran2022decentralized, JMLR:v24:22-0044}.
Decentralized stochastic gradient descent (D-SGD) is a popular decentralized training algorithm which enables simultaneous model training on a massive number of workers without the need for a central server \citep{xiao2004fast, lopes2008diffusion, nedic2009distributed, NIPS2017_f7552665, pmlr-v119-koloskova20a}. In D-SGD, each worker communicates only with its directly connected neighbors; see a detailed background in \cref{subsec: decentralized-learning}. This decentralization avoids the requirements of a costly central server with heavy communication burdens and potential privacy risks. The existing literature demonstrates that the massive models on edge can converge to a steady consensus model \citep{lu2011gossip, shi2015extra}, with %equal 
asymptotic linear speedup in convergence rate \citep{NIPS2017_f7552665} similar to distributed centralized SGD (C-SGD) \citep{dean2012large, li2014communication}. Therefore, D-SGD offers a promising distributed learning solution with significant advantages in communication efficiency \citep{ying2021bluefog}, privacy \citep{nedic2020distributed} and scalability \citep{NIPS2017_f7552665}.

Despite these merits, it is regrettable that the existing theories claim decentralization to invariably undermines generalization \citep{sun2021stability, zhu2022topology, deng2023stability}, which contradicts the following unique phenomena in decentralized deep learning:

\begin{itemize}[leftmargin=*]
    \item D-SGD can outperform C-SGD in large-batch settings, achieving higher validation accuracy and smaller validation-training accuracy gap, despite both being fine-tuned \citep{zhang2021loss};
    \item A non-negligible consensus distance (see \cref{eq: consensus-distance}) at middle phases of decentralized training can improve generalization over centralized training \citep{kong2021consensus}.  
\end{itemize}

% \footnote{There is no layer-wise scaling of learning rate \citep{You2020Large} in \citep{zhang2021loss}.}

These unexplained phenomena indicates the existence of a non-negligible \textbf{Gap} between existing theories and deep learning experiments, which we attribute to the overlook of important characteristics of decentralized learning in existing literature. Accordingly, the primary \textbf{Goal} of our study is to thoroughly investigate the underexamined characteristics of decentralized learning, in an effort to to bridge the gap.

\begin{figure*}[t!]
\begin{subfigure}[ResNet-18 on CIFAR-10 (C-SGD), 128 total batch size]{.33\textwidth}
  \centering
  % include fourth image
  \includegraphics[width=1.0\linewidth]{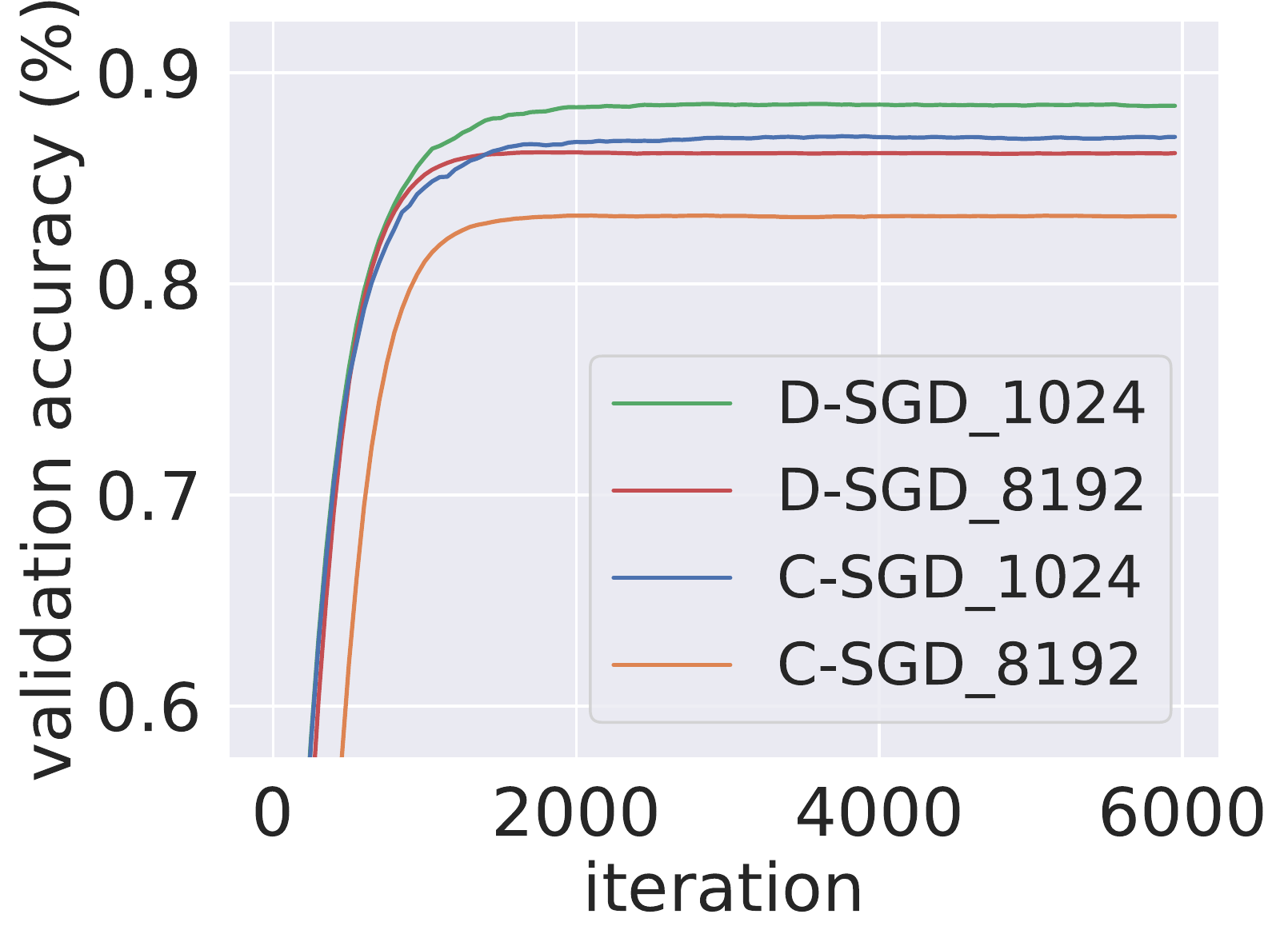}  
  \caption{AlexNet}
\end{subfigure}
\begin{subfigure}[ResNet-18 on CIFAR-10 (C-SGD), 1024 total batch size]{.33\textwidth}
  \centering
  % include fourth image
  \includegraphics[width=1.0\linewidth]{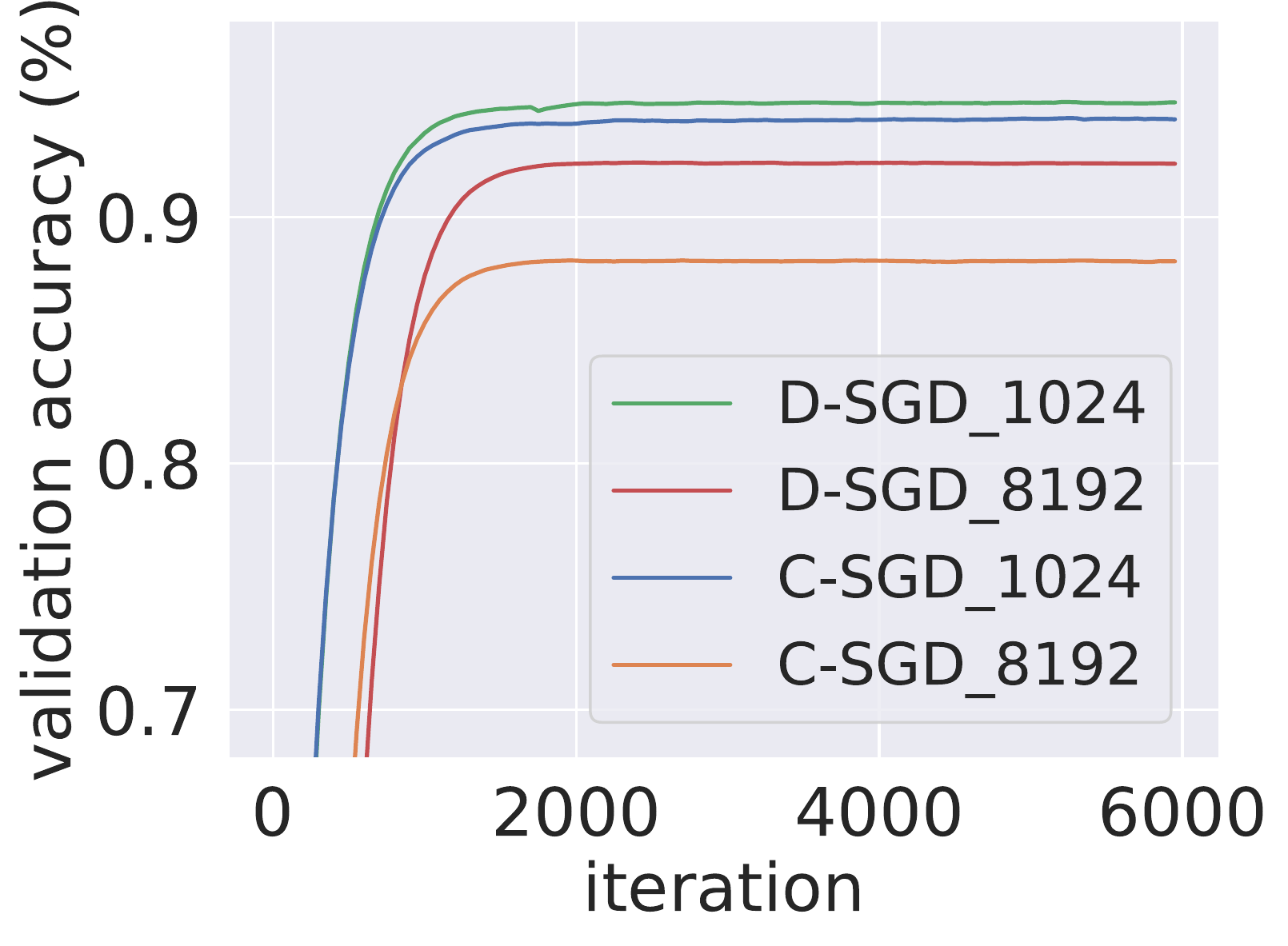}  
  \caption{ResNet-18}
\end{subfigure}
\begin{subfigure}[DenseNet-121 on CIFAR-10 (C-SGD), 8196 total batch size]{.33\textwidth}
  \centering
  % include fourth image
  \includegraphics[width=1.0\linewidth]{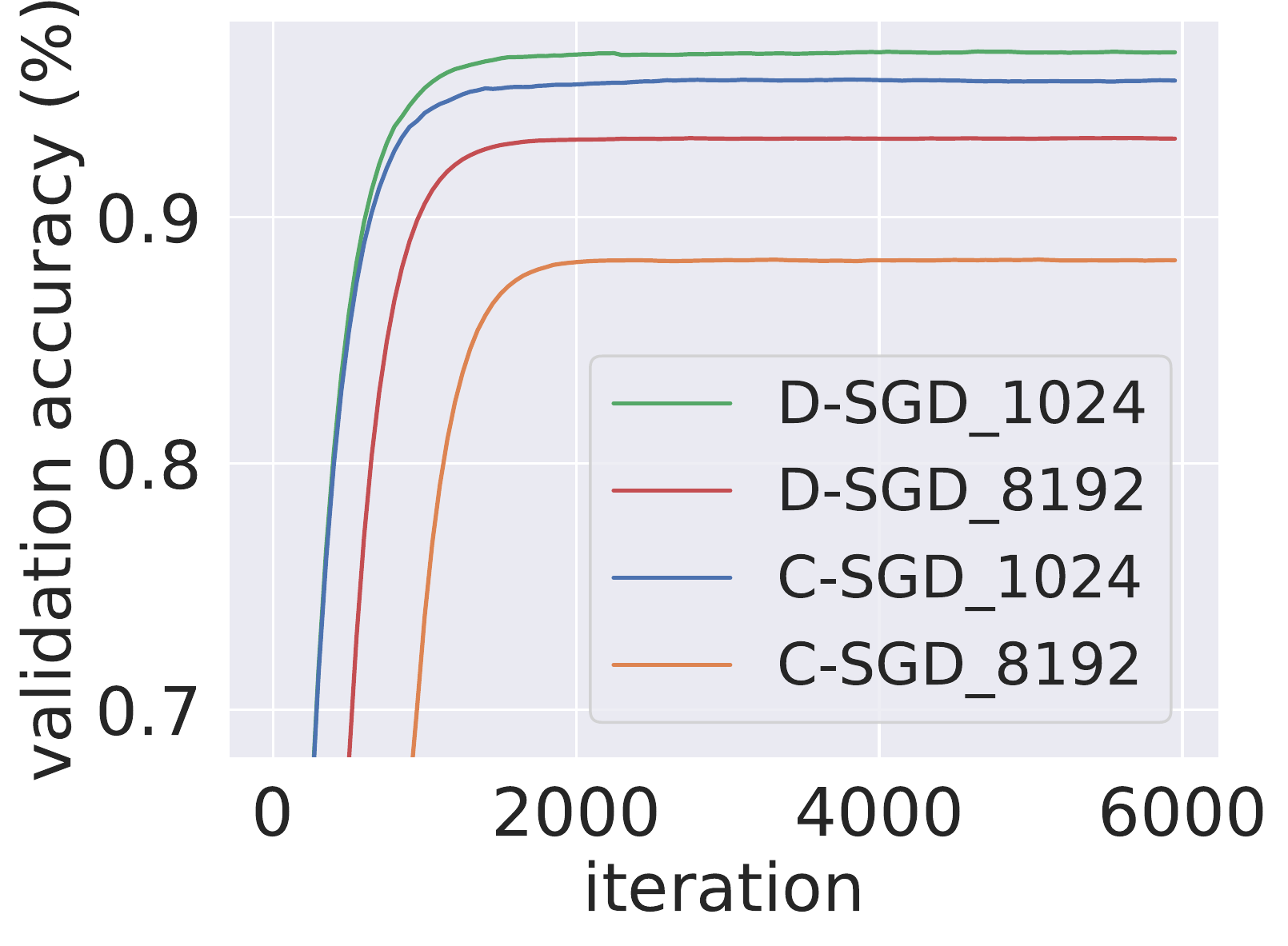}
  \caption{DenseNet-121}
\end{subfigure}
\caption{The validation accuracy comparison of C-SGD and D-SGD (ring topology) on CIFAR-10. The number of workers is set as 16, with a local batch size of 64 and 512 per worker, resulting in total batch sizes of 1024 and 8192, respectively. Validation accuracy comparison of C-SGD and D-SGD  with other topologies and on Tiny ImageNet are included in \cref{fig: test-accuracy_2} and \cref{fig: test-accuracy_3}, respectively. The training accuracy is almost $100\%$ everywhere. Exponential moving average is employed to smooth the validation accuracy curves. The training setting is included in \cref{sec: additional-empirical-result}.}
\label{fig: test-accuracy}
\end{figure*}

Directly analyzing the dynamics of the diffusion-like decentralized learning systems, instead of relying on upper bounds, can be challenging. Instead, we aim to establish a relationship between D-SGD and other centralized training algorithms.
In recent years, there has been a growing interest in techniques that aim to improve the generalization of deep learning models. One of the most popular techniques is sharpness-aware minimization (SAM) \citep{foret2021sharpnessaware,kwon2021asam,zhuang2022surrogate,du2022efficient,pmlr-v162-kim22f}, which is designed to explicitly minimize a sharpness-based perturbed loss; see the detailed background of SAM in \cref{subsec: intro-SAM}. 
% Specifically, SAM is a mini-max type algorithm which modifies the original loss function to evaluate the maximum loss value within the a small neighborhood around the current iterate. 
Empirical studies have shown that SAM substantially improves the generalization of multiple architectures, including convolutional neural networks \citep{wu2020adversarial, foret2021sharpnessaware}, vision transformers \citep{dosovitskiy2021an} and large language models \citep{bahri-etal-2022-sharpness}.
Average-direction SAM \citep{howsharpness2023}, a kind of SAM variants where sharpness is calculated as the (weighted) average within a small neighborhood around the current iterate, has been shown to generalize on par with vanilla SAM \citep{ujvary2022rethinking}. 

In this paper, we  provide a completely new perspective for understanding decentralized learning by showing that

\begin{tcolorbox}[notitle, rounded corners, colframe=darkgrey, colback=white, boxrule=2pt, boxsep=0pt, left=0.15cm, right=0.17cm, enhanced, shadow={2.5pt}{-2.5pt}{0pt}{opacity=5,mygrey},toprule=2pt, before skip=0.65em, after skip=0.75em 
    %   boxed title style={%
    %     rounded corners, 
    %     % borderline={1cm}{1cm}{SaddleBrown,solid},
    %     rounded corners=northwest, 
    %     colback=tcbcolframe, 
    %     boxrule=0pt, boxsep=3pt},
    %     attach boxed title to top left={
    %     yshift= -0.5mm},
    %   fonttitle=\large,
    %   title={\textbf{Motivating question}}
    ]
\emph{
    % {\centering  D-SGD and average-direction SAM are asymptotically equivalent.\\}
    {\centering {\fontsize{8pt}{13.2pt}\selectfont D-SGD and average-direction SAM are asymptotically equivalent.}\\}
}
\end{tcolorbox}

Specifically, our contributions are summarized below. 
\begin{itemize}[leftmargin=*]
    \item We prove that D-SGD asymptotically minimizes the loss function of an average-direction sharpness-aware minimization algorithm with zero additional computation (see \cref{th: dsgd-sam}), which directly connects decentralized training and centralized training. The asymptotic equivalence also implies a regularization-optimization trade-off in decentralized training. Our theory is applicable to arbitrary communication topologies (see \cref{def:dou-matrix}) and general \textbf{non-convex and non-$\beta$-smooth} (see \cref{def:l-smooth}) objectives (e.g., deep neural networks training). 
    % To prove the main results, we apply a second-order multivariate Taylor expansion \citep{konigsberger2013analysis} on the gradient diversity (see \cref{eq: taylor-expansion}) and derive a Hessian-dependent noise in D-SGD.  
    \item The equivalence further reveals the potential benefits of the decentralized training paradigm, which challenges the previously held belief that centralized training is optimal. We demonstrate three advantages of training with decentralized models based on the equivalence: (1) there exists a surprising free uncertainty estimation mechanism in D-SGD, where the weight diversity matrix $\bm{\Xi}{\scriptstyle (t)}$ is learned to estimate $\Sigma_q$, the intractable covariance of the posterior;(2) D-SGD has a gradient smoothing effect, which improves training stability (see \cref{coro: smoothing}); and that (3) the sharpness regularizer of D-SGD does not decrease as the total batch size increases (see \cref{th: large-batch}), which justifies the superior generalizability of D-SGD than C-SGD, especially in large-batch settings where gradient variance remains low. Our empirical results also fully support our theory (see \cref{fig: test-accuracy} and \cref{fig: 3d-minima-cifar10}).
\end{itemize}

To the best of our knowledge, our work is the first to establish the equivalence of D-SGD and average-direction SAM, which constitutes a direct connection between decentralized training and centralized training algorithms.
This breakthrough makes it easier to analyze the diffusion-like decentralized systems, whose exact dynamics were considered challenging to understand. The theory further sheds light on the potential benefits of decentralized training paradigm.
While our theory primarily focuses on vanilla D-SGD, it could be directly extended to general decentralized gradient-based algorithms.
 We anticipate the insights derived from our work will help bridge the decentralized learning and SAM communities, and pave the way for the development of fast and generalizable decentralized algorithms.

%% file: section/2-related_work.tex
\section{Related work}
\label{related_work}

\begin{figure*}[ht!]
  \centering
  \includegraphics[width=0.9\linewidth]{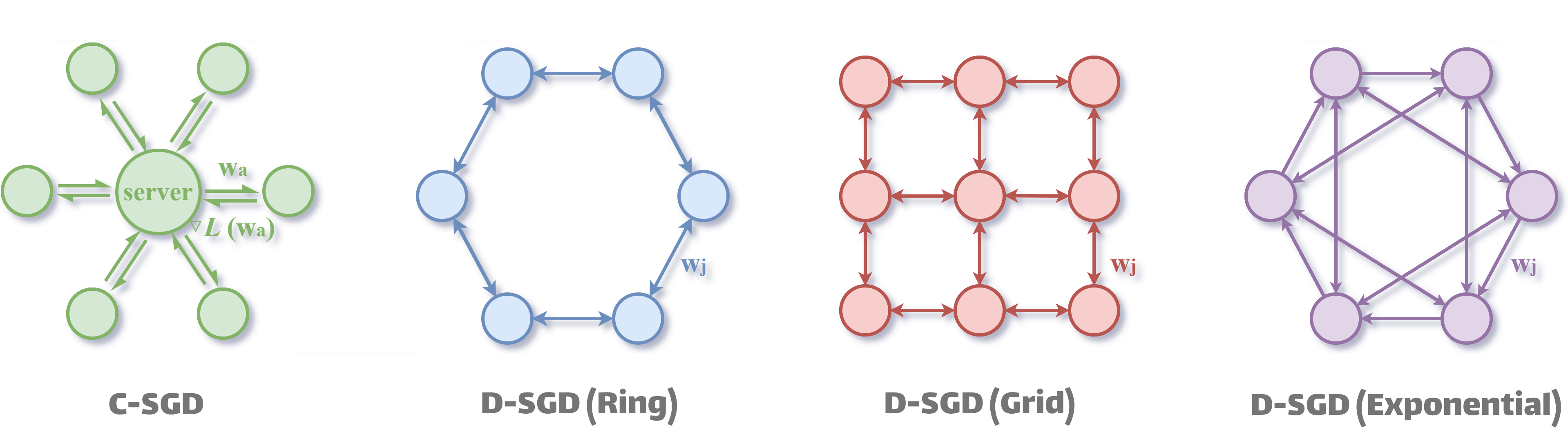}  
  \caption{An illustration of centralized SGD and decentralized SGD. In C-SGD there is a central server aggregating global information while D-SGD relies only on peer-to-peer communication to diffuse information among workers.}
\label{fig: centralized-decentralized}
\end{figure*}

\textbf{Flatness and generalization.}
The flatness of the minimum has long been regarded as a proxy of generalization in the machine learning literature \citep{hochreiter1997flat, NIPS2017_10ce03a1,  izmailov2018averaging,  Jiang*2020Fantastic}. 
Intuitively, the loss around a flat minimum varies slowly in a large neighborhood, while a sharp minimum increases rapidly in a small neighborhood \citep{hochreiter1997flat}.
Through the lens of the minimum description length theory \citep{rissanen1983universal},  flat minimizers, due to their specification with lower precision, tend to generalize better than sharp minimizers \citep{keskar2017on}. From a Bayesian perspective, sharp minimizers have highly concentrated posterior parameter distributions, indicating a greater specialization to the training set and thus are less robust to data perturbations than flat minimizers \citep{mackay1992practical, chaudhari2019entropy}.

\textbf{Generalization of D-SGD.} 
Recently works by \citet{sun2021stability}, \citet{zhu2022topology} and \citet{deng2023stability} prove that decentralization introduces an additional positive term into the generalization error bounds, which suggests that decentralization may hurt generalization. However, these conservative upper bounds do not account for the unique phenomena in decentralized learning, such as the superior generalization performance of D-SGD in large batch scenarios.
\citet{gurbuzbalaban2022heavy} offers an alternative perspective, showing that D-SGD with large, sparse topology has a heavier tail in parameter distribution than C-SGD in some cases, indicating better generalizability potential.
Compared with \citet{gurbuzbalaban2022heavy}, our theory is generally applicable to arbitrary communication topologies and learning rates.
Another work by \citet{zhang2021loss} demonstrates that decentralization introduces a landscape-dependent noise, which may improve the convergence of D-SGD. However, the impact of noise on the local geometry of the D-SGD trajectory and its effect on generalization remains unexplored. In contrast, we theoretically justify the flatness-seeking behavior of Hessian-dependent noise in D-SGD and then establish the asymptotic equivalence between D-SGD and SAM. Based on the equivalence, we prove that D-SGD has superior potential in generalizability compared with C-SGD, especially in large-batch settings.

%% file: section/3-preliminaries.tex
\section{Preliminaries}
\label{preliminaries}

%\textbf{Supervised learning.} 
\textbf{Basic notations.}
Suppose that $\xcal \subseteq \rbb^{d_x}$ and $\ycal \subseteq \rbb$ are the input and output spaces, respectively. We denote the training set as $\mu=\left\{z_{1}, \ldots, z_{N}\right\}$, where $z_{\zeta}=\left(x_{\zeta}, y_{\zeta}\right), \zeta=1,\dots, N $ are sampled independent and identically distributed (i.i.d.) from an unknown data distribution $\dcal$ defined on $\mathcal{Z}=\mathcal{X} \times \mathcal{Y}$. 
The goal of supervised learning is to learn a predictor (or hypothesis) $g(\bw; \cdot )$, parameterized by $\bw\in \rbb^d$ of an arbitrary finite  dimension $d$, to approximate the mapping between the input variable $x\in \mathcal{X}$ and the output variable $y\in \mathcal{Y}$, based on the training set $\mu$. The function $c: \mathcal{Y} \times \mathcal{Y} \mapsto \mathbb{R}^{+}$ are defined to evaluate the prediction performance of hypothesis $g$. The loss of a hypothesis $g$ with respect to (w.r.t.) the example $z_{\zeta}=(x_{\zeta}, y_{\zeta})$ is denoted by $
\mL(\bw; z_{\zeta})=c(g(\bw;x_{\zeta}), y_{\zeta})$, which measures the effectiveness of the learned model $\bw$. The empirical and population risks of $\bw$ are then defined as follows:
\begin{equation*}
    \mL^{\mu}_{\bw}=\frac{1}{N}\sum_{\zeta=1}^{N} \mL(\bw;z_\zeta),~~  \mL_{\bw}=\ebb_{z\sim D}[\mL(\bw;z)].
\end{equation*}
\textbf{Distributed learning.}
Distributed learning trains a model $\bw$ jointly using multiple workers 
\citep{shamir2014distributed}. In this framework, the $j$-th worker $(j\unaryequal 1,\dots, m)$ can access local training examples $\mu_j\unaryequal\{z_{j,1}, \ldots, z_{j,|\mu_j|}\}$, drawn from the local empirical distribution $\mathcal{D}_j$. In this setup, the global empirical risk of $\bw$ becomes
\begin{align*}
    \mL^{\mu}_{\bw} = \frac{1}{m}\sum_{j=1}^{m} \mL^{\mu_j}_{\bw},
\end{align*}
where $\mL^{\mu_j}_{\bw} = \frac{1}{|\mu_j|}\sum_{\zeta=1}^{|\mu_j|}\mL({\bw};z_{j, \zeta})$
denotes the local empirical risk on the $j$-th worker and $z_{j, \zeta}\in\mu_{j}$, where $\zeta =1,\dots, |\mu_j|$, represent the local training data.

\textbf{Distributed centralized stochastic gradient  descent (C-SGD).}\footnote{``Centralized" refers to the fact that in C-SGD, there is a central server receiving local weights or gradients information (see \cref{fig: centralized-decentralized}). C-SGD defined above is identical to the FedAvg algorithm \citep{pmlr-v54-mcmahan17a} under the condition that the local step is set as $1$ and all local workers are selected by the server in each round (see \cref{subsec: centralized-learning}). To avoid misunderstandings, we include the term ``distributed" in C-SGD to differentiate it from traditional single-worker SGD \citep{cauchy1847methode, Robbins1951ASA}.} 
In C-SGD \citep{dean2012large, li2014communication}, the de facto distributed training algorithm, there is only one centralized model $\bw_{a}{\scriptstyle (t)}$.
C-SGD updates the model by
\begin{align}\label{def: c-sgd}
    \bw_{a}{\scriptstyle (t+1)} 
    = \bw_{a}{\scriptstyle (t)}-\eta\cdot\frac{1}{m}\sum_{j=1}^{m}\overbrace{  \nabla\mL^{\mu_j{\scriptstyle (t)}}(\bw_{a}{\scriptstyle (t)})}^{\text {Local gradient computation}},
\end{align}
where $\eta$ denotes the learning rate (step size), $\mu_j{\scriptstyle (t)}=\{z_{j,1}, \ldots, z_{j,|\mu_j{\scriptstyle (t)}|}\}$ denotes the local training batch independent and identically distributed (i.i.d.) drawn from the local empirical data distribution $\mathcal{D}_j$ at the $t$-th iteration, and $\nabla \mL^{\mu_j{\scriptstyle (t)}}_{\bw}=\nabla \mL^{\mu_j{\scriptstyle (t)}}(\bw) = \frac{1}{|\mu_j{\scriptstyle (t)}|}\sum_{\zeta{\scriptstyle (t)}=1}^{|\mu_j{\scriptstyle (t)}|}\nabla\mL({\bw};z_{j, \zeta{\scriptstyle (t)}})$ stacks for the local mini-batch gradient of $\mL$ w.r.t. the first argument $\bw$. The total batch size of C-SGD at the $t$-th iteration is $|\mu{\scriptstyle (t)}| = \sum_{j=1}^{m}|\mu_j{\scriptstyle (t)}|$.
Please refer to \cref{subsec: centralized-learning} for more details of (distributed) centralized learning.
In the next section, we will show that C-SGD equals the single-worker SGD with a larger batch size. 

\textbf{Decentralized stochastic gradient  descent (D-SGD).} In model decentralization scenarios, only peer-to-peer communication is allowed.
The goal of D-SGD in the setup is to learn a consensus model $\bw_{a}{\scriptstyle (t)}=\frac{1}{m}\sum_{j=1}^m\bw_{j}{\scriptstyle (t)}$ on $m$ workers through gossip communication, where $\bw_{j}{\scriptstyle (t)}$ stands for the $d$-dimensional local model on the $j$-th worker.
We denote ${\bp} = [\bp_{j,k}] \in \mathbb{R}^{m\times m}$ as a doubly stochastic gossip matrix (see \cref{def:dou-matrix}) that characterizes the underlying topology $\mathcal{G}$. The vanilla Adapt-While-Communicate  (AWC) version of the mini-batch D-SGD \citep{nedic2009distributed, NIPS2017_f7552665} updates the model on the $j$-th worker by
\begin{equation}\label{def: dec-sgd}
    \bw_{j}{\scriptstyle (t+1)} = \overbrace{\sum_{k=1}^{m}\bp_{j,k} \bw_{k}{\scriptstyle (t)}}^{\text {Communication}}-\eta\cdot\overbrace{  \nabla\mL^{\mu_j{\scriptstyle (t)}}(\bw_{j}{\scriptstyle (t)})}^{\text {Local gradient computation}}.
\end{equation}
For a more detailed background of decentralized learning, please kindly refer to \cref{subsec: decentralized-learning}.

%% file: section/4-theoretical_results.tex
\section{Theoretical results}
In this section, we establish the equivalence between D-SGD and average-direction SAM under general non-convex and non-$\beta$-smooth objectives $\mL$. We also provide a proof sketch to offer a more intuitive understanding.
The equivalence further showcases the potential superiority of learning with decentralized models.
Specifically, we prove that the additional noise introduced by decentralization leads to a gradient smoothing effect, which could stabilize optimization. 
Additionally, we show that the sharpness regularizer in D-SGD does not decrease as the total batch size increases, which guarantees generalizability in large-batch settings.

\subsection{Equivalences of Decentralized SGD and average-direction SAM}\label{sec: dsgd-sam}
In this subsection, we prove that D-SGD implicitly performs average-direction sharpness-aware minimization (SAM), followed by detailed implications. 

\begin{theorem}[D-SGD as SAM\label{th: dsgd-sam}]
Suppose $\mL \in C^4\left(\mathbb{R}^d\right)$, i.e., $\mL$ is four times continuously differentiable. The mean iterate of the global averaged model of D-SGD, defined by $\bw_{a}{\scriptstyle (t)}=\frac{1}{m}\sum_{j=1}^m\bw_{j}{\scriptstyle (t)}$, is provided as follows:\footnote{Taking expectation over super batch $\mu{\scriptscriptstyle (t)}$ means taking expectations over all local mini-batches $\mu_j{\scriptscriptstyle (t)}$ for all $j\unaryequal1,\dots, m$, which eliminates the randomness of all training data at $t$-th iteration, represented by $z_{j, \zeta{\scriptstyle (t)}}$ for all $\zeta{\scriptstyle (t)}\unaryequal1,\dots,|\mu_j{\scriptstyle (t)}|$ and  $j\unaryequal1,\dots, m$. 

In D-SGD (see \cref{def: dec-sgd}), the ``virtual" global averaged model $\bw_{a}{\scriptstyle (t)}=\frac{1}{m}\sum_{j=1}^m\bw_{j}{\scriptstyle (t)}$ is primarily employed for theoretical analysis, since there is no central server to aggregate the information from local workers. However, analyzing $\bw_{a}{\scriptstyle (t)}$ remains practical as it characterizes the overall performance of D-SGD.} 
\begin{align*}
\ebb_{\mu{\scriptscriptstyle (t)}}&[\bw_{a}{\scriptstyle (t+1)}]\nonumber\\
=&\bw_{a}{\scriptstyle (t)}-\eta\color{DarkBlue}{\underbrace{\color{black}{\tikzmarknode{descent_direction}{\hlmath{LightBlue}{\ebb_{\epsilon \sim \mathcal{N}(0, \bm{\Xi}{\scriptstyle (t)})}[\nabla\mL_{\bw_{a}{\scriptstyle (t)}+\epsilon}]}}}}_{\text{asymptotic descent direction}}}\nonumber\\
&\unaryplus\underbrace{\mathcal{O}(\eta\ \ebb_{\epsilon \sim \mathcal{N}(0, \bm{\Xi}{\scriptstyle (t)})}\|\epsilon\|_2^3\unaryplus\frac{\eta}{m}\sum_{j=1}^{m}\|\bw_{j}{\scriptstyle (t)}\unaryminus\bw_{a}{\scriptstyle (t)}\|_2^3)}_{\text{higher-order residual terms}},
\end{align*}
% \begin{tikzpicture}[overlay,remember picture,>=stealth,nodes={align=right,inner ysep=1pt},<-]
%     % For "S"
%     \path (descent_direction.south) ++ (0,-1.5em) node[anchor=north west,color=blue!67] (mean){asymptotic descent direction};
%     \draw [color=blue!57](descent_direction.south) |- ([xshift=-0.3ex,color=blue!67]mean.south east);
% \end{tikzpicture}
 where $\bm{\Xi}{\scriptstyle (t)}= \frac{1}{m}\sum_{j=1}^{m}(\bw_{j}{\scriptstyle (t)}\unaryminus\bw_{a}{\scriptstyle (t)}){(\bw_{j}{\scriptstyle (t)}\unaryminus\bw_{a}{\scriptstyle (t)})}^{\top} \in \rbb^{d\times d}$ denotes the weight diversity matrix and $\nabla\mL_{\bw_{a}{\scriptstyle (t)}+\epsilon}$ denotes the gradient value of $\mL(\bw)$ at $\bw_{a}{\scriptstyle (t)}+\epsilon$, i.e., $\nabla\mL({\bw})|_{\bw={\bw_{a}{\scriptstyle (t)}+\epsilon}}$.
\end{theorem}
The proof is deferred to \cref{sec:proof-dsgd-sam}.

\textbf{Asymptotic equivalence.}
According to \cref{eq: taylor-expansion} and \cref{prop: order-of-residuals}, $\ebb_{\epsilon \sim \mathcal{N}(0, \bm{\Xi}{\scriptstyle (t)})}[\nabla\mL_{\bw_{a}{\scriptstyle (t)}+\epsilon}]$ is of the order $\mL_{\bw_{a}{\scriptstyle (t)}}\unaryplus\mathcal{O}{(\frac{1}{m}\sum_{j=1}^{m}\|\bw_{j}{\scriptstyle (t)}\unaryminus\bw_{a}{\scriptstyle (t)}\|_2^2)}$ while the residuals are of the higher-order $\mathcal{O}{(\frac{1}{m}\sum_{j=1}^{m}\|\bw_{j}{\scriptstyle (t)}\unaryminus\bw_{a}{\scriptstyle (t)}\|_2^3)}$. Therefore, $\ebb_{\epsilon \sim \mathcal{N}(0, \bm{\Xi}{\scriptstyle (t)})}[\nabla\mL_{\bw_{a}{\scriptstyle (t)}+\epsilon}]$ gradually dominates the optimization direction as the local models are near consensus (i.e., $\bw_{j}{\scriptstyle (t)}\unaryrightarrow \bw_{a}{\scriptstyle (t)}, \forall j$) and the descent direction of D-SGD asymptotically approaches $\ebb_{\epsilon \sim \mathcal{N}(0, \bm{\Xi}{\scriptstyle (t)})}[\nabla\mL_{\bw+\epsilon}]$. See \cref{def: asymptotic-equivalence} for details on the asymptotic equivalence.
% \ztt{Although residuals are negligible only when the local models are near consensus, the equivalence is still significant in determining the limiting dynamics and further the generalizability of D-SGD.}

\textbf{Sharpness regularization.}
According to \cref{th: dsgd-sam}, D-SGD asymptotically optimizes $\ebb_{\epsilon \sim \mathcal{N}(0, \bm{\Xi}{\scriptstyle (t)})}[\mL_{\bw+\epsilon}]$, an averaged perturbed loss in a ``basin” around $\bw$, rather than the original point-loss.
To further clarify, we can split ``true objective" of D-SGD near consensus into the original loss plus an average-direction sharpness:
\begin{align*}
    \ebb_{\mu{\scriptstyle (t)}}[\mL_\bw^{\text{\tiny D-SGD}}]\approx\underbrace{\ \mL_{\bw}\ }_{\text{\textit{original loss}}}+\color{DarkBlue}{\underbrace{\color{black}{\tikzmarknode{sharpness-aware_regularizer}{\hlmath{LightBlue}{\ebb_{\epsilon \sim \mathcal{N}(0, \bm{\Xi}{\scriptstyle (t)})}[\mL_{\bw+\epsilon}-\mL_{\bw}]}}}}_{\text{\textit{sharpness-aware regularizer}}}}.
\end{align*}
The second term $\ebb_{\epsilon \sim \mathcal{N}(0, \bm{\Xi}{\scriptstyle (t)})}[\mL_{\bw+\epsilon}-\mL_{\bw}]$ measures the weighted average sharpness at $\bw$, which is a special form of the average-direction sharpness \citep{howsharpness2023}. 
\cref{th: dsgd-sam} proves that D-SGD minimizes the loss function of an average-direction SAM asymptotically, which provides a direct connection between decentralized learning and centralized learning.
As \cref{th: dsgd-sam} only assumes $\mL$ to be continuous and fourth-order differentiable, the result is generally applicable to \textbf{non-convex non-$\beta$-smooth} problems, including deep neural networks training.

% The generalization benefit of optimizing average-direction sharpness is discussed in \cref{th: generalization-bound}. This work reveals the sharpness regularization effect of D-SGD in one-step. We leave the study of the multi-step dynamics of D-SGD to future work.

We note that the sharpness regularizer in D-SGD is directly controlled by $\bm{\Xi}{\scriptstyle (t)}$, whose magnitude can be measured by the {\it consensus distance}, a key component characterizing the overall convergence of D-SGD \citep{kong2021consensus},
\begin{equation}
    \Tr{(\bm{\Xi}{\scriptstyle (t)})}= \frac{1}{m}\sum_{j=1}^{m}(\bw_{j}{\scriptstyle (t)}\unaryminus\bw_{a}{\scriptstyle (t)})^{\top}{(\bw_{j}{\scriptstyle (t)}\unaryminus\bw_{a}{\scriptstyle (t)})}.
    \label{eq: consensus-distance}
\end{equation}
\cref{th: dsgd-sam} provides a theoretical explanation for the phenomena observed in \citep{kong2021consensus}: (1) Controlling consensus distance below a threshold in the initial training phases makes the SAM-type term quickly dominates the residual terms, thus ensuring good optimization; (2) Sustaining a non-negligible consensus distance at the middle phases maintains the sharpness regularization effect and therefore improves generalization over centralized training.

The implicit regularization effect in D-SGD shares similar insights with interesting studies on local SGD and federated learning, revealing that global coherence is not always optimal and a certain degree of drift in clients could be benign \citep{localsgdgeneralizes2023, chor2023more}. Specifically, \citet{localsgdgeneralizes2023} proves that the dissimilarity between local models induces a gradient noise, which drives the iterate to drift faster to flatter minima. 
Despite the shared characteristics, the consensus distance in decentralized learning is notably unique. The magnitude of the consensus distance exhibits dynamic adjustments.
\cref{prop: consensus-distance-decrease} shows that if the learning rate is smaller than a certain threshold, then the consensus distance gradually decreases during training, indicating that the ``searching space'' of $\epsilon$ is relatively large in the initial training phase and then gradually declines. In addition, as shown in \cref{prop: consensus-distance}, a small spectral gap (see \cref{def:spectral-gap}) of the underlying communication topology (see \cref{fig: centralized-decentralized}) leads to larger consensus distance, the magnitude of perturbation radius.
According to \citet{foret2021sharpnessaware}, a large perturbation radius ensures a lower generalization upper bound. However, the validation performance of D-SGD is not always satisfactory on large and sparse topologies with a small spectral gap \citep{kong2021consensus}, as there is regularization-optimization trade-off in D-SGD (please refer to the discussion in \cref{sec: discussion}).

\textbf{Variational interpretation of D-SGD. \label{page: val-dsgd}} In the variational formulation \citep{zellner1988optimal}, $\min_{\bw}\ebb_{\epsilon \sim \mathcal{N}(0, \bm{\Xi}{\scriptstyle (t)})}[\mL_{\bw+\epsilon}]$ is referred to as the variational optimization \citep{rockafellar2009variational}. 
% \ztt{when the approximate posterior Q\mathcal{Q} is chosen as a Gaussian with mean \bw(t)\bw{\scriptstyle (t)} and covariance \bmΞ(t)\bm{\Xi}{\scriptstyle (t)}}.
 \cref{th: dsgd-sam} shows that D-SGD not only optimizes $\ebb_{\epsilon \sim \mathcal{N}(0, \bm{\Xi}{\scriptstyle (t)})}$ with respect to $\bw$, but also inherently estimates uncertainty for free: The weight diversity matrix $\bm{\Xi}{\scriptstyle (t)}$ (i.e., the empirical covariance matrix of $\bw_{j}{\scriptstyle (t)})$ implicitly estimate $\Sigma_q$, the intractable posterior covariance,
\begin{align*}
    \bm{\Xi}{\scriptstyle (t)}
    =\frac{1}{m}\sum_{j=1}^{m}(\bw_{j}{\scriptstyle (t)}\unaryminus\bw_{a}{\scriptstyle (t)}){(\bw_{j}{\scriptstyle (t)}\unaryminus\bw_{a}{\scriptstyle (t)})}^{\top}
    \approx \Sigma_q.
\end{align*}
Note that $\bm{\Xi}{\scriptstyle (t)}$ is ``used" implicitly along with the update of local models without any additional computational budget.
The free uncertainty estimation mechanism indicates the uniqueness of the noise from decentralization, which depends both on  the local loss landscape and the posterior.

\textbf{Comparison of D-SGD and vanilla SAM.} 
The loss function of vanilla SAM \citep{foret2021sharpnessaware} can be written in the following form:
\begin{align*}
\mL^{\text{\tiny SAM}}(\bw, \Sigma)\unaryequal\max_{\epsilon\in\rbb^d\mid\epsilon^T \Sigma^{-1} \epsilon \leq d}[\mL(\bw\unaryplus\epsilon)\unaryminus\mL(\bw)]\unaryplus\mL(\bw),
\end{align*}
where the covariance matrix $\Sigma$ is set as $\frac{\rho^2}{d} I$, with $\rho$ being the perturbation radius and $I$ representing the identity matrix. 
Interestingly, $\rho$ in SAM plays a similar role as $\bm{\Xi}{\scriptstyle (t)}$ in D-SGD.
However, in the SAM that D-SGD approximates, the covariance matrix $\bm{\Xi}{\scriptstyle (t)}$ is learned adaptively during training.
Moreover, the iterate of D-SGD involves higher-order residuals, whereas vanilla SAM does not.
The third difference is that vanilla SAM minimizes a worst-case sharpness $\max_{\epsilon^T \Sigma^{-1} \epsilon \leq d} L(\bw+\epsilon)$ while D-SGD implicitly minimizes an average-direction sharpness (or a Bayes loss). However, the loss of vanilla SAM always upper bounds the Bayes loss \citep{möllenhoff2023sam}, and they are close to each other in high dimensions where samples from $\mathcal{N}(\bw, \Sigma)$ concentrate around the ellipsoid $(\bw-\epsilon)^{\top} \Sigma^{-1}(\bw-\epsilon)=d$ \citep{vershynin2018high}.
In addition, the sharpness regularization effect of D-SGD incurs \textbf{zero} additional computational overhead compared to SAM, which requires performing backpropagation twice at each iteration.

\textbf{Comparison with related works.} 
Initial efforts have viewed D-SGD as a centralized algorithm penalizing the weight norm $\|\bp^{-\frac{1}{2}}\bW\|^2_{\bI-\bp}$, a quantity similar to the consensus distance, where $\bW= [\bw_{1}, \cdots, \bw_{m}]^T \in \rbb^{m\times d}$ collects $m$ local models \citep{yuan2021decentlam, gurbuzbalaban2022heavy}. However, little effort has been made so far to analyze the ``interplay'' between weight diversity measures, such as the consensus distance, and the local geometry of the D-SGD iterate. Our work fills this gap by showing the flatness-seeking behavior of the Hessian-consensus dependent noise in D-SGD and then exhibiting the asymptotic equivalence between D-SGD and SAM.

\subsection{Proof sketch and idea}
To impart a stronger intuition, we summarize the proof sketch of \cref{th: dsgd-sam} and explain the motivation behind our proof idea. Full proof is deferred to \cref{sec:proof}.

% The proof scheme for this part is organized as follows:

\textbf{(1) Derive the iterate of the averaged model $\bw_{a}{\scriptstyle (t)}$.}
% maintains proximity to the local models $\bw_{j}\ {\scriptstyle (t)}$ for all $j\unaryequal1,\dots, m$ \citep{on_the-convergence, JMLR:v23:19-939}.

Directly analyzing the dynamics of the diffusion-like decentralized systems where information is gradually spread across the network is non-trivial.
Instead, we focus on $\bw_{a}{\scriptstyle (t)}=\frac{1}{m}\sum_{j=1}^m\bw_{j}{\scriptstyle (t)}$, the global averaged model of D-SGD, whose update can be written as follows,
\begin{align}
&\bw_{a}{\scriptstyle (t+1)}\nonumber\\
&\unaryequal\bw_{a}{\scriptstyle (t)}\unaryminus\eta\big[\nabla \mL^{\mu{\scriptscriptstyle (t)}}_{\bw_{a}{\scriptstyle (t)}}
\unaryplus\underbrace{\frac{1}{m}\sum_{j=1}^{m}(\nabla \mL^{\mu_j{\scriptscriptstyle (t)}}_{\bw_{j}{\scriptstyle (t)}}\unaryminus\nabla \mL^{\mu_j{\scriptscriptstyle (t)}}_{\bw_{a}{\scriptstyle (t)}})}_{\text {\textit{gradient diversity among local workers}}}\big],\label{eq: graident-decomposition}
\end{align} 
as $\frac{1}{m}\sum_{j=1}^{m}\sum_{k=1}^{m}\bp_{j,k} \bw_{k}{\scriptstyle (t)}\unaryequal\frac{1}{m}\sum_{k=1}^{m}\bw_{k}{\scriptstyle (t)}\unaryequal\bw_{a}{\scriptstyle (t)}$.

\cref{eq: graident-decomposition} shows that decentralization introduces an additional noise, which characterizes the gradient diversity between the global averaged model $\bw_{a}{\scriptstyle (t)}$ and the local models $\bw_{j}{\scriptstyle (t)}$ for $j\unaryequal1, \dots, m$, compared with its centralized counterpart.\footnote{We note that the concept of gradient diversity is distinct from that in \citep{pmlr-v84-yin18a}, as it quantifies the variation of the gradients of one single model on different data points. The gradient diversity in our paper shares similarities with the gradient bias of local workers in federated learning (FL) literature \citep{wang2020tackling, reddi2021adaptive, wang2022unreasonable}. }
Therefore, we note that 
\begin{tcolorbox}[notitle, rounded corners, colframe=darkgrey, colback=lightblue, 
       boxrule=2pt, boxsep=0pt, left=0.15cm, right=0.17cm, enhanced, 
       shadow={2.5pt}{-2.5pt}{0pt}{opacity=5,mygrey},
       toprule=2pt, before skip=0.65em, after skip=0.75em 
    ]
\emph{
    % {\centering  D-SGD and average-direction SAM are asymptotically equivalent.\\}
    {analyzing the gradient diversity is the major challenge of decentralized (gradient-based) learning.}
}
\end{tcolorbox}

One can deduce directly from \cref{eq: graident-decomposition} that distributed centralized SGD, whose gradient diversity remains zero, equals the standard single-worker mini-batch SGD with an equivalently large batch size. 

\textbf{Insight.}
We also note that the gradient diversity equals to zero on quadratic objective $\mL$ (see \cref{prop: gradient-diversity-quadratic}). Therefore, the quadratic approximation of loss functions $\mL$ \citep{pmlr-v97-zhu19e, ibayashi2021quasi,liu2021noise,ziyin2022strength} might be insufficient to characterize how decentralization impacts the training dynamics of D-SGD, especially on neural network loss landscapes where quadratic approximation may not be accurate even around minima \citep{JML-1-247}. To better understand the dynamics of D-SGD on complex landscapes, it is crucial to consider \hltext{VeryLightBlue}{\textit{higher-order geometric information}} of objective $\mL$. In the following, we approximate the gradient diversity using Taylor expansion, instead of analyzing it on non-convex non-$\beta$-smooth loss $\mL$ directly, which is highly non-trivial.

\textbf{(2) Perform Taylor expansion on the gradient diversity.}
 Since $\mL \in C^4\left(\mathbb{R}^d\right)$, we can perform a second-order Taylor expansion on the gradient diversity around $\bw_{a}{\scriptstyle (t)}$:
\begin{align*}
\frac{1}{m}&\sum_{j=1}^{m}(\nabla \mL^{\mu_j{\scriptscriptstyle (t)}}_{\bw_{j}{\scriptstyle (t)}}\unaryminus\nabla \mL^{\mu_j{\scriptscriptstyle (t)}}_{\bw_{a}{\scriptstyle (t)}})
\unaryequal \frac{1}{m}\sum_{j=1}^{m} \mH^{\mu_j{\scriptscriptstyle (t)}}_{\bw_{a}{\scriptstyle (t)}}\unarycdot(\bw_{j}{\scriptstyle (t)}\unaryminus\bw_{a}{\scriptstyle (t)})\nonumber\\
&\unaryplus \frac{1}{2m}\sum_{j=1}^{m} \mT^{\mu_j{\scriptscriptstyle (t)}}_{\bw_{a}{\scriptstyle (t)}}\otimes[(\bw_{j}{\scriptstyle (t)}\unaryminus\bw_{a}{\scriptstyle (t)})(\bw_{j}{\scriptstyle (t)}\unaryminus\bw_{a}{\scriptstyle (t)})^{\top}],
\end{align*} 
plus residual terms $\mathcal{O}(\frac{1}{m}\sum_{j=1}^{m}\|\bw_{j}{\scriptstyle (t)}\unaryminus\bw_{a}{\scriptstyle (t)}\|_2^3)$.
Here $\mH^{\mu_j{\scriptscriptstyle (t)}}_{\bw_{a}{\scriptstyle (t)}}\triangleq \frac{1}{|\mu_j{\scriptscriptstyle (t)}|}\sum_{\zeta{\scriptscriptstyle (t)}=1}^{|\mu_j{\scriptscriptstyle (t)}|}\mH({\bw_{a}{\scriptscriptstyle (t)}};z_{j, \zeta{\scriptscriptstyle (t)}})$ denotes the empirical Hessian matrix evaluated at $\bw_{a}{\scriptscriptstyle (t)}$ and $\mT^{\mu_j{\scriptscriptstyle (t)}}_{\bw_{a}{\scriptscriptstyle (t)}} \triangleq \frac{1}{|\mu_j{\scriptscriptstyle (t)}|}\sum_{\zeta{\scriptscriptstyle (t)}=1}^{|\mu_j{\scriptscriptstyle (t)}|}\mT({\bw_{a}{\scriptscriptstyle (t)}};z_{j, \zeta{\scriptscriptstyle (t)}})$ stacks for the empirical third-order partial derivative tensor at $\bw_{a}{\scriptstyle (t)}$, where $\mu_j{\scriptstyle (t)}$ and $z_{j, \zeta{\scriptstyle (t)}}$ follows the notation in \cref{def: c-sgd}.

% Analogous to the works investigating the SGD dynamics \citep{JMLR:v18:17-214, pmlr-v97-zhu19e, ziyin2022strength}, we then calculate the expectation of the gradient diversity.
% The expectation of gradient diversity is calculated first as follows.
As $\bw_{a}{\scriptstyle (t)}$ and local models $\bw_{j}{\scriptstyle (t)}\ (j\unaryequal1,\dots, m)$ are only correlated with the super batch before the $t$-th iteration (see \cref{def: dec-sgd}), taking expectation over $\mu{\scriptscriptstyle (t)}$ provides
\begin{align*}
\ebb_{\mu{\scriptscriptstyle (t)}}&\big[\frac{1}{m}\sum_{j=1}^{m}(\nabla \mL^{\mu_j{\scriptscriptstyle (t)}}_{\bw_{j}{\scriptstyle (t)}}\unaryminus\nabla \mL^{\mu_j{\scriptscriptstyle (t)}}_{\bw_{a}{\scriptstyle (t)}})\big]\nonumber\\
=& \mH_{\bw_{a}{\scriptstyle (t)}} \cdot \underbrace{\frac{1}{m}\sum_{j=1}^{m} (\bw_{j}{\scriptstyle (t)}\unaryminus\bw_{a}{\scriptstyle (t)})}_{= 0}\nonumber\\
&\unaryplus  \frac{1}{2}\mT_{\bw_{a}{\scriptstyle (t)}}\otimes\big[\frac{1}{m}\sum_{j=1}^{m} (\bw_{j}{\scriptstyle (t)}\unaryminus\bw_{a}{\scriptstyle (t)})(\bw_{j}{\scriptstyle (t)}\unaryminus\bw_{a}{\scriptstyle (t)})^{\top}\big],
\end{align*} 
plus residual terms $\mathcal{O}(\frac{1}{m}\sum_{j=1}^{m}\|\bw_{j}{\scriptstyle (t)}\unaryminus\bw_{a}{\scriptstyle (t)}\|_2^3)$, where $\mH_{\bw_{a}{\scriptstyle (t)}}\unaryequal\ebb_{\mu_j{\scriptscriptstyle (t)}} [\mH^{\mu_j{\scriptscriptstyle (t)}}_{\bw_{a}{\scriptstyle (t)}}]$ and $\mT_{\bw_{a}{\scriptstyle (t)}}\unaryequal\ebb_{\mu_j{\scriptscriptstyle (t)}} [\mT^{\mu_j{\scriptscriptstyle (t)}}_{\bw_{a}{\scriptstyle (t)}}]$.

% Therefore, the $i$-th entry of the expected gradient diversity can be further written in the following form:
Then the $i$-th entry of the expected gradient diversity can be written in the following form:
\begin{align}
&\ebb_{\mu{\scriptscriptstyle (t)}}\big[\frac{1}{m}\sum_{j=1}^{m}(\partial_i \mL^{\mu_j{\scriptscriptstyle (t)}}_{\bw_{j}{\scriptstyle (t)}}\unaryminus\partial_i \mL^{\mu_j{\scriptscriptstyle (t)}}_{\bw_{a}{\scriptstyle (t)}})\big]\nonumber\\
\unaryequal& \frac{1}{2}\underbrace{\sum_{l, s} \partial_{i l s }^3 \mL_{\bw_{a}{\scriptstyle (t)}} \frac{1}{m}\sum_{j=1}^{m}{(\bw_{j}{\scriptstyle (t)}\unaryminus\bw_{a}{\scriptstyle (t)})}_{l} {(\bw_{j}{\scriptstyle (t)}\unaryminus\bw_{a}{\scriptstyle (t)})}_s}_{\unaryequal\partial_i \sum_{l s} \partial_{l s}^2 \mL_{\bw} \left(\frac{\sum_{j=1}^{m}}{m}{(\bw_{j}{\scriptscriptstyle (t)}\tinyminus\bw_{a}{\scriptscriptstyle (t)})}_{l} {(\bw_{j}{\scriptscriptstyle (t)}\tinyminus\bw_{a}{\scriptscriptstyle (t)})}_s\right)\big|_{\bw=\bw_{a}{\scriptscriptstyle (t)}}}\nonumber\\
&\unaryplus \mathcal{O}(\frac{1}{m}\sum_{j=1}^{m}\|\bw_{j}{\scriptstyle (t)}\unaryminus\bw_{a}{\scriptstyle (t)}\|_2^3 ),
\label{eq: taylor-expansion}
\end{align} 
where $(\bw_{j}{\scriptstyle (t)}\unaryminus\bw_{a}{\scriptstyle (t)})_l$ denotes the $l$-th entry of the vector $\bw_{j}{\scriptstyle (t)}\unaryminus\bw_{a}{\scriptstyle (t)}$. The equality in the brace is due to Clairaut’s theorem \citep{rudin1976principles}. Details are deferred to \cref{sec:proof-dsgd-sam}. The right hand side (RHS) of this equality without taking value $\bw_{a}{\scriptstyle (t)}$ is the $i$-th partial derivative  of
\begin{align}
    &\Tr(\nabla^2 \mL_{\bw}\bm{\Xi}{\scriptscriptstyle (t)})\nonumber\\
    &=\Tr(\nabla^2 \mL_{\bw}\ebb_{\epsilon \sim \mathcal{N}(0, \bm{\Xi}{\scriptscriptstyle (t)})}[\epsilon\epsilon^T])\nonumber\\
    &=\ebb_{\epsilon \sim \mathcal{N}(0, \bm{\Xi}{\scriptscriptstyle (t)})}[\epsilon^T\nabla^2 \mL_{\bw}\epsilon]\nonumber\\
    &= \underbrace{\ebb_{\epsilon \sim \mathcal{N}(0, \bm{\Xi}{\scriptscriptstyle (t)})}[2(\mL_{\bw+\epsilon}\unaryminus\mL_{\bw})}_{\text{average-direction sharpness at $\bw$}}\unaryplus\mathcal{O}{(\|\epsilon\|_2^3)}].
\label{eq: dsgd-sam}
\end{align}
Proof complete by combining \cref{eq: graident-decomposition} and \cref{eq: dsgd-sam}.
% \begin{align*}
% \ebb_{\mu{\scriptscriptstyle (t)}}&[\bw_{a}{\scriptstyle (t+1)}]\nonumber\\
% =&\bw_{a}{\scriptstyle (t)}-\eta\nabla\underbrace{\ebb_{\epsilon \sim \mathcal{N}(0, \bm{\Xi}{\scriptstyle (t)})}[\mL_{\bw_{a}{\scriptstyle (t)}+\epsilon}]}_{\text{ asymptotic descent direction}}\nonumber\\
% &\unaryplus\underbrace{\mathcal{O}(\eta\ \ebb_{\epsilon \sim \mathcal{N}(0, \bm{\Xi}{\scriptstyle (t)})}\|\epsilon\|_2^3\unaryplus\frac{\eta}{m}\sum_{j=1}^{m}\|\bw_{j}{\scriptstyle (t)}\unaryminus\bw_{a}{\scriptstyle (t)}\|_2^3 )}_{\text{higher-order residual terms}}.
% \end{align*} 
{\color{magenta}\qed}

The proof sketch above outlines the high-level intuition of the flatness-seeking behavior of D-SGD.

\begin{tcolorbox}[notitle, rounded corners, colframe=darkgrey, colback=lightblue, 
       boxrule=2pt, boxsep=0pt, left=0.15cm, right=0.17cm, enhanced, 
       shadow={4.5pt}{-4.5pt}{0pt}{opacity=5,mygrey},
       toprule=2pt, before skip=0.65em, after skip=0.75em ]
\emph{
    % {\centering  D-SGD and average-direction SAM are asymptotically equivalent.\\}
    {\textbf{High-level intuition}: Model decentralization introduces gradient diversity among local models (see \cref{eq: graident-decomposition}), which induces a unique Hessian-consensus dependent noise. This noise directs the optimization trajectory of D-SGD towards regions with lower average-direction sharpness $\ebb_{\epsilon \sim \mathcal{N}(0, \bm{\Xi}{\scriptstyle (t)})}[\mL_{\bw+\epsilon}\unaryminus\mL_{\bw}]$.}
}
\end{tcolorbox}

\subsection{Gradient smoothing effect of decentralization}
Previous literature has shown the gradient stabilization effect of isotropic Gaussian noise injection \citep{bisla2022low,liu2022random}.
According to \cref{th: dsgd-sam}, decentralization can be interpreted as the injection of Gaussian noise into gradient.
There arises a natural question that whether or not the noise introduced from decentralization, which is not necessarily isotropic, would smooth the gradient.
We employ the following Corollary to answer this question.

\begin{corollary}[Gradient smoothing effect\label{coro: smoothing}]
     Given that the vanilla loss function $\mL_{\bw}$ is $\alpha$-Lipschitz continuous and the gradient $\nabla\mL_{\bw}$ is $\beta$-Lipschitz continuous. We can conclude that the gradient $\nabla\mL_{\bw+\epsilon}$ where $\epsilon\sim\mathcal{N}(0,\bm{\Xi}{\scriptstyle (t)})$ is $\min \left\{\frac{\sqrt{2}\alpha}{\sigma_{\textnormal{min}}}, \beta\right\}$-Lipschitz continuous, where $\sigma_{\textnormal{min}}=\lambda_{\textnormal{min}}(\bm{\Xi}{\scriptstyle (t)})$, the smallest eigenvalue of $\bm{\Xi}{\scriptstyle (t)}$.
\end{corollary}
\cref{coro: smoothing} implies that if the lower bound of noise magnitude satisfies  $\sigma_{\textnormal{min}}\geq\frac{\sqrt{2}\alpha}{\beta}$, then the noise $\epsilon$ can make the Lipschitz constant of gradients smaller, therefore leading to gradient smoothing. 
% We note that the gradient smoothing effect is stronger in the initial training phase when $\sigma_{\textnormal{min}}$ stays large, which is consistent with the observation that the training curves of D-SGD are more stable than C-SGD in large-batch settings.
The gradient smoothing effect exhibited by D-SGD aligns with two empirical observations in large-batch settings: (1) the training curves of D-SGD are notably more stable than those of C-SGD, and (2) D-SGD can converge with larger learning rates, as reported in \citep{zhang2021loss}.
The proof is deferred to \cref{sec: smoothing}. Further research directions include dynamical stability analysis \citep{kim2023stability, wu2023} of D-SGD.

\begin{figure*}[t!]
\begin{subfigure}[ResNet-18 on CIFAR-10 (C-SGD), 128 total batch size]{.33\textwidth}
  \centering
  % include fourth image
  \includegraphics[width=1.0\linewidth]{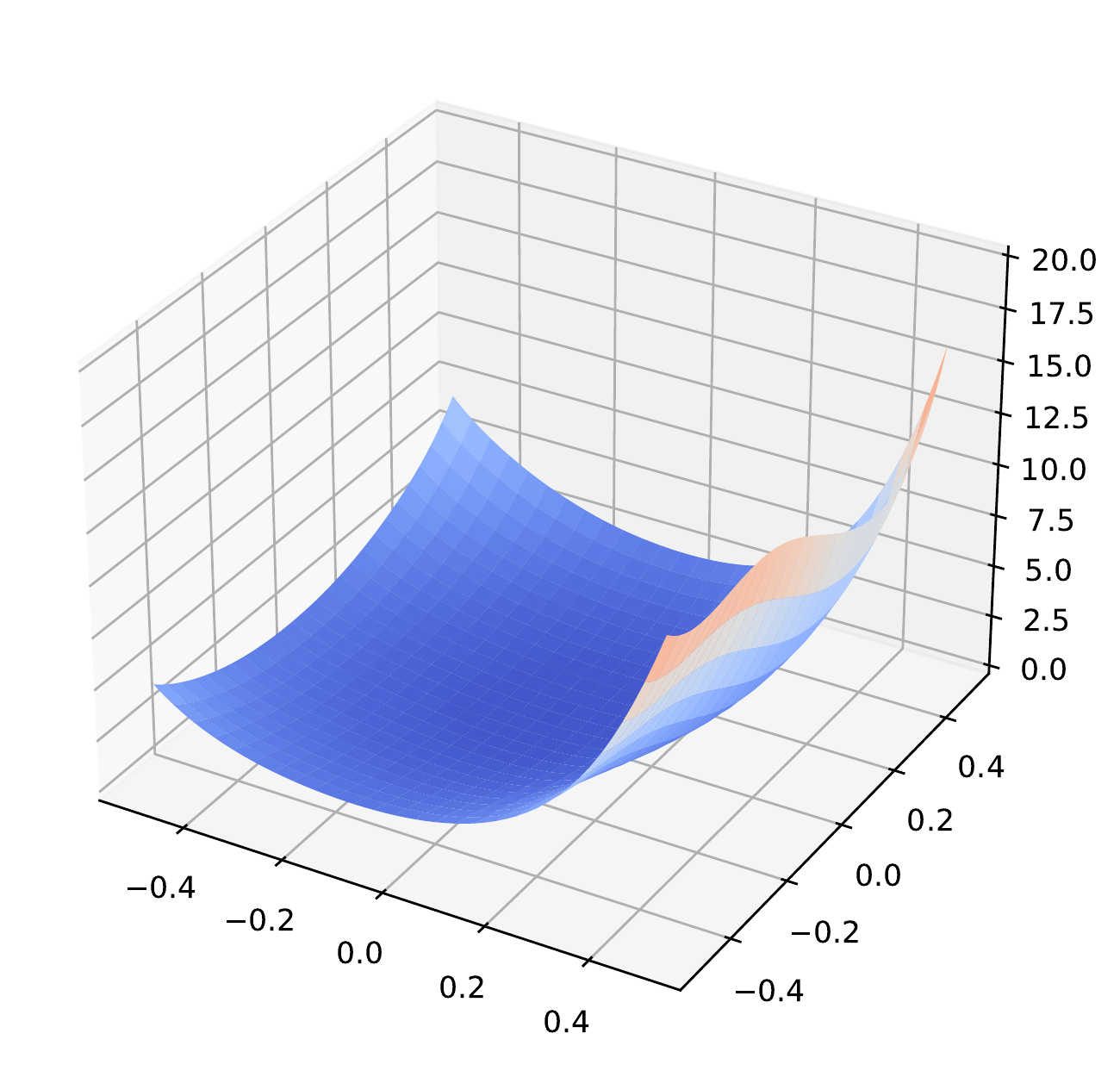}  
  \caption{C-SGD, 128 total batch size}
\end{subfigure}
\begin{subfigure}[ResNet-18 on CIFAR-10 (C-SGD), 1024 total batch size]{.33\textwidth}
  \centering
  % include fourth image
  \includegraphics[width=1.0\linewidth]{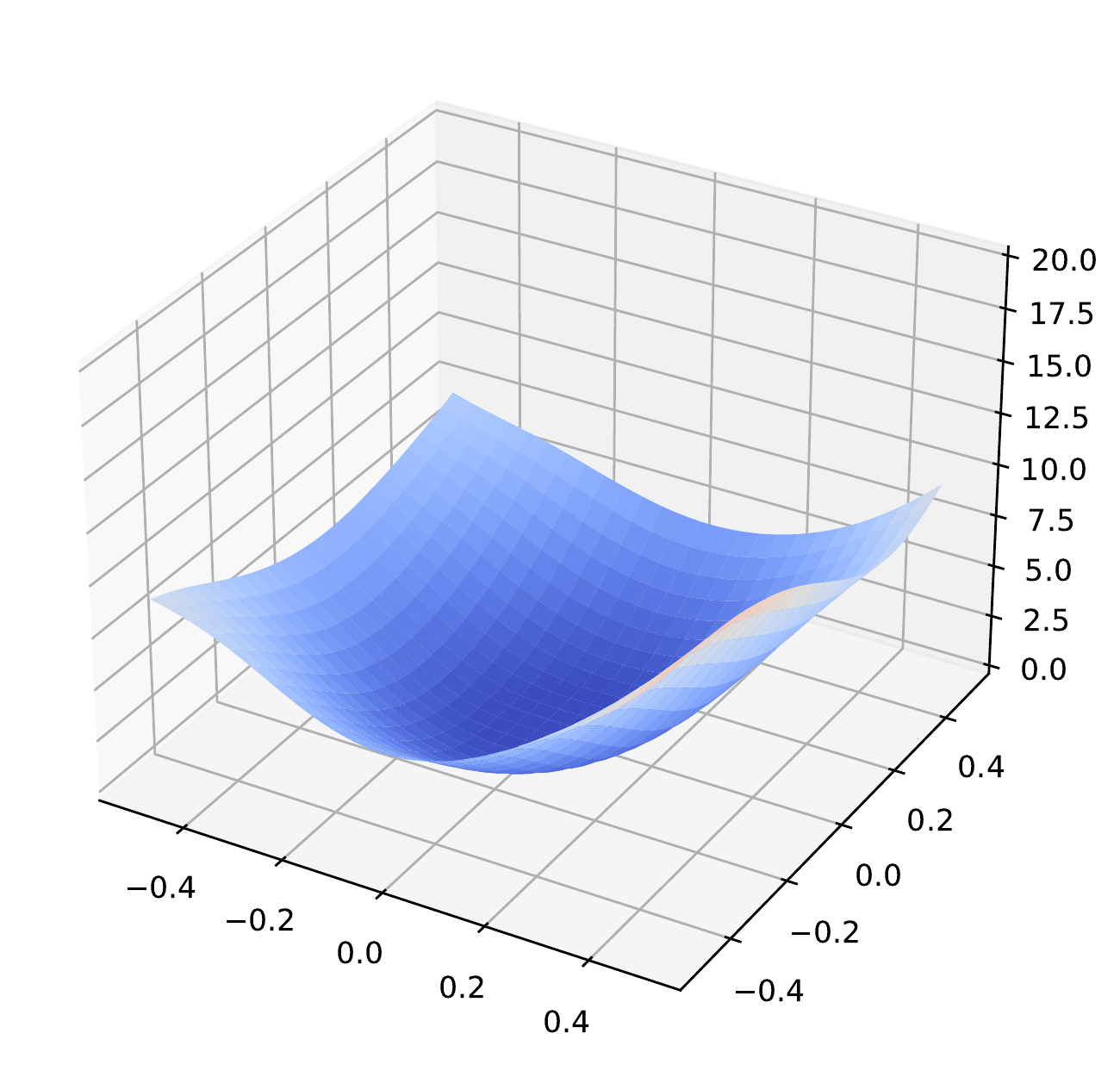}  
  \caption{C-SGD, 1024 total batch size}
\end{subfigure}
\begin{subfigure}[ResNet-18 on CIFAR-10 (C-SGD), 8196 total batch size]{.33\textwidth}
  \centering
  % include fourth image
  \includegraphics[width=1.0\linewidth]{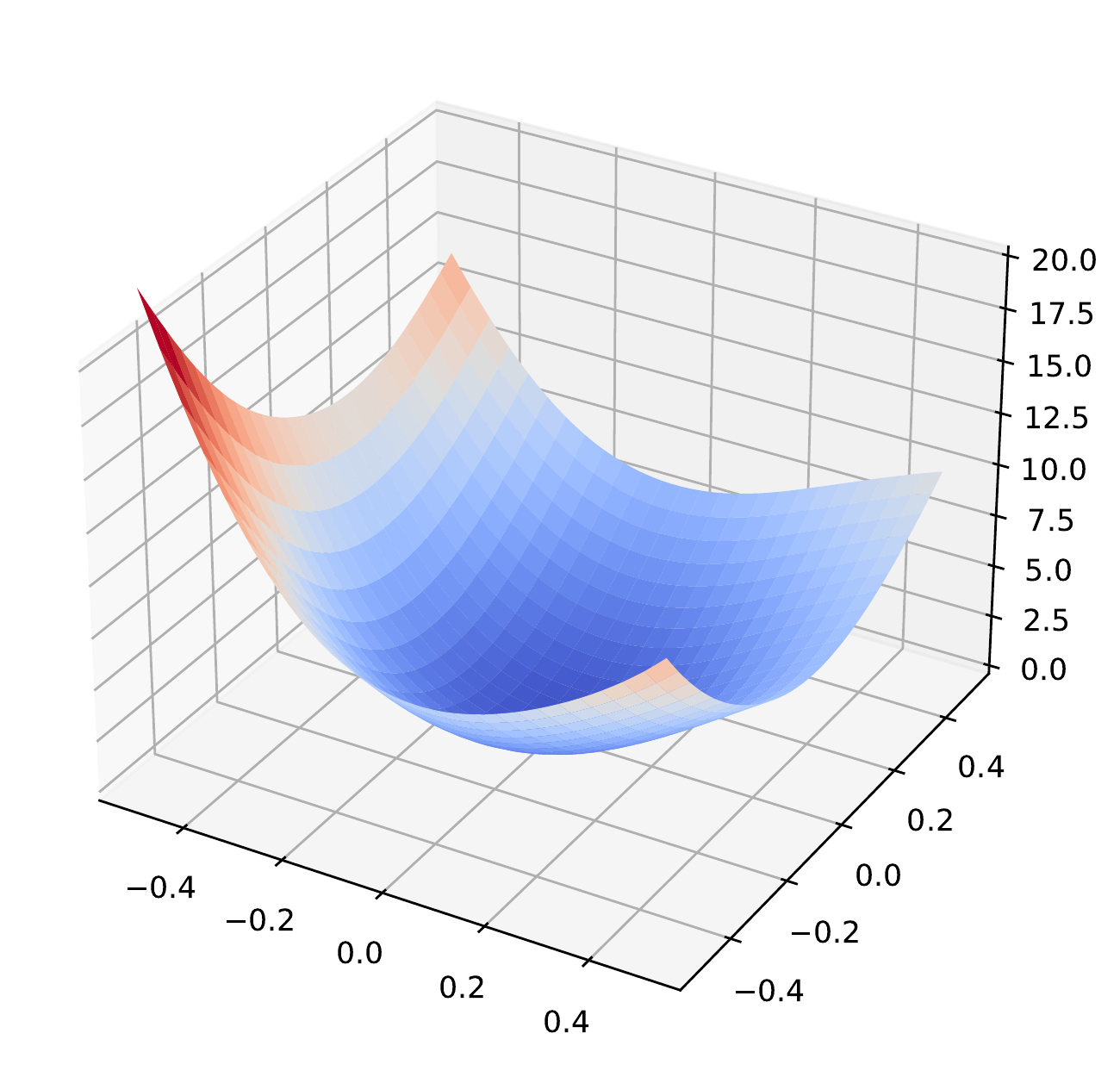}  
  \caption{C-SGD, 8192 total batch size}
\end{subfigure}

\medskip

\begin{subfigure}[ResNet-18 on CIFAR-10 (D-SGD), 128 total batch size]{.33\textwidth}
  \centering
  % include fourth image
  \includegraphics[width=1.0\linewidth]{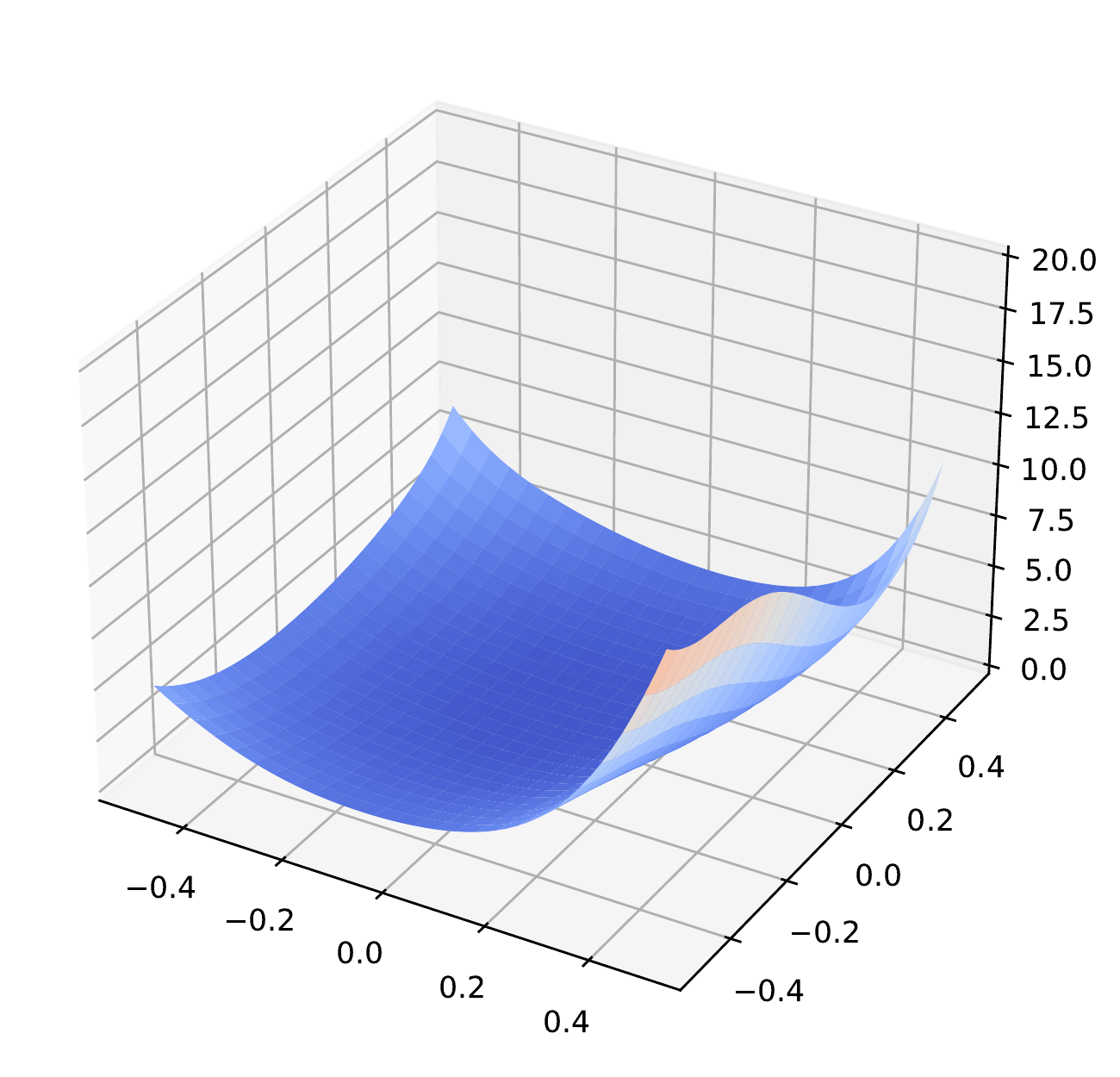}  
  \caption{D-SGD, 128 total batch size}
\end{subfigure}
\begin{subfigure}[ResNet-18 on CIFAR-10 (D-SGD), 1024 total batch size]{.33\textwidth}
  \centering
  % include fourth image
  \includegraphics[width=1.0\linewidth]{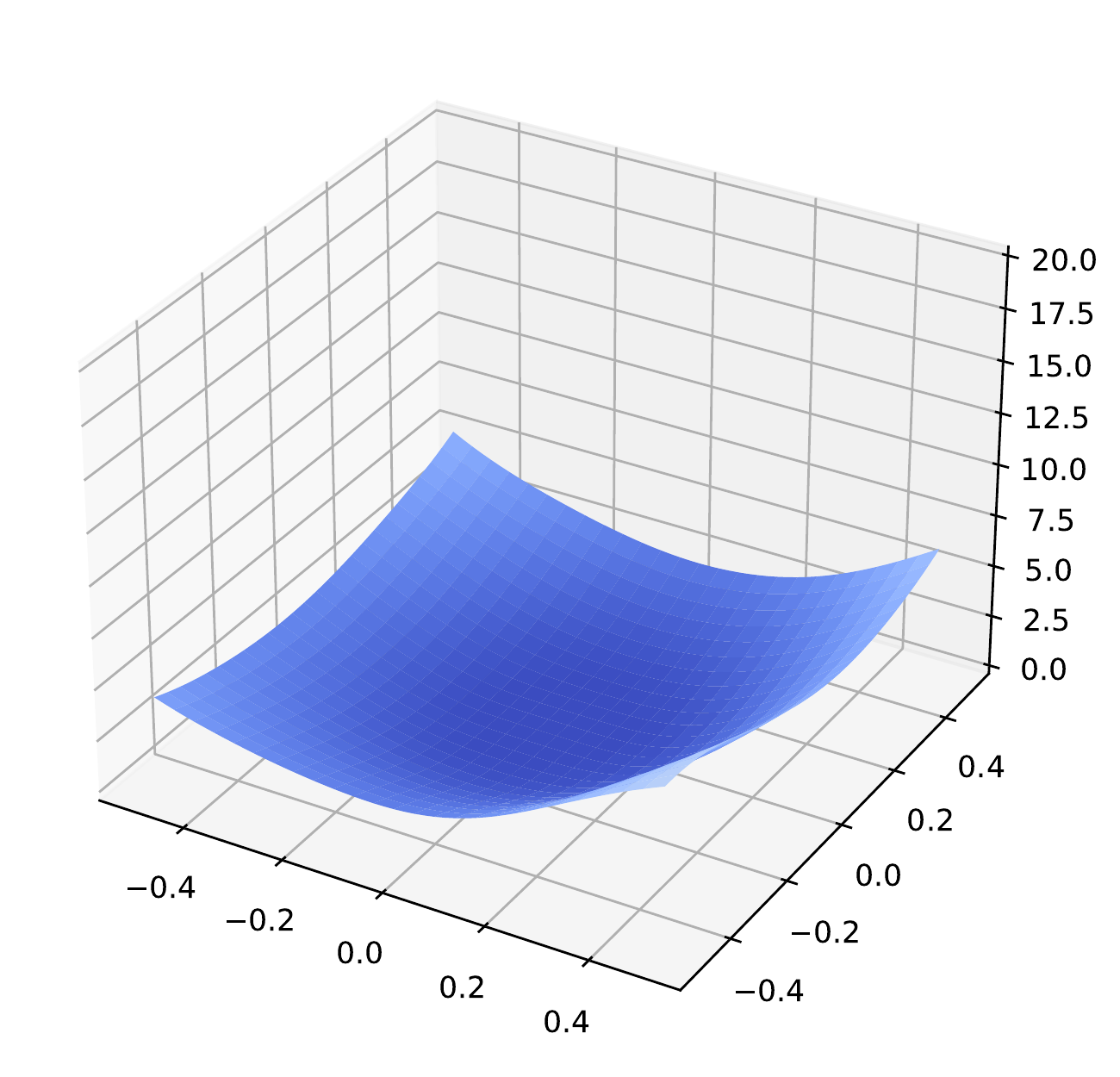}  
  \caption{D-SGD, 1024 total batch size}
\end{subfigure}
\begin{subfigure}[ResNet-18 on CIFAR-10 (D-SGD), 8196 total batch size]{.33\textwidth}
  \centering
  % include fourth image
  \includegraphics[width=1.0\linewidth]{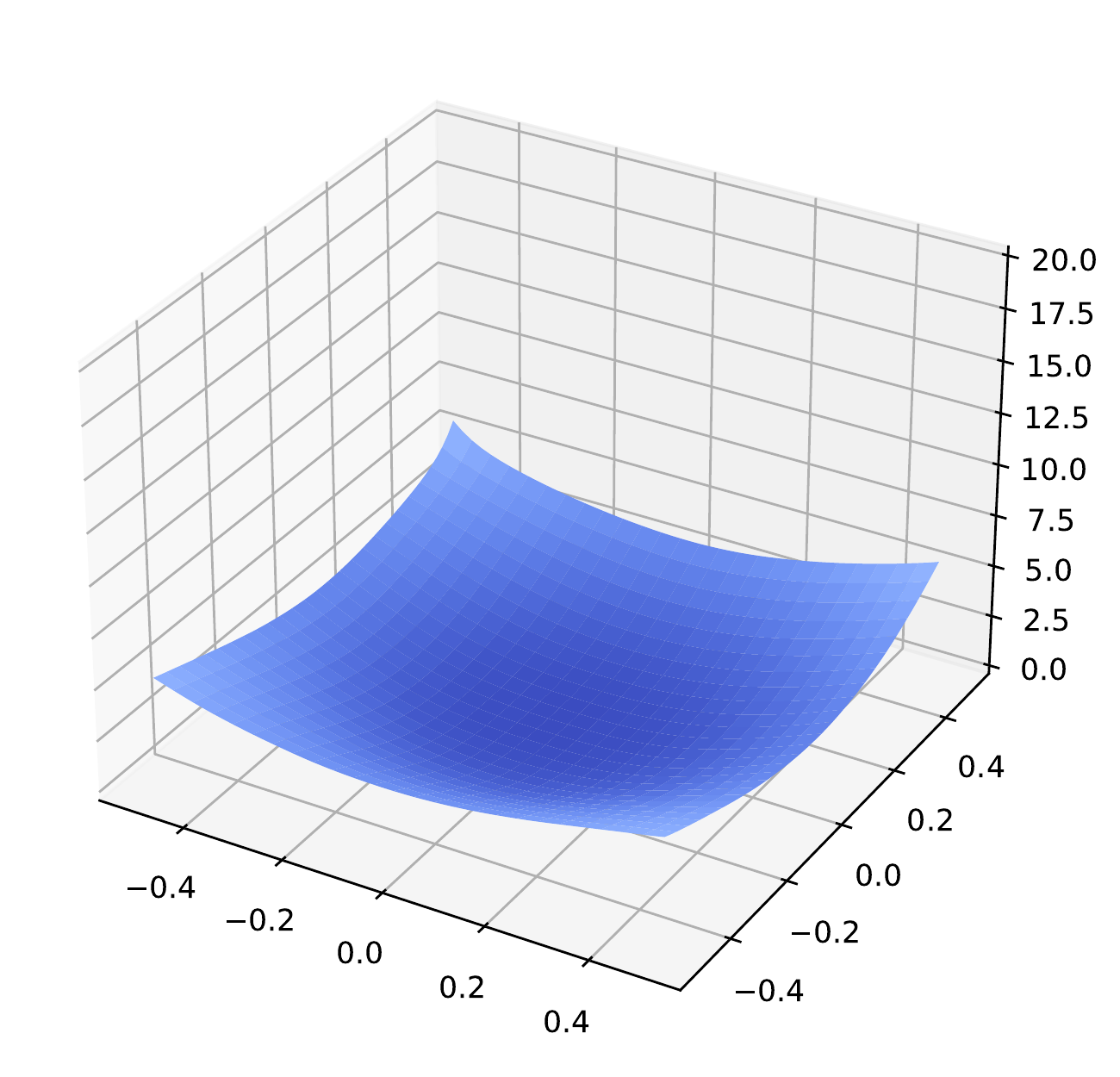}  
  \caption{D-SGD, 8192 total batch size}
\end{subfigure}
\caption{Minima 3D visualization of ResNet-18 trained on CIFAR-10 using C-SGD and D-SGD (ring topology).}
\label{fig: 3d-minima-cifar10}
\end{figure*}

\subsection{Generalization benefit of decentralization: the role of total batch size}
\label{subsec: generalization-large-batch}

In practice, data and model decentralization ordinarily imply large total batch sizes, as a massive number of workers are involved in the system in many practical scenarios. Large-batch training can enhance the utilization of super computing facilities and further speed up the entire training process. Thus, studying the large-batch setting is crucial for fully understanding the application of D-SGD. 

Despite the importance, theoretical understanding of the generalization of large-batchdecentralized training remains an open problem.\footnote{Please refer to \cref{subsec: generalization-in-large-batch} for the detailed discussion on the generalization of large-batch training.} 
This subsection examines the implicit regularization of D-SGD with respect to (w.r.t.) total batch size $|\mu|$, and compares it to C-SGD.
To investigate the impact of $|\mu|$, we analyze the effect of the gradient variance, in addition to the gradient expectation studied in \cref{sec: dsgd-sam}. 

\begin{theorem}\label{th: large-batch}
Let $B=|\mu|$ denote the total batch size. With a probability greater than $1-\mathcal{O}(\frac{B}{(N-B)\eta^2})$, D-SGD implicit minimizes the following objective function:\footnote{If we apply the linear scaling rule (LSR) when increasing the total batch size $B$ (i.e., $\frac{\eta}{B}=s$ where $s$ is a constant), the probability becomes $1-\mathcal{O}{(\frac{1}{(N-B)Bs^2})}\approx 1$ and thus is not  uninformative.}
\begin{align*}
    &\mL_{\bw}^{\text{\tiny D-SGD}}
    \unaryequal\mL_{\bw}^\mu \unaryplus\underbrace{\Tr(\mH_\bw^\mu \bm{\Xi}{\scriptstyle (t)})\unaryplus\frac{\eta}{4}\Tr({(\mH^\mu_\bw)}^2 \bm{\Xi}{\scriptstyle (t)})}_{\text{batch size independent sharpness regularizer}}
     \nonumber\\
    &\unaryplus\underbrace{\tikzmarknode{kappa}{\hlmath{LightBlue}{\kappa}}\cdot \frac{1}{N}\sum_{j=1}^{N}\left[\|\nabla\mL_{\bw}^j\unaryminus \nabla\mL_{\bw}^\mu\|_2^2 \unaryplus
    \Tr({(\mH_\bw^j\unaryminus\mH_\bw^\mu)}^2\bm{\Xi}{\scriptstyle (t)})\right]}_{\text{batch size dependent variance regularizer}}\nonumber\\
    &\unaryplus \frac{\eta}{4}\|\nabla\mL_{\bw}^\mu\|_2^2\unaryplus\mathcal{R}^{A}\unaryplus\mathcal{O}{(\eta^2)},
\end{align*}
where $\kappa=\frac{\eta}{B}\cdot\frac{N\unaryminus B}{(N\unaryminus1)}$, and $\mathcal{R}^{A}$ absorbs all higher-order residuals. The empirical gradient $ \nabla\mL^{\mu}_{\bw}$ on the super-batch $\mu$ are averaged over the one-sample gradients $\nabla \mL_{\bw}^j\ (j=1,\dots,m)$. Similarly, the empirical Hessian $\mH_\bw^\mu$ is an average of $\mH_{\bw}^j=\mH(\bw;z_j)\ (j=1,\dots,m)$.
\end{theorem}

 A corresponding implicit regularization of C-SGD (and SGD) is established in \cref{lemma: implicit-reg-sgd}, which demonstrates that C-SGD has an implicit gradient variance reduction mechanism to improve generalization (see \cref{fig: gradient-variance}). However, as the total batch size $B$ approaches the total training sample size $N$, the regularization term diminishes rapidly, even in the case when the learning rate scales with the total batch size, since the ratio $\kappa=\frac{N-B}{N-1}$ converges to $0$ gradually.

On the contrary, \cref{th: large-batch} proves that the sharpness regularization terms in D-SGD do not decrease as the total batch size increases, unlike in C-SGD, which theoretically justifies the potential superior generalizability of D-SGD in large-batch settings. 
The underlying intuition is that decentralization introduces additional noise, which compensates for the noise required for D-SGD to generalize well in large-batch scenarios.
The proof is included in \cref{sec:proof-large-batch}.

%% file: section/5-empirical_results.tex
\section{Empirial results}
\label{sec: empirical-results}

This section empirically validates our theory. 
We introduce the experimental setup and then study how decentralization impacts minima flatness via local landscape visualization.

\textbf{Dataset and architecture.} Decentralized learning is simulated in a dataset-centric setup by uniformly partitioning data among multiple workers (GPUs) to accelerate the training process. Vanilla D-SGD with various commonly used topologies (see \cref{fig: centralized-decentralized}) and C-SGD are employed to train image classifiers on CIFAR-10 \citep{krizhevsky2009learning} and Tiny ImageNet \citep{le2015tiny} with AlexNet \citep{krizhevsky2017imagenet}, ResNet-18 \citep{he2016identity} and DenseNet-121 \citep{Huang_2017_CVPR}. Detailed implementation setting is inclued in \cref{sec: additional-empirical-result}.\footnote{In our experiments, the ImageNet pre-trained  models are used as initializations to achieve better final validation performance. The conclusions still stand for training from scratch.

On CIFAR-10, we use deterministic topology. On Tiny ImageNet, we use deterministic topology with random neighbor shuffling, which can increase the effective spectral gap of the underlying communication matrix \citep{Zhang2020ImprovingEI}. We adjust the effective spectral gap by introducing random shuffle to accommodate the ``optimal temperature" of models on different datasets.}

As demonstrated in \cref{fig: test-accuracy}, \cref{fig: test-accuracy_2} and \cref{fig: test-accuracy_3}, D-SGD could outperform C-SGD in terms of generalizability in large-batch settings.\footnote{This is due to the fact that the training accuracy of D-SGD is almost surely $100\%$ in all settings, making validation accuracy a reliable measure of generalizability.

Note that it is not a rigorous claim that decentralization will always improve generalization in large-batch settings. The experiments reveal the  generalization potential of decentralized learning.} 
We also note that the gap in generalizability between D-SGD and C-SGD in a large-batch scenario is larger on the CIFAR-10 dataset, which we attribute to the smaller $\kappa$ value (see \cref{th: large-batch}).
To further support our claim that D-SGD favors flatter minima than C-SGD in large-batch scenarios, we plotted the minima learned by both algorithms using the loss landscape visualization tool in \citet{li2018visualizing}. The resulting plots are shown in \cref{fig: 3d-minima-cifar10}, along with additional plots in \cref{sec: additional-empirical-result}.

These figures demonstrate that D-SGD could learn flatter minima than C-SGD in large-batch settings, and this difference in flatness becomes larger as the total batch size increases.
These observations are consistent with the claims made in \cref{th: dsgd-sam} and \cref{th: large-batch} that D-SGD favors flatter minima than its centralized counterpart, especially in the large-batch scenarios.
Future work includes visualizing the whole trajectories of D-SGD.

%% file: section/7-conclusion.tex
\section{Discussion and Broader Impact}\label{sec: discussion}

\textbf{Question 1: How to scale learning rate w.r.t. batch size and spectral gap in decentralized deep learning?}

As proved in \cref{th: dsgd-sam}, the training dynamics of D-SGD and C-SGD are completely different. There also exists potential landscape smoothing and batch size-independent regularization effect of D-SGD (\cref{coro: smoothing} and \cref{th: large-batch}). Consequently, we conjecture that D-SGD could be more ``tolerable" to hyperparameters such as learning rate than C-SGD. However, the existing tricks for hyperparameter tuning are tailored for C-SGD (e.g., linear scaling rule). A natural question is that can we design a general scaling strategy of learning rate as a function of batch size and the spectral gap of the communication topology, in order to maintain generalizability of D-SGD in large-batch settings?

\textbf{Question 2: Would continuously reducing spectral gap  improve the validation performance of D-SGD?}

The answer is  negative, as D-SGD inherently exhibits a regularization-optimization trade-off.
On the one hand, a sparse topology with a too-small spectral gap leads to large consensus distance (see proof in \cref{prop: consensus-distance}), which also slows down the converges of the higher-order residual terms, and consequently, hampers the optimization of the original objective.\footnote{A sparse topology refers to the topology whose neighbor number is relatively smaller than the total number of workers. A sparse topology always has a small spectral gap (see \cref{def:spectral-gap})}
On the other hand, although a small spectral gap increases the sharpness regularization effect, it also makes the optimization of $\ebb_{\epsilon \sim \mathcal{N}(0, \bm{\Xi}{\scriptstyle (t)})}[\nabla\mL_{\bw_{a}{\scriptstyle (t)}+\epsilon}]$ difficult.
In SAM, a too large perturbation radius (or neighborhood size) $\rho$ incurs divergence \citep{pmlr-v162-andriushchenko22a,mueller2022perturbing}. Similarly, a topology with a small spectral gap leads to a large $\Tr{(\bm{\Xi{\scriptscriptstyle (t)}})}$, the magnitude of the variance of $\epsilon$. A large $\Tr{(\bm{\Xi{\scriptscriptstyle (t)}})}$ results in an increased search space for the optimization problem $\min_{\bw}\ebb_{\epsilon \sim \mathcal{N}(0, \bm{\Xi}{\scriptstyle (t)})}[\mL_{\bw+\epsilon}-\mL_{\bw}]$, making it more difficult to solve. These intuitions align with the observation that the D-SGD converges slowly on sparse topologies \citep{NIPS2017_f7552665,pmlr-v119-koloskova20a}.
Based on the analysis, we conjecture that there exists ``\textbf{sweet spots}'', which balance the sharpness regularization introduced by proper decentralization and the optimization issues originated from sparse topologies. Finding these ``sweet spots'' then becomes a promising direction in a decentralized deep learning.

% \textbf{Future work 1: Analyze the dynamics of the local (and personalized) models of D-SGD.}

% This work primarily focuses on the properties of the averaged model, while the dynamics of local models remain unexplored. A promising but challenging direction is analyzing the dynamics of the local (and personalized) models.

\textbf{Future work 1: Fine-grained convergence and generalization analyses of decentralized learning algorithms.}

We have demonstrated several potential benefits of D-SGD based on the established equivalence.
Another direct extension of our work is to utilize the connection between decentralized learning and centralized learning to improve the existing convergence and generalization bounds of decentralized learning algorithms.

\textbf{Future work 2: Provide adversarial robustness guarantees for 
decentralized learning algorithms.} 

\citet{cao2023decentralized} find that decentralized stochastic gradient algorithms are more adversarial robust than their centralized counterpart in certain scenarios. They boldly conjecture that the superiority could be attributed to their inclination towards flatter minimizers. The question is can we provide rigorous adversarial robustness guarantees of decentralized learning algorithms, and develop more robust algorithms?

\textbf{Future work 3: Bridge decentralized learning and SAM.} 

An interesting question arising from the asymptotic equivalence is that does D-SGD share the properties of SAM, beyond generalizablity, including better interpretability \citep{andriushchenko2023sharpness} and transferability \citep{chen2022when}? Furthermore, it is worth exploring whether the insights gained from SAM \citep{pmlr-v162-andriushchenko22a, möllenhoff2023sam, howsharpness2023} can be utilized to design more effective decentralized algorithms.

\section{Conclusion}

This paper challenges the conventional belief that centralized learning is optimal in generalizablity and establishes a surprising asymptotic equivalence of decentralized SGD and average-direction SAM. This asymptotic equivalence further demonstrates a regularization-optimization trade-off and three advantages of D-SGD: (1) there exists a surprising free uncertainty estimation mechanism in D-SGD to estimate the intractable posterior covariance; (2) D-SGD has a gradient smoothing effect; and (3) the sharpness regularization effect of D-SGD does not decrease as total batch size increases, which justifies the superior generalizablity of D-SGD over centralized SGD (C-SGD) in large-batch settings. 
Although our theory focuses primarily on the vanilla decentralized SGD, we believe our theoretical insights are applicable to a broad range of decentralized gradient-based algorithms. 
Code is available at \href{https://github.com/Raiden-Zhu/ICML-2023-DSGD-and-SAM}{\faGithub~D-SGD and SAM}.

%% file: section/acknowledgement
%\newpage
\section*{Acknowledgement}

This work is supported by National Natural Science Foundation of China (U20B2066, 61976186), the Zhejiang Provincial Key Research and Development Program of China (2021C01164), and the Fundamental Research Funds for the Central Universities (2021FZZX001-23, 226-2023-00048).
Prof Dacheng Tao is partially supported by Australian Research Council Project FL-170100117.

The authors appreciate Xinran Gu for the valuable discussions on Local SGD and the limiting dynamics of D-SGD.
The authors also thank Batiste Le Bars for the constructive comments on the asymptotic equivalence and Theorem 4.

%% file: section/appendix.tex
\onecolumn
\appendix
\numberwithin{equation}{section}
\numberwithin{theorem}{section}
\numberwithin{remark}{section}
\numberwithin{definition}{section}
\numberwithin{assumption}{section}
\numberwithin{figure}{section}
\numberwithin{table}{section}
\renewcommand{\thesection}{{\Alph{section}}}
\renewcommand{\thesubsection}{\Alph{section}.\arabic{subsection}}
\renewcommand{\thesubsubsection}{\Alph{section}.\arabic{subsection}.\arabic{subsubsection}}

\newpage
\section{Background\label{sec:background}}

\subsection{Distributed centralized training (data decentralization)\label{subsec: centralized-learning}}

Efficiently training large-scale models on massive amounts of data is challenging yet important for real-world applications \citep{narayanan2021efficient, shen2023efficient}.
To handle an increasing amount of data and parameters, distributed training on multiple workers emerges. Traditional distributed training systems usually follow a centralized setup (parameter server).

\textbf{Distributed centralized stochastic gradient  descent (C-SGD).}
In C-SGD \citep{dean2012large, li2014communication}, the de facto distributed training algorithm, there is only one centralized model $\bw_{a}{\scriptstyle (t)}$.
C-SGD updates the model by
\begin{align}\label{eq: c-sgd}
    \bw_{a}{\scriptstyle (t+1)} 
    = \bw_{a}{\scriptstyle (t)}-\eta\cdot\frac{1}{m}\sum_{j=1}^{m}\overbrace{  \nabla\mL^{\mu_j{\scriptstyle (t)}}(\bw_{a}{\scriptstyle (t)})}^{\text {Local gradient computation}},
\end{align}
where $\eta$ is the learning rate, $\mu_j{\scriptstyle (t)}=\{z_{j,1}, \ldots, z_{j,|\mu_j{\scriptstyle (t)}|}\}$ denotes the local training batch independent and identically distributed (i.i.d.) drawn from the data distribution $\mathcal{D}$ at the $t$-th iteration, and $\nabla \mL^{\mu_j{\scriptstyle (t)}}_{\bw}=\nabla \mL^{\mu_j{\scriptstyle (t)}}(\bw) = \frac{1}{|\mu_j{\scriptstyle (t)}|}\sum_{\zeta{\scriptstyle (t)}=1}^{|\mu_j{\scriptstyle (t)}|}\nabla\mL({\bw};z_{j, \zeta{\scriptstyle (t)}})$ stacks for the local mini-batch gradient of $\mL$ w.r.t. the first argument $\bw$. The total batch size of C-SGD at $t$-th iteration is $|\mu{\scriptstyle (t)}| = \sum_{j=1}^{m}|\mu_j{\scriptstyle (t)}|$.

``Centralized" refers to the fact that in C-SGD, there is a central server receiving weight or gradient information from local workers (see \cref{fig: centralized-decentralized}). 
To clarify, the process of updating the global model with the globally averaged gradients, as shown in equation \ref{eq: c-sgd}, is equivalent to averaging the local weights after performing local gradient updates on the global model:
\begin{align}\label{eq: c-sgd2}
    &\bw_{j}{\scriptstyle (t+1)} 
    = \bw_{j}{\scriptstyle (t)}-\eta\cdot  \nabla\mL^{\mu_j{\scriptstyle (t)}}(\bw_{a}{\scriptstyle (t)}), \quad \bw_{a}{\scriptstyle (t+1)}=\frac{1}{m}\sum_{j=1}^{m}\bw_{j}{\scriptstyle (t+1)},\nonumber\\
    \Rightarrow &\bw_{a}{\scriptstyle (t+1)} 
    = \bw_{a}{\scriptstyle (t)}-\eta\cdot\frac{1}{m}\sum_{j=1}^{m}\overbrace{  \nabla\mL^{\mu_j{\scriptstyle (t)}}(\bw_{a}{\scriptstyle (t)})}^{\text {Local gradient computation}}.
\end{align}
C-SGD defined in \cref{eq: c-sgd} and \cref{eq: c-sgd2} are mathematically identical to the Federated Averaging algorithm \citep{pmlr-v54-mcmahan17a} under the condition that the local step is set as $1$ and all local workers are selected by the server in each round. To avoid misunderstandings, we include the term ``distributed" in C-SGD to differentiate it from traditional single-worker SGD \citep{cauchy1847methode, Robbins1951ASA}.

\subsection{Decentralized training (data and model decentralization)\label{subsec: decentralized-learning}}

 \textbf{Limitations of server-based training.} Despite convenience and scalability, central server-based training scheme suffers from two main issues: (1) a centralized communication protocol slows down training since central servers are easily overloaded, especially in low-bandwidth or high-latency cases \citep{NIPS2017_f7552665}; (2) there exists a potential information leakage through privacy attacks on the gradients transmitted to central server despite decentralizing data using Federated Learning \citep{NEURIPS2019_60a6c400,NEURIPS2020_c4ede56b,Yin_2021_CVPR,wang2023reconstructing}. As an alternative, decentralized learning (machine learning in a peer-to-peer architecture) allows workers to balance the load on the central server \citep{NIPS2017_f7552665}, as well as  maintain confidentiality \citep{warnat2021swarm}. 

\textbf{Development of decentralized algorithms.} The earliest work of classical decentralized optimization can be  traced  back  to \citet{tsitsiklis1984problems}, \citet{tsitsiklis1986distributed} and \citet{nedic2009distributed}. Decentralized SGD, a direct combination of decentralization and gradient-based optimization, has been extended to various contexts, including time-varying topologies \citep{nedic2014distributed,lu2020decentralized,pmlr-v119-koloskova20a,ying2021exponential}, directed topologies \citep{assran2019stochastic,taheri2020quantized,song2022communicationefficient}, asynchronous settings \citep{lian2018asynchronous,xu2021dp,nadiradze2021asynchronous,bornstein2023swift}, personalized settings \citep{li2022learning, shi2023towards}, data-heterogeneous scenarios \citep{tang2018d,vogels2021relaysum,pmlr-v206-le-bars23a,pmlr-v202-shi23d} and Byzantine-robust versions \citep{9084329,Farhadkhani2023}.
Although our theory focuses primarily on the vanilla decentralized SGD, we anticipate our theoretical insights will be applicable to a broad range of decentralized algorithms.

We then summarize some commonly used notions regarding decentralized training in the following.
\begin{definition}[Doubly Stochastic Matrix\label{def:dou-matrix}]
Let $\mathcal{G}=(\mathcal{V},\mathcal{E})$ stand for the decentralized communication topology, where $\mathcal{V}$ denotes the set of $m$ computational nodes and $\mathcal{E}$ represents the edge set. For any given topology $\mathcal{G}=(\mathcal{V},\mathcal{E})$,
the doubly stochastic gossip matrix ${\bp} = [\bp_{j,k}] \in \mathbb{R}^{m\times m}$ is defined on the edge set $\mathcal{E}$ that satisfies
\begin{itemize}[leftmargin=*]
    \item ${\bp} = {\bp}^{\top}$ (symmetric);
    \item If $j\neq k$ and $(j,k) \notin {\cal E}$, then $\bp_{j,k} =0$ (disconnected) and otherwise, $\bp_{j,k} >0$ (connected);
    \item $\bp_{j,k}\in [0,1]\ \ \forall k, l$ and $\sum_k \bp_{j,k}=\sum_l \bp_{j,k}=1$ (standard weight matrix for undirected graph).
\end{itemize}
\end{definition}

The doubly stochasticity of the gossip matrices is a standard assumption in decentralized learning \citep{NIPS2017_f7552665, pmlr-v119-koloskova20a}.
It is worth noticing that our theory is generally applicable to \textbf{arbitrary communication topologies} whose gossip matrices are doubly stochastic. 

In the following we illustrate some commonly-used communication topologies.

\begin{figure*}[h!]
  \centering
  % include fourth image
  \includegraphics[width=0.7\linewidth]{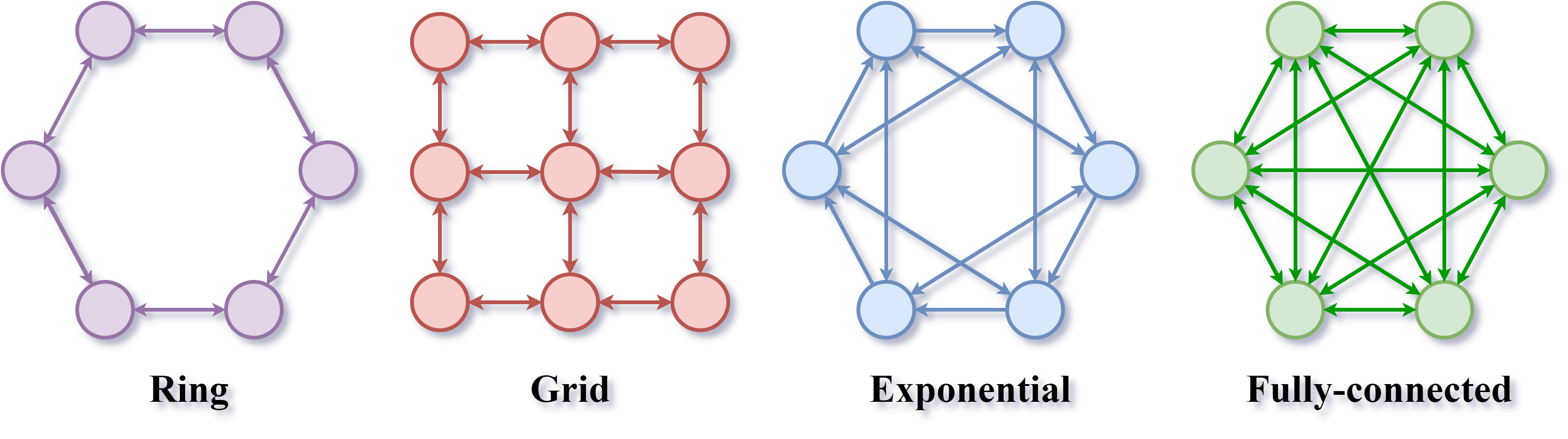}  
  \caption{An illustration of some commonly used topologies.}
\label{fig: topologies}
\end{figure*}

The intensity of gossip communications is measured by the spectral gap \citep{seneta2006non} of $\bp$. 

\begin{definition}[Spectral Gap\label{def:spectral-gap}]
Denote $\lambda=\max \left\{\left|\lambda_{2}\right|,\left|\lambda_{m}\right|\right\}$ where $\lambda_i\ (i=2,\dots,m)$ is the $i$-th largest eigenvalue of gossip matrix ${\bp}\in\rbb^{m\times m}$.
  The spectral gap of a gossip matrix $\bp$ can be defined as follows:
  \begin{equation*}
  {\textnormal{spectral gap}} := 1- \lambda.
  \end{equation*}
\end{definition}
According to the definition of doubly stochastic matrix (\cref{def:dou-matrix}), we have $0\leq\lambda<1$. The spectral gap measures the connectivity of the communication topology. A topology is considered sparse if its communication matrix has a small spectral gap close to $0$, while a topology is considered dense if its communication matrix has a large spectral gap close to $1$.

% \textbf{Influence of communication topologies.}

\subsection{Sharpness-aware minimization\label{subsec: intro-SAM}}

Sharpness-aware minimization (SAM) is proposed by \citet{foret2021sharpnessaware} to minimize a perturbed loss function for the purpose of improving generalization, which is studied concurrently by \citet{wu2020adversarial} and \citet{zheng2021regularizing}. Subsequently, various SAM variants emerge, including adaptive SAM \citep{kwon2021asam}, surrogate gap guided SAM \citep{zhuang2022surrogate}, LookSAM \citep{liu2022towards}, Fisher SAM \citep{pmlr-v162-kim22f}, random SAM \citep{liu2022random}, sparse SAM \citep{mi2022make}, variational SAM \citep{ujvary2022rethinking} and Bayes SAM \citep{möllenhoff2023sam}.
\begin{definition}[Vanilla SAM \citep{foret2021sharpnessaware}\label{def: SAM}]
The loss function of vanilla SAM is defined as follows:
\begin{align*}
\mL^{\text{\tiny SAM}}(\bw)=\max_{\|\in\|_p \leq \rho}\mL(\bw+\epsilon).
\end{align*}
\citet{foret2021sharpnessaware} propose to use a first-order approximation to simplify the max step:
\begin{align*}
\mL^{\text{\tiny SAM}}(\bw)\approx\max _{\|\in\|_p \leq \rho}[\mL(\bw)+\epsilon^{\top}\nabla\mL(\bw)],
\end{align*}
where $\epsilon^*=\rho\frac{\nabla\mL(\bw)}{\|\nabla\mL(\bw)\|_2}$ is the close-form solution. 

Therefore, the gradient update of vanilla SAM becomes
\begin{align*}
\nabla\mL^{\text{\tiny SAM}}(\bw)\approx\nabla\mL(\bw+\epsilon^*)=\nabla\mL(\bw+\rho\frac{\nabla\mL(\bw)}{\|\nabla\mL(\bw)\|_2}).
\end{align*}
\end{definition}

\begin{definition}[Average-direction SAM \citep{howsharpness2023}\label{def:AD-SAM}]
Average-direction SAM (AD-SAM) minimizes 
\begin{align*}
\mL^{\text{\tiny AD-SAM}}(\bw)=\ebb_{\epsilon \sim \mathcal{N}(0, I)}[\mL({\bw+\epsilon^{\top}\frac{\nabla\mL(\bw)}{\|\nabla\mL(\bw)\|_2}})].
\end{align*}
\end{definition}
Average-direction is named because it minimizes an averaged loss in a ``basin” around $\bw$, rather than the original point-loss.
Actually, the generalization bound of vanilla SAM in \citet{foret2021sharpnessaware} upper bounds the generalization error by the average-direction sharpness $\mL^{\text{\tiny AD-SAM}}(\bw)-\mL(\bw)$. However, \citet{howsharpness2023} proves that average-direction SAM actually minimizes $\mL(\bw)+\Tr({\mH(\bw)})$ rather than $\max_{\|\in\|_p \leq \rho}\mL(\bw+\epsilon)$.

The average-direction SAM that defined in our paper is sightly different from that in \citep{howsharpness2023}:
\begin{align*}
    {\underbrace{{\tikzmarknode{sharpness-aware_regularizer}{\ebb_{\epsilon \sim \mathcal{N}(0, \bm{\Xi}{\scriptstyle (t)})}[\mL_{\bw+\epsilon}-\mL_{\bw}]}}}_{\text{\textit{average-direction sharpness}}}}.
\end{align*}
This kind of AD-SAM depends both on the local landscape and consensus, which is discussed in detail in \cref{sec: dsgd-sam}.

\subsection{Generalization in large-batch training\label{subsec: generalization-in-large-batch}}
Large-batch training is of significant interest for deep learning deployment, which can contribute to a significant speed-up in training neural networks \citep{goyal2017accurate, 10.1145/3225058.3225069, JMLR:v20:18-789}. Unfortunately, it is widely observed that in the centralized learning setting, large-batch training often suffers from a drastic generalization degradation, even with fine-tuned hyperparameters, from both empirical \citep{chen2016scalable, keskar2017on,hoffer2017train,JMLR:v20:18-789,smith2020generalization,smith2021on} and theoretical \citep{he2019control, li2021validity} aspects. An explanation of this phenomenon is that large-batch training lacks sufficient gradient noise to escape ``sharp" minima \citep{smith2020generalization}. 

\textbf{Mitigating generalization issues in large-batch training.}
Linear scaling rule (LSR) is a widely used hyper-parameter-free rule to make up the noise in large-batch deep learning \citep{he2016deep, goyal2017accurate, bottou2018optimization, smith2020generalization}, which states that a fixed learning rate to total batch size ratio allows maintaining generalization performance when the total batch size increases. Apart from LSR, various optimization techniques have been proposed to reduce the gap, including learning rate warmup \citep{smith2017cyclical}, Layer-wise Adaptive Rate Scaling (LARS) \citep{you2017large} and Layer-wise Adaptive Moments (LAMB)  \citep{You2020Large}. Is worth noticing that decentralization could be a general training technique in a data-centric setup, which can be combined with these approaches to further improve generalization in large-batch training.

\subsection{$\beta$-smoothness\label{subsec: smoothness}}

Smoothness is a fundamental property of the objective function in optimization, which characterizes the behavior of its gradient with respect to changes in the parameters \citep{boyd2004convex}. In particular, $\beta$-smoothness quantifies the rate at which the objective function varies with respect to changes in the input variables, which is demonstrated as follow: 
\begin{definition}[$\beta$-smoothness\label{def:l-smooth}]
$\mL$ is $\beta$-smooth if for any $z$ and $\bw, \widetilde{\bw}\in\rbb^d$,
  \begin{equation}
    \big\|\nabla \mL(\bw;z)-\nabla \mL(\widetilde{\bw};z)\big\|_2\leq \beta\|\bw-\widetilde{\bw}\|_2.
  \end{equation}
\end{definition}
An objective function that lacks $\beta$-smoothness (i.e., non-$\beta$-smoothness) can exhibit sudden variations in its gradient with respect to the input variables, making it difficult to predict the behavior of the function and to design efficient optimization algorithms.
However,  $\beta$-smoothness is generally difficult to ensure at the beginning and intermediate phases of deep neural network training \citep{NEURIPS2020_2e2c4bf7}. It is noteworthy that our theoretical framework does not rely on the $\beta$-smoothness assumption of the objective function $\mL$, which renders it suited for different stages of deep neural network training.

\subsection{Explanation of tensor product\label{subsec: high-dimensional-taylor-expansion}}
% High dimensional Taylor expansion
The tensor product between a third-order tensor $\mT\in\rbb^{d\times d\times d}$ and a second-order tensor (matrix)  $\mM\in\rbb^{d\times d}$ in the proof of \cref{th: dsgd-sam} is defined as 
\begin{align*}
    \underbrace{{(\mT\otimes\mM)}_i}_{\text {the $i$-th entry}} = \text{grandsum}{(\mT_i\odot\mM)},
\end{align*}
where $\mT_i\in\rbb^{d\times d}$ is a second-order tensor (matrix), $\odot$ denotes the Hadamard product \citep{davis1962norm}, and the $\text{grandsum}(\cdot)$ \citep{merikoski1984trace} of a second-order tensor (matrix) $\tilde{\mM}$ satisfies $\text{grandsum}(\tilde{\mM}) = \sum_{i,j} \tilde{\mM}_{ij}$.

\newpage
\section{Experimental Setup and Additional Results\label{sec: additional-empirical-result}}

This section provides a comprehensive account of the experimental setup, along with supplementary experiments comparing the generalization performance and the sharpness of the minima of C-SGD and D-SGD. The validation accuracy comparison of C-SGD and D-SGD includes three scenarios: on multiple communication topologies, without pretraining, and using layer-wise learning rate tuning.

\textbf{Dataset and architecture.} Decentralized learning is simulated in a dataset-centric setup by uniformly partitioning data among multiple workers (GPUs) to accelerate training. Vanilla D-SGD with various commonly used topologies (see \cref{fig: centralized-decentralized}) and C-SGD are employed to train image classifiers on CIFAR-10 \citep{krizhevsky2009learning} and Tiny ImageNet \citep{le2015tiny} with AlexNet \citep{krizhevsky2017imagenet}, ResNet-18 \citep{he2016identity} and DenseNet-121 \citep{Huang_2017_CVPR}. The ImageNet pretrained  models are used as initializations to achieve better final validation performance.

\begin{figure*}[h!]
\centering
\begin{subfigure}[ResNet-18 on CIFAR-10 (C-SGD), 128 total batch size]{.33\textwidth}
  \centering
  % include fourth image
  \includegraphics[width=1.0\linewidth]{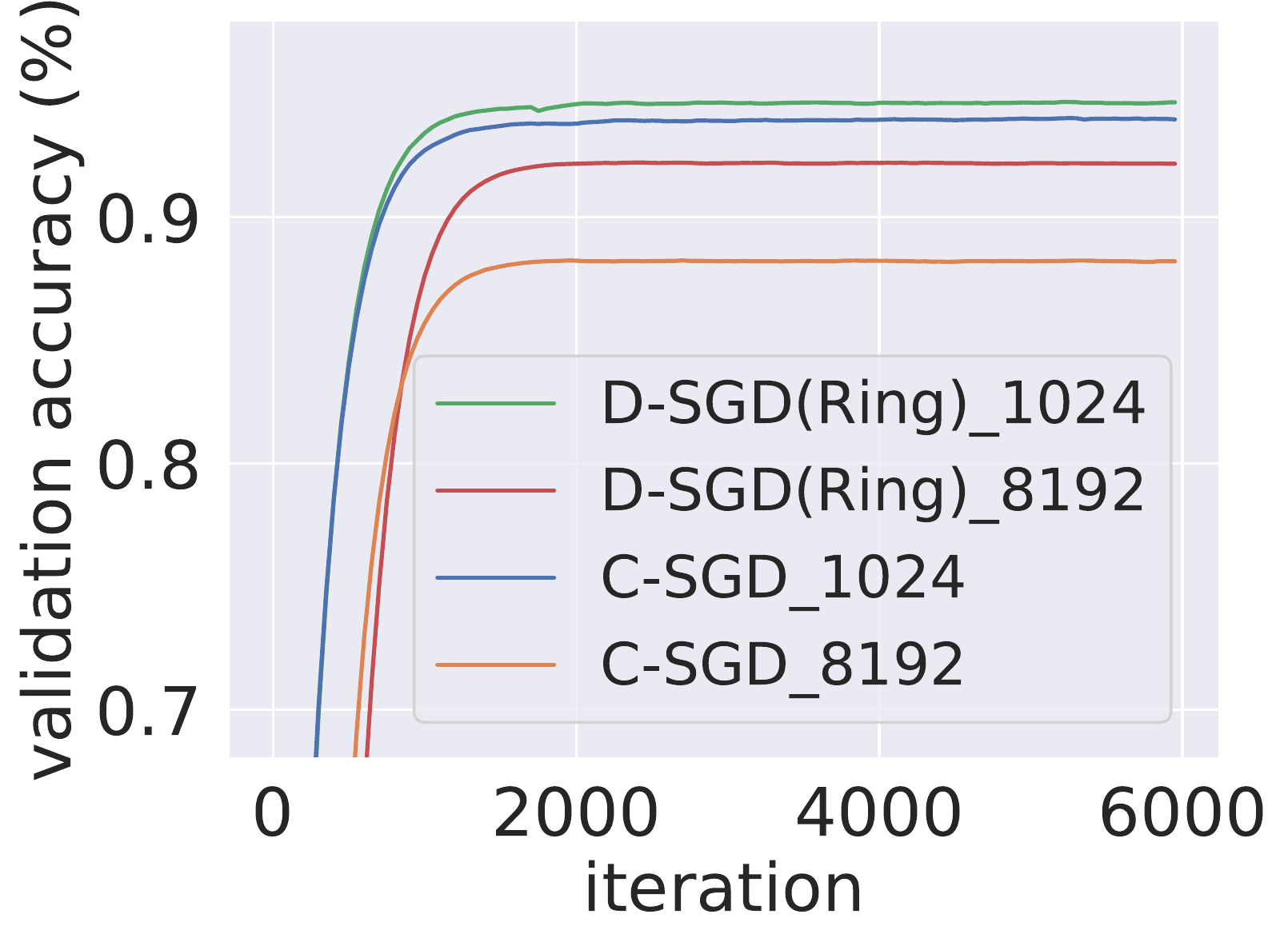}  
\end{subfigure}
\begin{subfigure}[ResNet-18 on CIFAR-10 (C-SGD), 1024 total batch size]{.33\textwidth}
  \centering
  % include fourth image
  \includegraphics[width=1.0\linewidth]{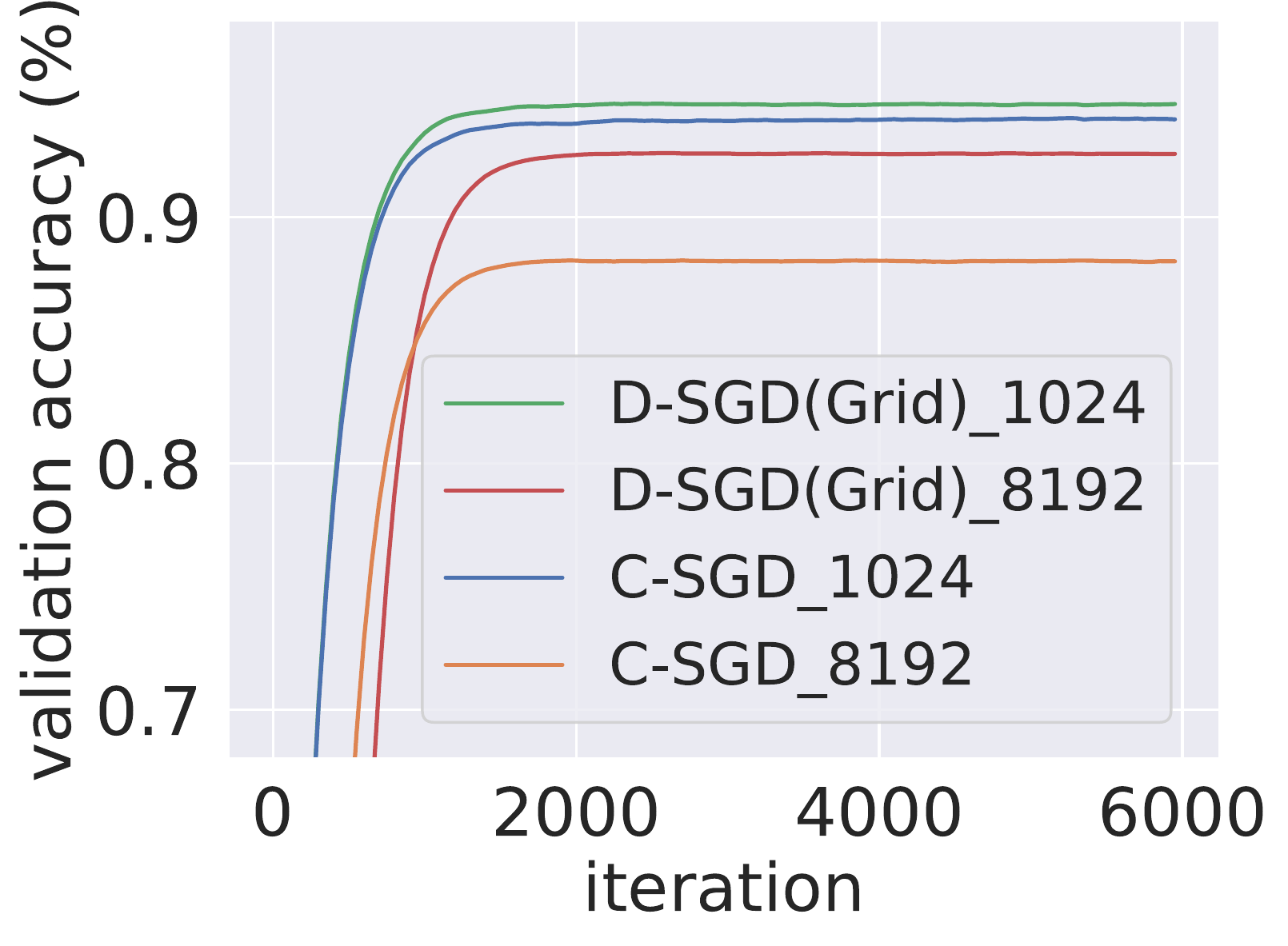}  
\end{subfigure}
\begin{subfigure}[ResNet-18 on CIFAR-10 (C-SGD), 8196 total batch size]{.33\textwidth}
  \centering
  % include fourth image
  \includegraphics[width=1.0\linewidth]{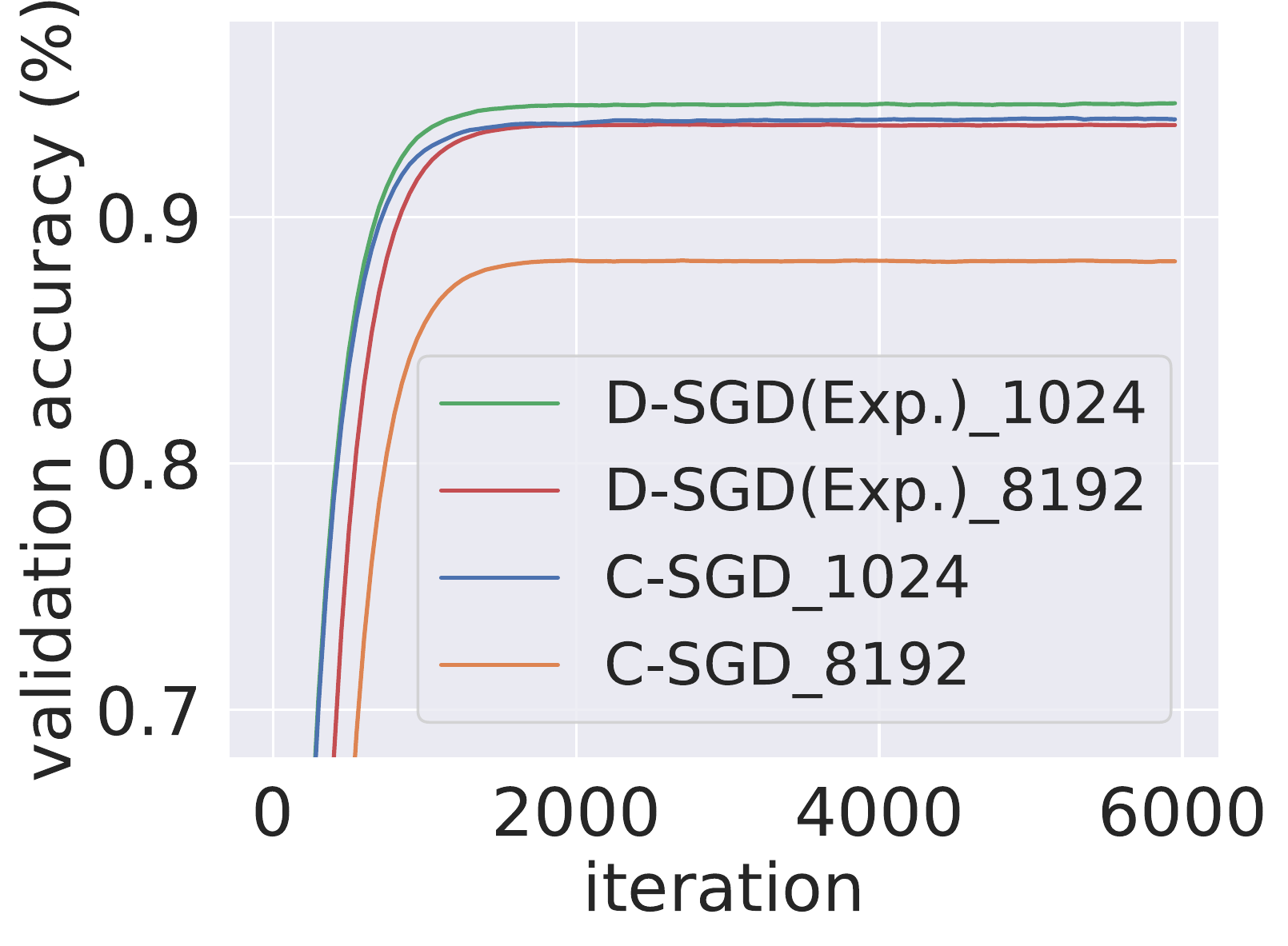}
\end{subfigure}
\caption{The validation accuracy comparison of ResNet-18 trained on \underline{CIFAR-10} using C-SGD and D-SGD with various topologies (see \cref{fig: centralized-decentralized}). The number of workers (one GPU as a worker) is set as 16; and the local batch size is set as 64, and 512 per worker (1024 and 8196 total batch size).}
\label{fig: test-accuracy_2}
\end{figure*}

\begin{figure*}[h!]
\centering
\begin{subfigure}[ResNet-18 on Tiny ImageNet (C-SGD), 128 total batch size]{.33\textwidth}
  \centering
  % include fourth image
  \includegraphics[width=1.0\linewidth]{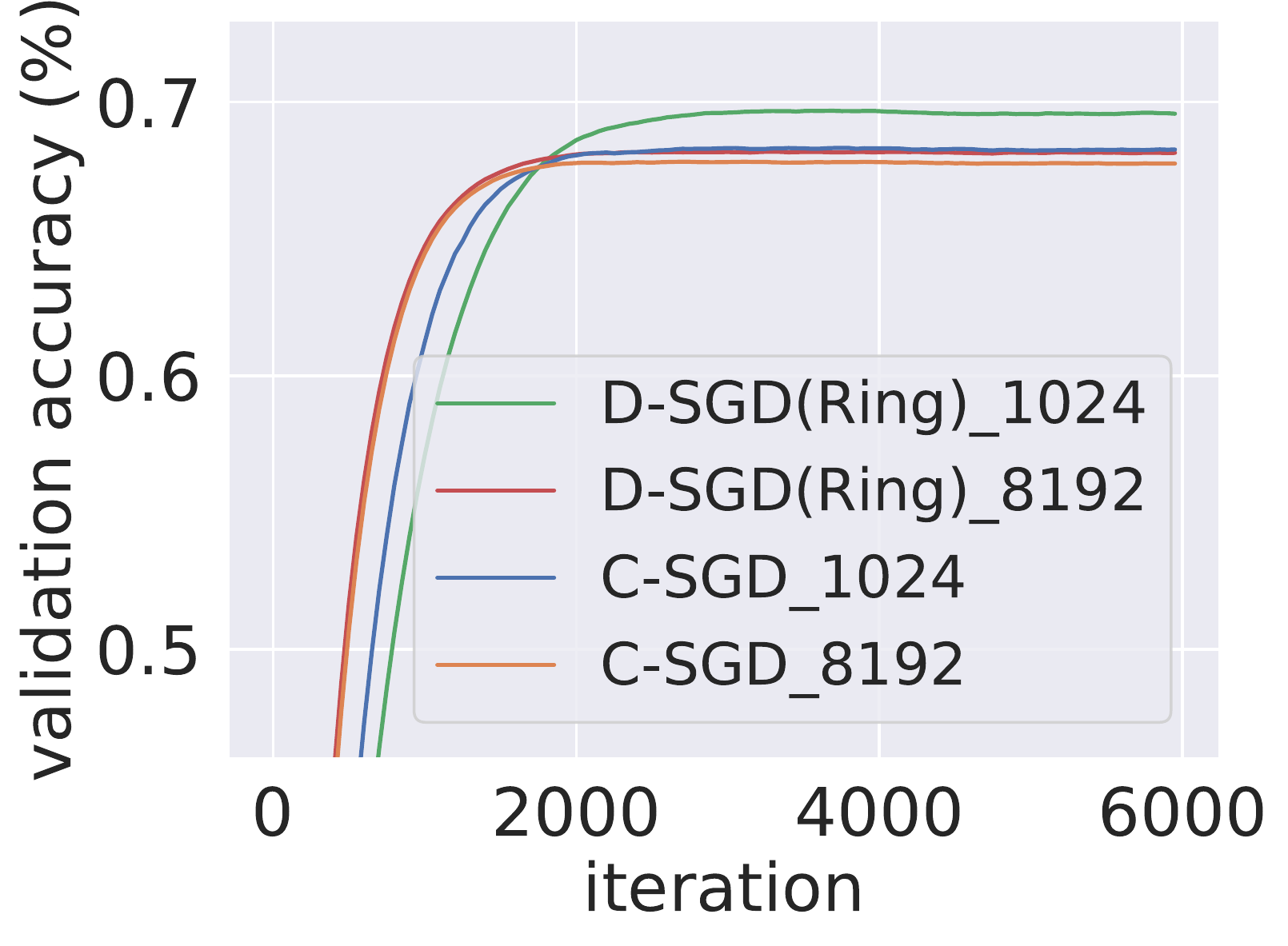}  
\end{subfigure}
\begin{subfigure}[ResNet-18 on Tiny ImageNet (C-SGD), 1024 total batch size]{.33\textwidth}
  \centering
  % include fourth image
  \includegraphics[width=1.0\linewidth]{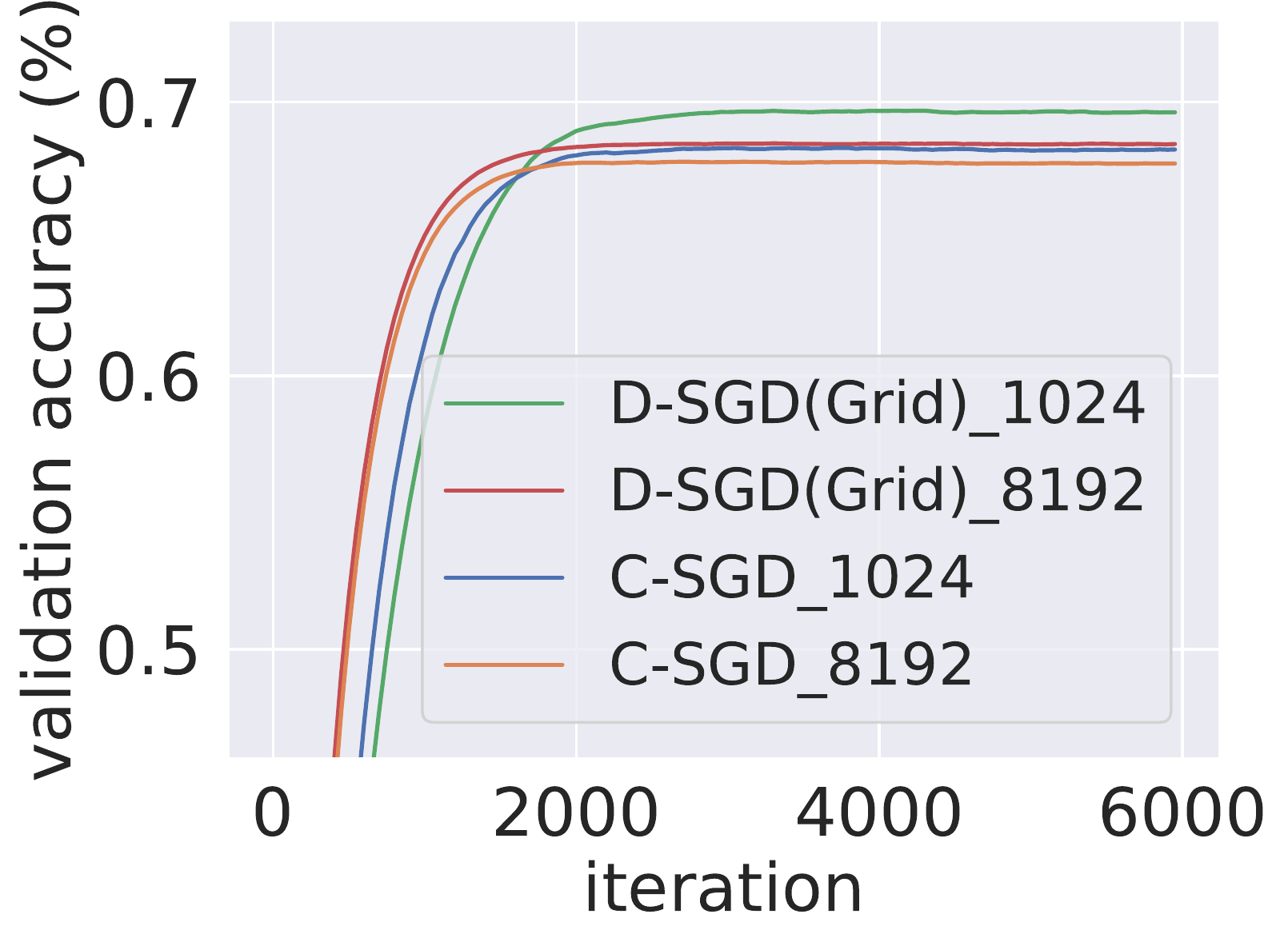}  
\end{subfigure}
\begin{subfigure}[ResNet-18 on Tiny ImageNet (C-SGD), 8196 total batch size]{.33\textwidth}
  \centering
  % include fourth image
  \includegraphics[width=1.0\linewidth]{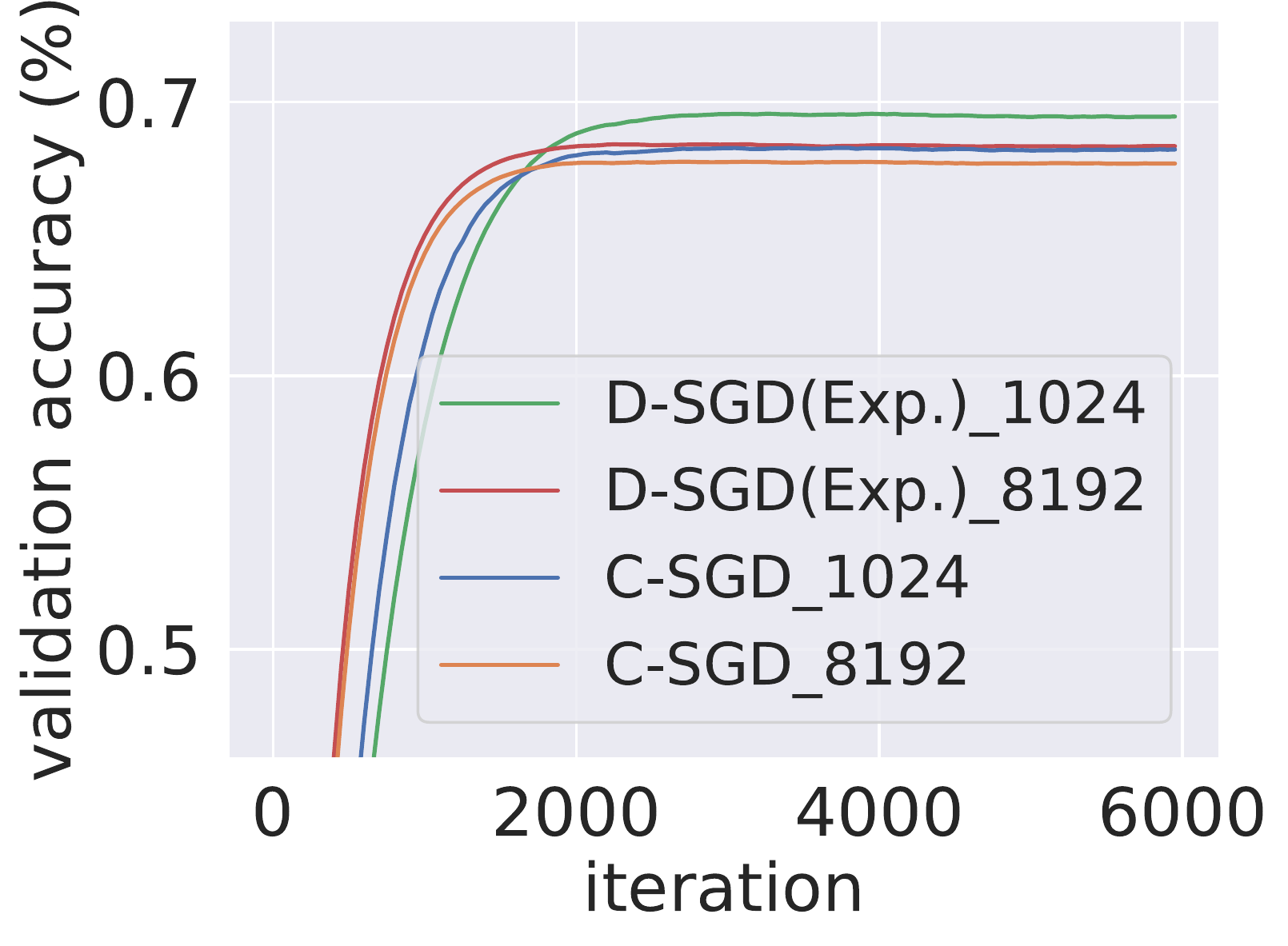}
\end{subfigure}
\caption{The validation accuracy comparison of ResNet-18 trained on \underline{Tiny ImageNet} using C-SGD and D-SGD with various topologies (see \cref{fig: centralized-decentralized}). The number of workers (one GPU as a worker) is set as 16; and the local batch size is set as 64, and 512 per worker.}
\label{fig: test-accuracy_3}
\end{figure*}

\textbf{Implementation setting.}  The number of workers (one GPU as a worker) is set as 16; and the local batch size is set as 8, 64, and 512 per worker in different cases.
For the case of local batch size 64, the initial learning rate is set as $0.1$ for ResNet-18 and ResNet-34 and $0.01$ for AlexNet, following the setup in \citep{zhang2021loss}. The learning rate is divided by $10$ when the model has passed the $2/5$ and $4/5$ of the total number of iterations \citep{he2016deep}.
We apply the learning rate warm-up \citep{smith2017cyclical} and the linear scaling law \citep{he2016deep, goyal2017accurate} to maintain generalization performance with increased total batch size. Batch normalization \citep{ioffe2015batch} is employed in training AlexNet.
In order to understand the effect of decentralization on the flatness of minima and generalization, all other training techniques are strictly controlled. The training accuracy is almost $100\%$ everywhere. Exponential moving average is employed to smooth the validation accuracy curves in \cref{fig: test-accuracy}, \cref{fig: test-accuracy_2} and \cref{fig: test-accuracy_3}. The code is based on PyTorch \citep{paszke2019pytorch}.

% including dropout \citep{JMLR:v15:srivastava14a}, data augmentation \citep{lecun1998gradient}, momentum \citep{qian1999momentum} and weight decay \citep{tihonov1963solution}

\textbf{Hardware environment.} The experiments are conducted on a computing facility with NVIDIA\textsuperscript{\textregistered} Tesla\textsuperscript{\texttrademark} V100 16GB GPUs and Intel\textsuperscript{\textregistered} Xeon\textsuperscript{\textregistered} Gold 6140 CPU @ 2.30GHz CPUs.

\cref{fig: test-accuracy_2} and \cref{fig: test-accuracy_3} show that D-SGD with different topologies could outperform C-SGD by a large margin in large-batch settings, which are consistent with the results shown in \cref{fig: test-accuracy}.

\textbf{Additional experiments without pretraining.}
Additional experiments are conducted to further investigate the impact of pretraining. All other training settings are kept the same as the previous experiments.
\begin{figure*}[h!]
\centering
\begin{subfigure}[ResNet-18 on CIFAR-10 (C-SGD), 1024 total batch size]{.4125\textwidth}
  \centering
  % include fourth image
  \includegraphics[width=0.8\linewidth]{section/Figures/Accuracy/230330_fig_cifar10_resnet18.pdf}  
  \caption{ResNet-18 (with ImageNet pretrained weights)}
\end{subfigure}
\begin{subfigure}[ResNet-18 on CIFAR-10 (C-SGD), 128 total batch size]{.4125\textwidth}
  \centering
  % include fourth image
  \includegraphics[width=0.8\linewidth]{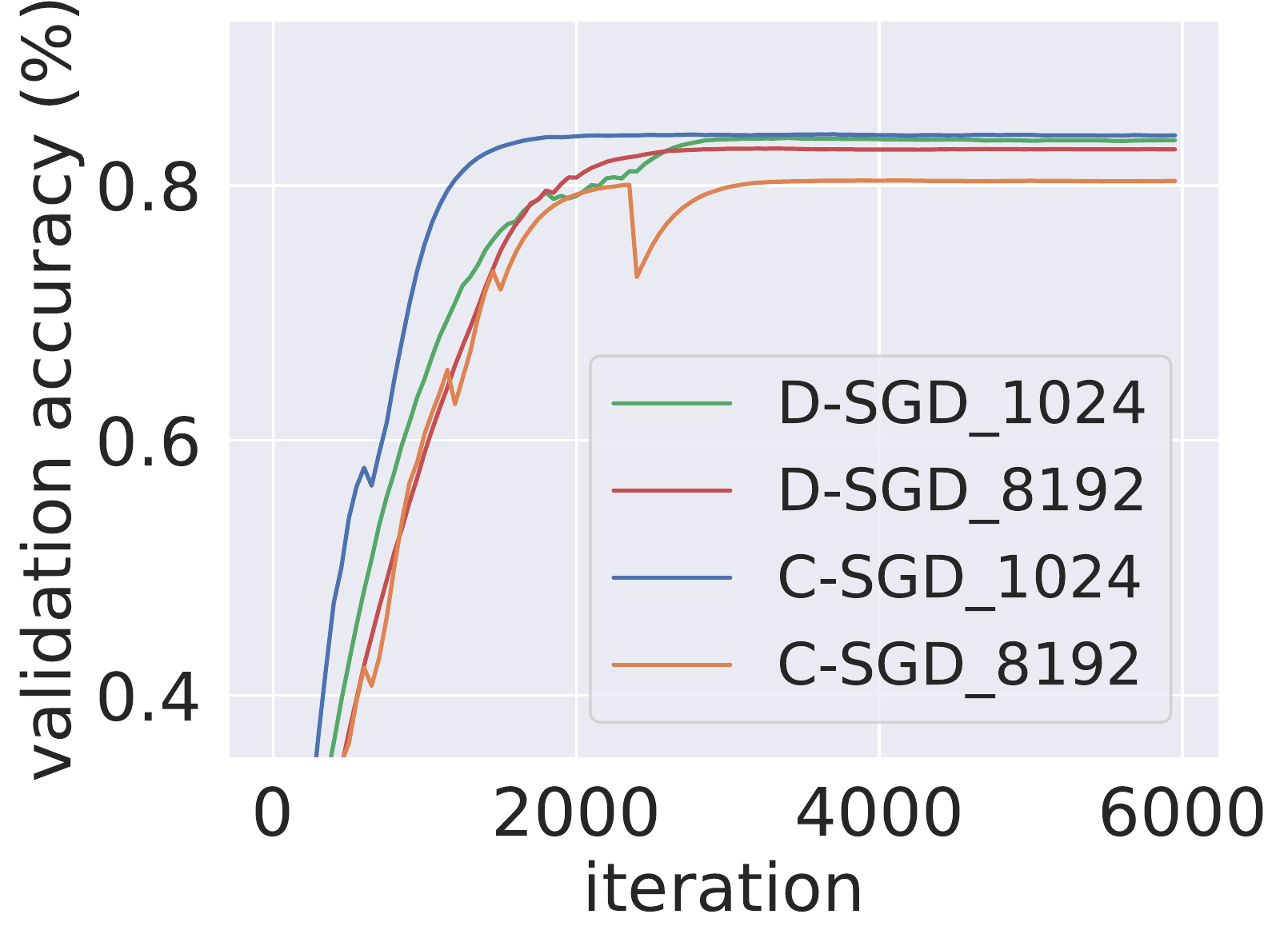} 
\caption{ResNet-18 (training from scratch)}
\end{subfigure}
\caption{The validation accuracy comparison of ResNet-18 with (left) and without pretraining (right) trained on CIFAR-10 using C-SGD and D-SGD. The number of workers (one GPU as a worker) is set as 16; and the local batch size is set as 64, and 512 per worker.}
\label{fig: test-accuracy_wo_pretraining}
\end{figure*}

One can observe from \cref{fig: test-accuracy_wo_pretraining} that  the gap in validation accuracy between C-SGD and D-SGD is alleviated in the training-from-scratch settings, which we attributes to the initial optimization difficulties of decentralized training without pretrained weights (see the discussion in \cref{sec: discussion}). The findings align with our insight that the generalization benefits of decentralization are more pronounced when optimization is not a significant obstacle.
We also note that without pretraining, the accuracy curves of D-SGD are notably smoother than that of C-SGD, which supports the theoretical results in \cref{coro: smoothing}.

\textbf{Additional experiments on LAMB.}
Large-batch training is of significant interest for deep learning deployment. However, linearly scaling the learning rate with the batch size can lead to generalization degradation \citep{JMLR:v20:18-789,smith2020generalization,smith2021on,li2021validity}. To address this issue, specialized methods have been developed to carefully tune the scaling factor between the learning rate and batch size, such as Layer-wise Adaptive Rate Scaling (LARS) \citep{you2017large} and Layer-wise Adaptive Moments (LAMB) \citep{You2020Large}. LARS calculates a scaling factor based on the ratio of the norm of the weight matrix to the norm of the weight gradients for each layer, while LAMB incorporates adaptive moment estimation. These methods have been shown to improve the convergence rate and generalization performance of large-batch training.
We compare the validation accuracy of centralized LAMB (C-LAMB) and decentralized LAMB (D-LAMB)\footnote{C-LAMB refers to the original LAMB and D-LAMB refers to the decentralized version of LAMB where the SGD optimizer in D-SGD is replaced by LAMB.}.
We follow the baseline learning rate setups (i.e., 0.0035 and 0.01) in \citet{You2020Large} and conduct experiments with other different learning rates, which is shown below. 

\begin{figure*}[h!]
\centering
\begin{subfigure}[ResNet-18 on CIFAR-10 (C-LAMB)]{.33\textwidth}
  \centering
  % include fourth image
  \includegraphics[width=1.0\linewidth]{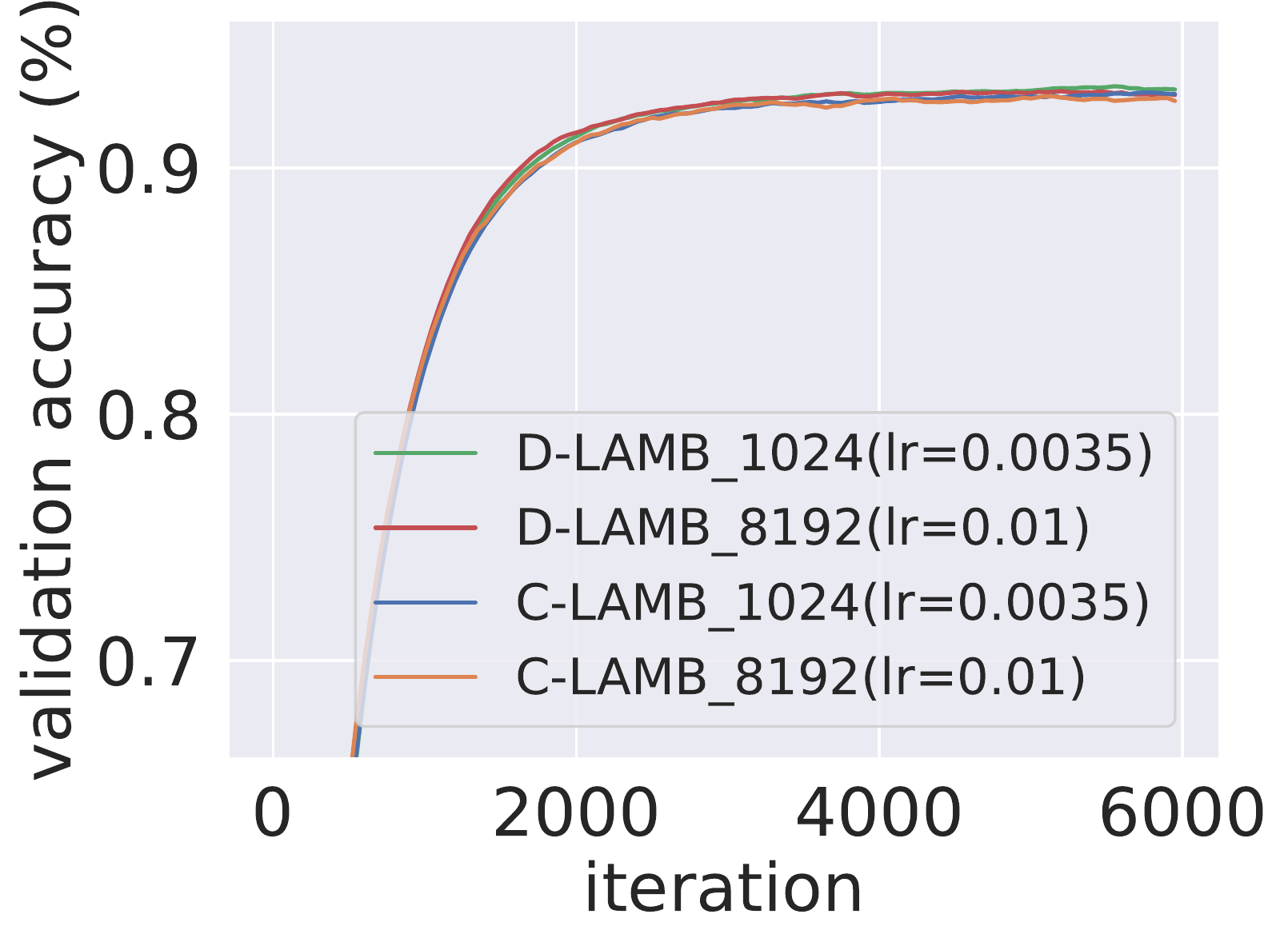}  
\end{subfigure}
\begin{subfigure}[ResNet-18 on CIFAR-10 (C-LAMB)]{.33\textwidth}
  \centering
  % include fourth image
  \includegraphics[width=1.0\linewidth]{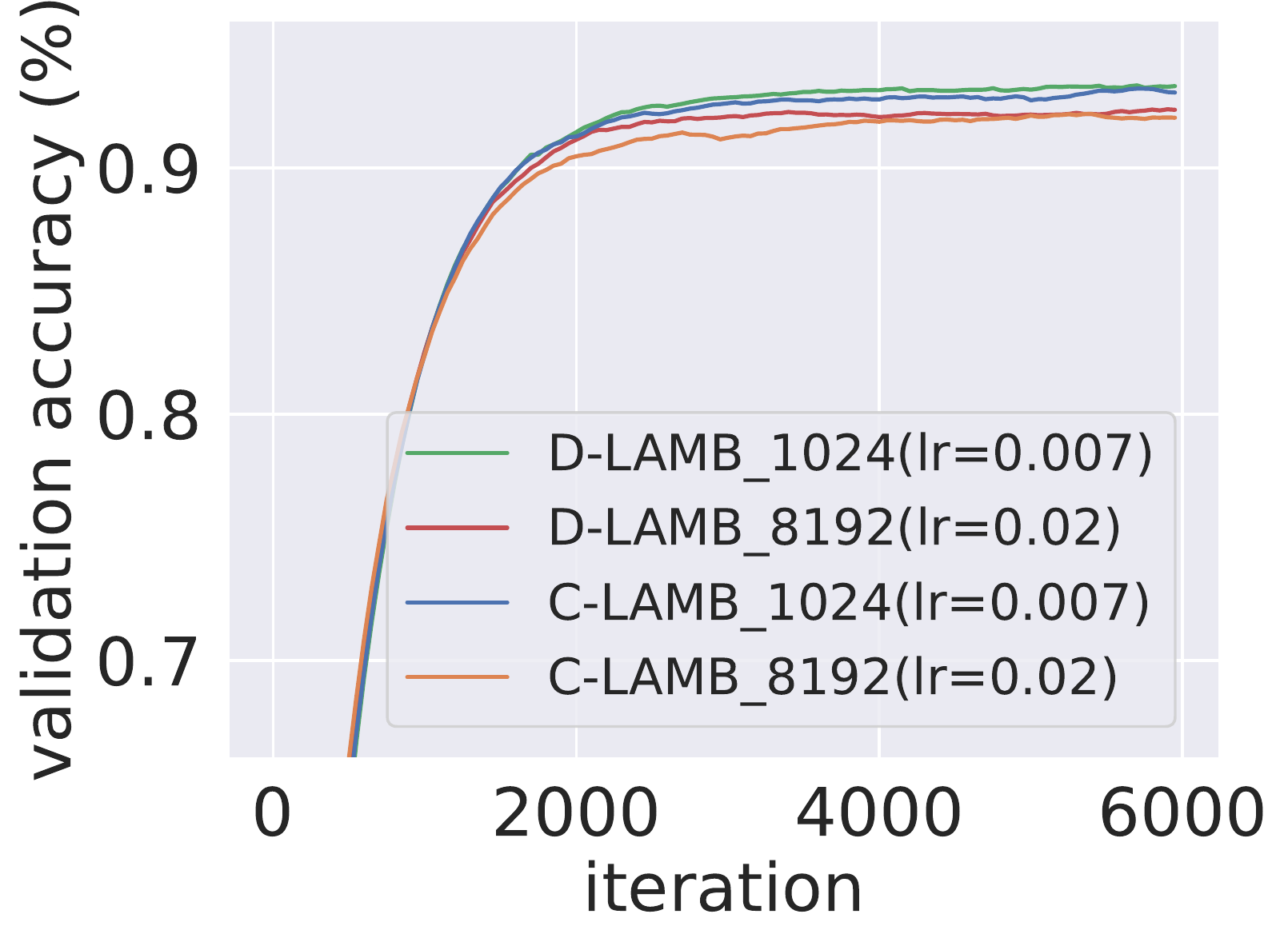}  
\end{subfigure}
\begin{subfigure}[ResNet-18 on CIFAR-10 (C-LAMB)]{.33\textwidth}
  \centering
  % include fourth image
  \includegraphics[width=1.0\linewidth]{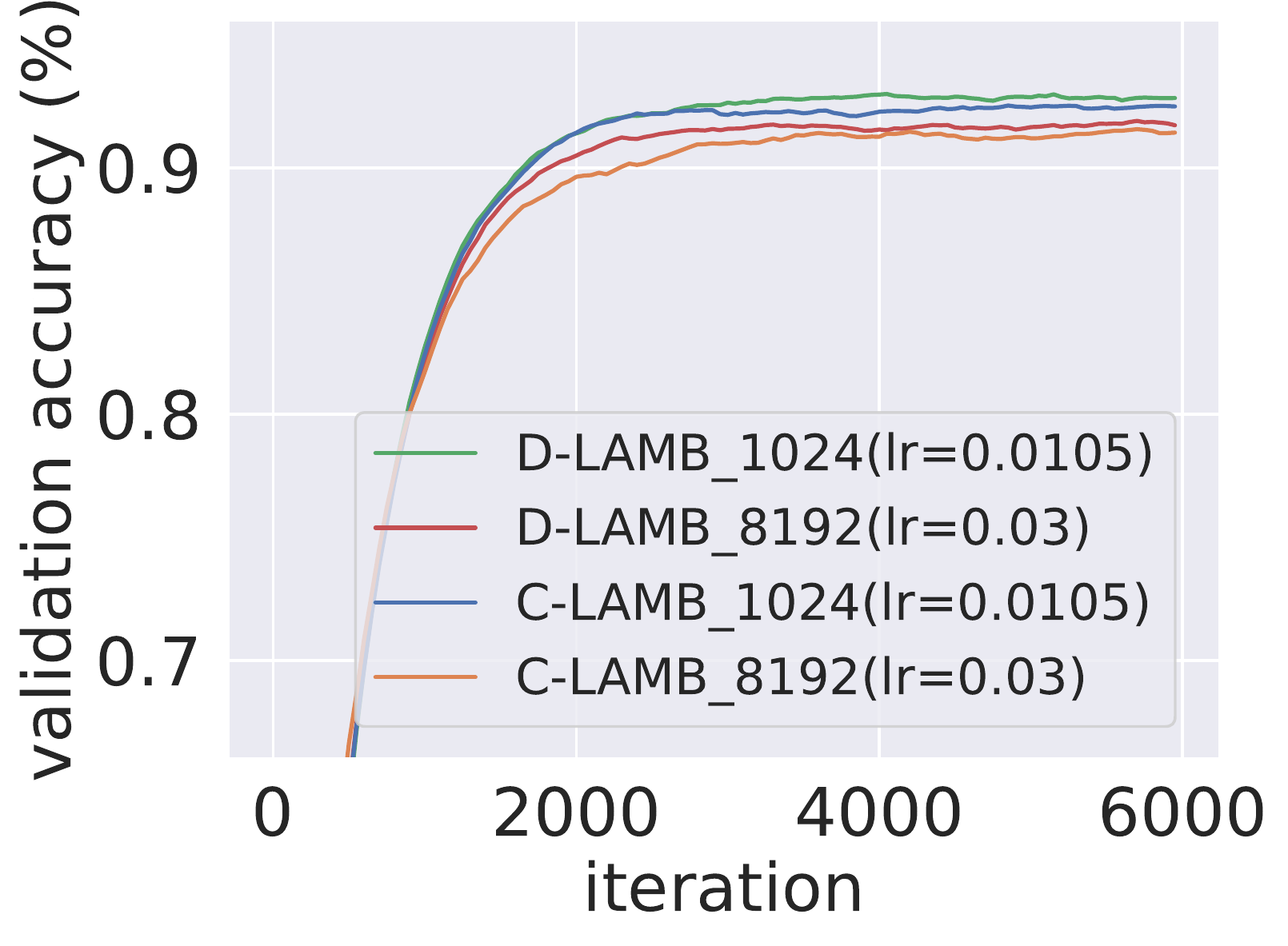}
\end{subfigure}
\caption{The validation accuracy comparison of ResNet-18 trained on CIFAR-10 using centralized LAMB (C-LAMB) and decentralized LAMB (D-LAMB, ring topology) with different learning rates. The number of workers (one GPU as a worker) is set as 16; and the local batch size is set as 512 per worker (8196 total batch size). The baseline learning rate setups (i.e., 0.0035 and 0.01) follow \citet{You2020Large}.} 
\label{fig: test-accuracy_4}
\end{figure*}

The best validation accuracy of  C-LAMB training  ResNet-18 on CIFAR-10 is 93.06 and 92.03 for 1024 and 8192 total batch size settings, respectively.
The best validation accuracy of  D-LAMB training  ResNet-18 on CIFAR-10 is 93.32 and 92.95 for 1024 and 8192 total batch size settings, respectively.
We find LAMB can mitigate the gap in generalizability between centralized training and decentralized training. However, decentralization still offers sight performance benefit on LAMB optimizer on multiple learning rates.
Is is worth noticing that compared with LARS and LAMB, decentralization incurs \hltext{VeryLightBlue}{\textit{zero additional computation}} overhead (e.g., no expensive computation of weight norm and gradient norm).

\textbf{Minima visualizations.}
The following figures depicts the local loss landscape round minima learned by C-SGD and D-SGD.

\begin{figure*}[h!]
\centering
\begin{subfigure}[ResNet-18 on CIFAR-10 (C-SGD), 128 total batch size]{.26\textwidth}
  \centering
  % include fourth image
  \includegraphics[width=.83\linewidth]{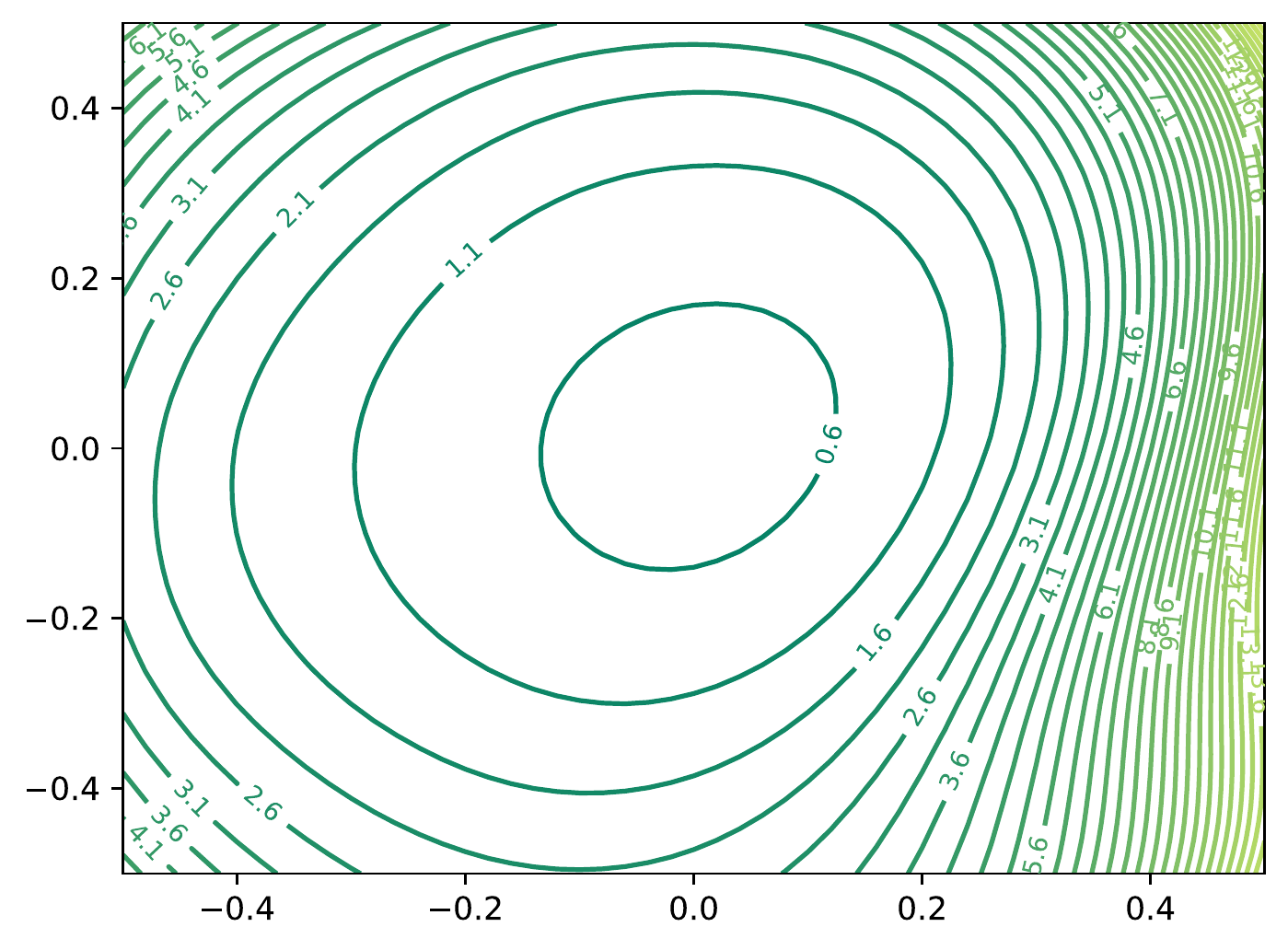}
  \caption{C-SGD, 128 total batch size}
\end{subfigure}
\begin{subfigure}[ResNet-18 on CIFAR-10 (C-SGD), 1024 total batch size]{.26\textwidth}
  \centering
  % include fourth image
  \includegraphics[width=.83\linewidth]{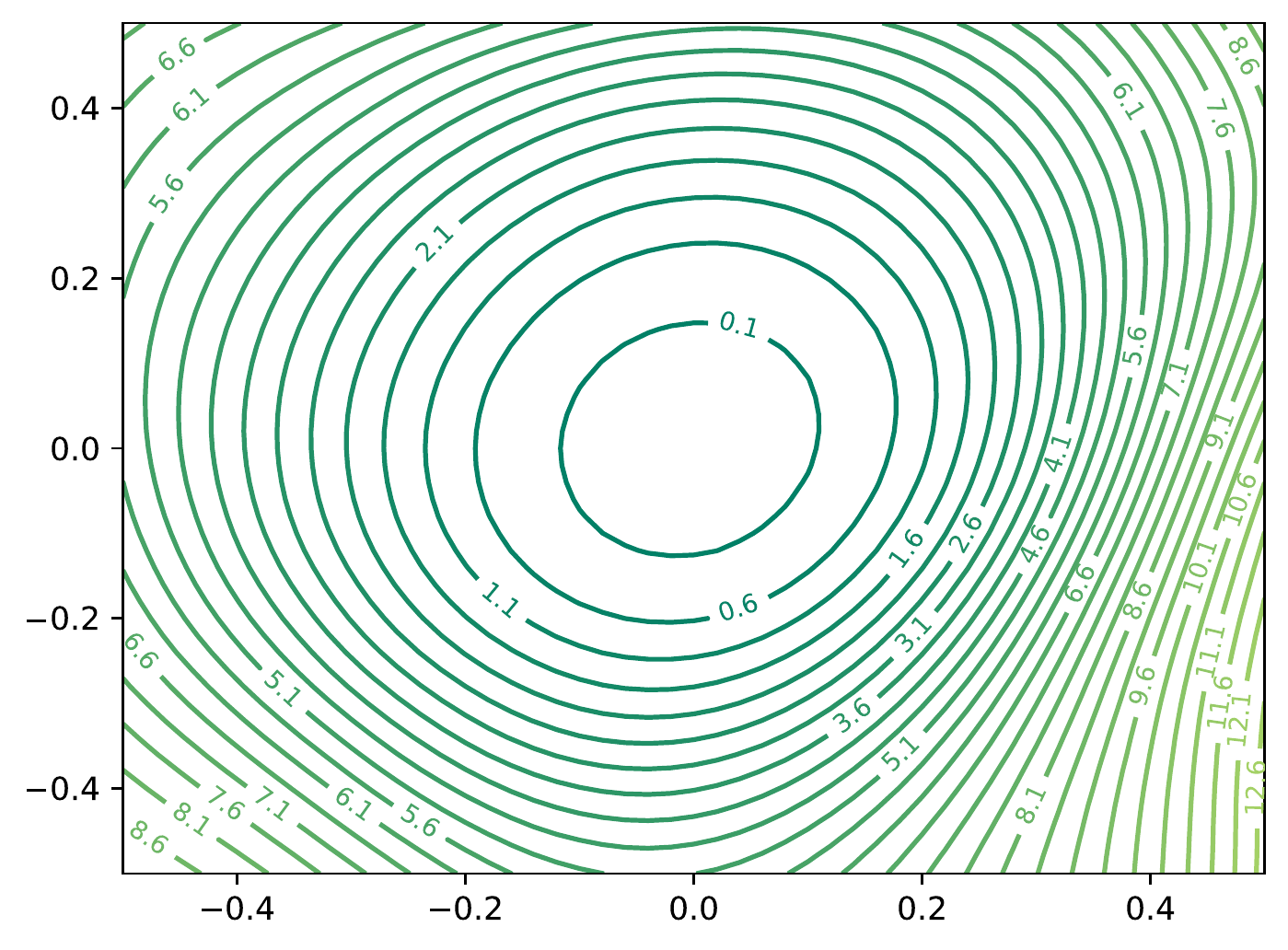}  
  \caption{C-SGD, 1024 total batch size}
\end{subfigure}
\begin{subfigure}[ResNet-18 on CIFAR-10 (C-SGD), 8196 total batch size]{.26\textwidth}
  \centering
  % include fourth image
  \includegraphics[width=.83\linewidth]{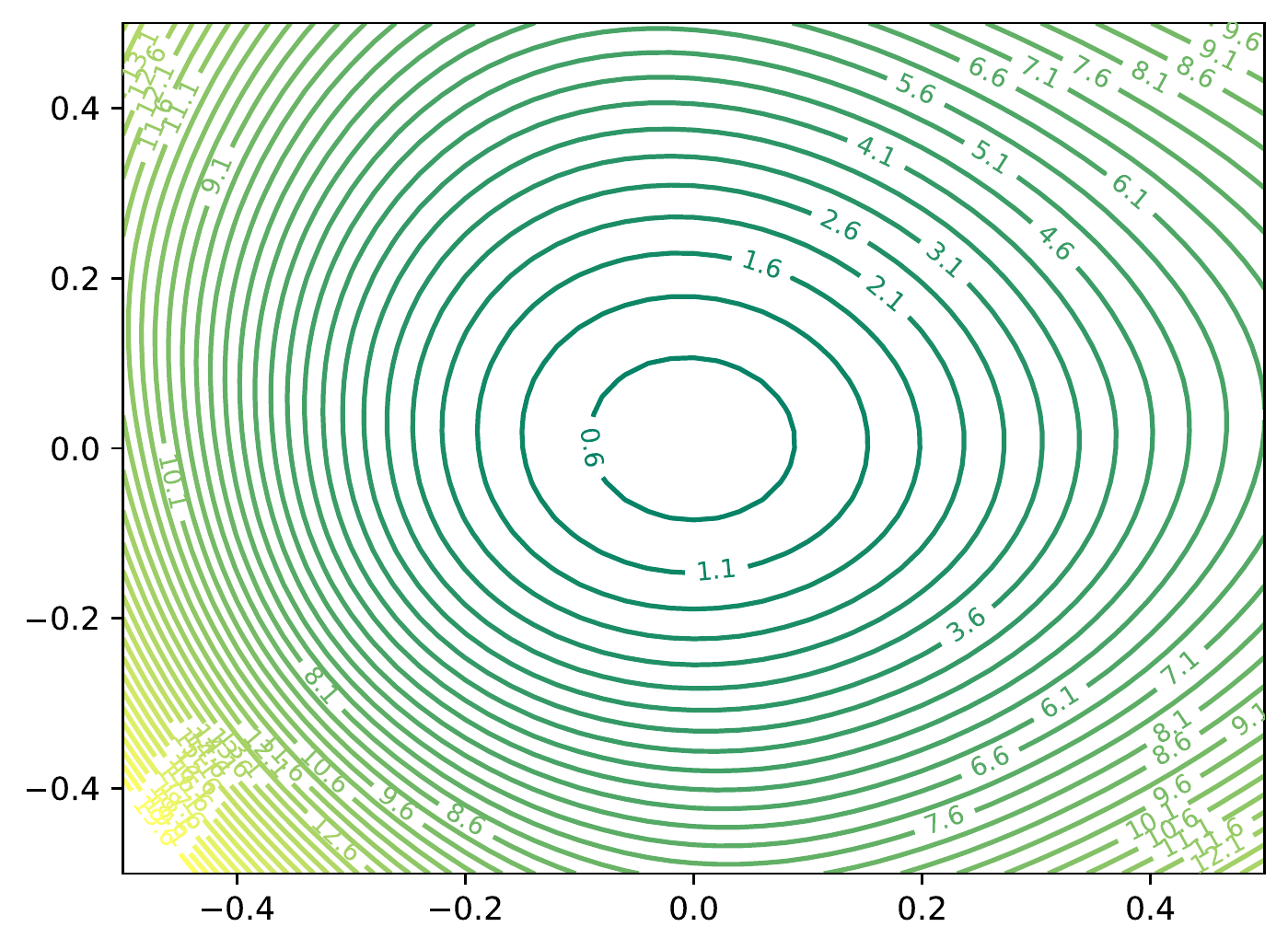}  
  \caption{C-SGD, 8196 total batch size}
\end{subfigure}
\medskip

\begin{subfigure}[ResNet-18 on CIFAR-10 (D-SGD), 128 total batch size]{.26\textwidth}
  \centering
  % include fourth image
  \includegraphics[width=.83\linewidth]{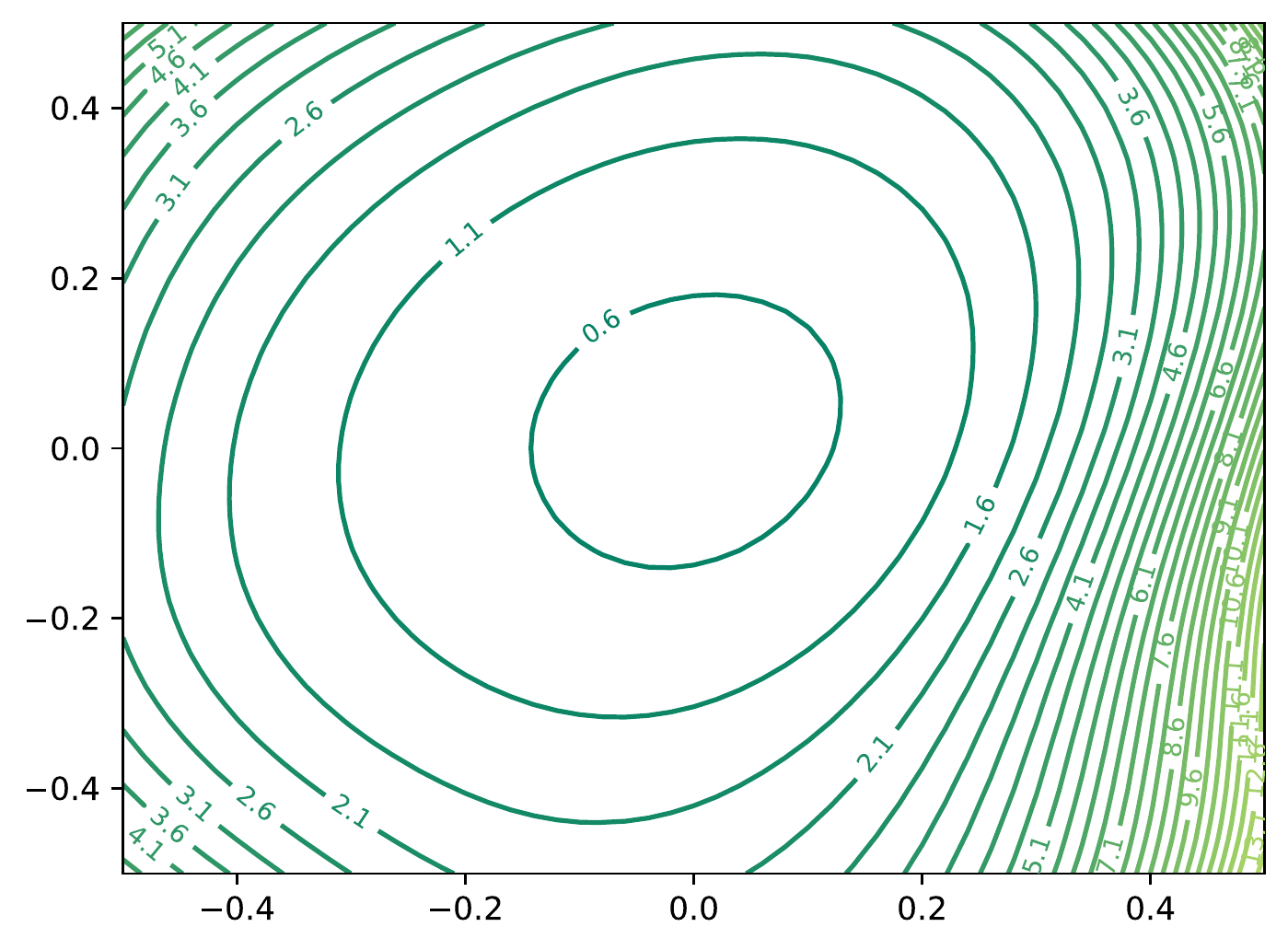}  
  \caption{D-SGD, 128 total batch size}
\end{subfigure}
\begin{subfigure}[ResNet-18 on CIFAR-10 (D-SGD), 1024 total batch size]{.26\textwidth}
  \centering
  % include fourth image
  \includegraphics[width=.83\linewidth]{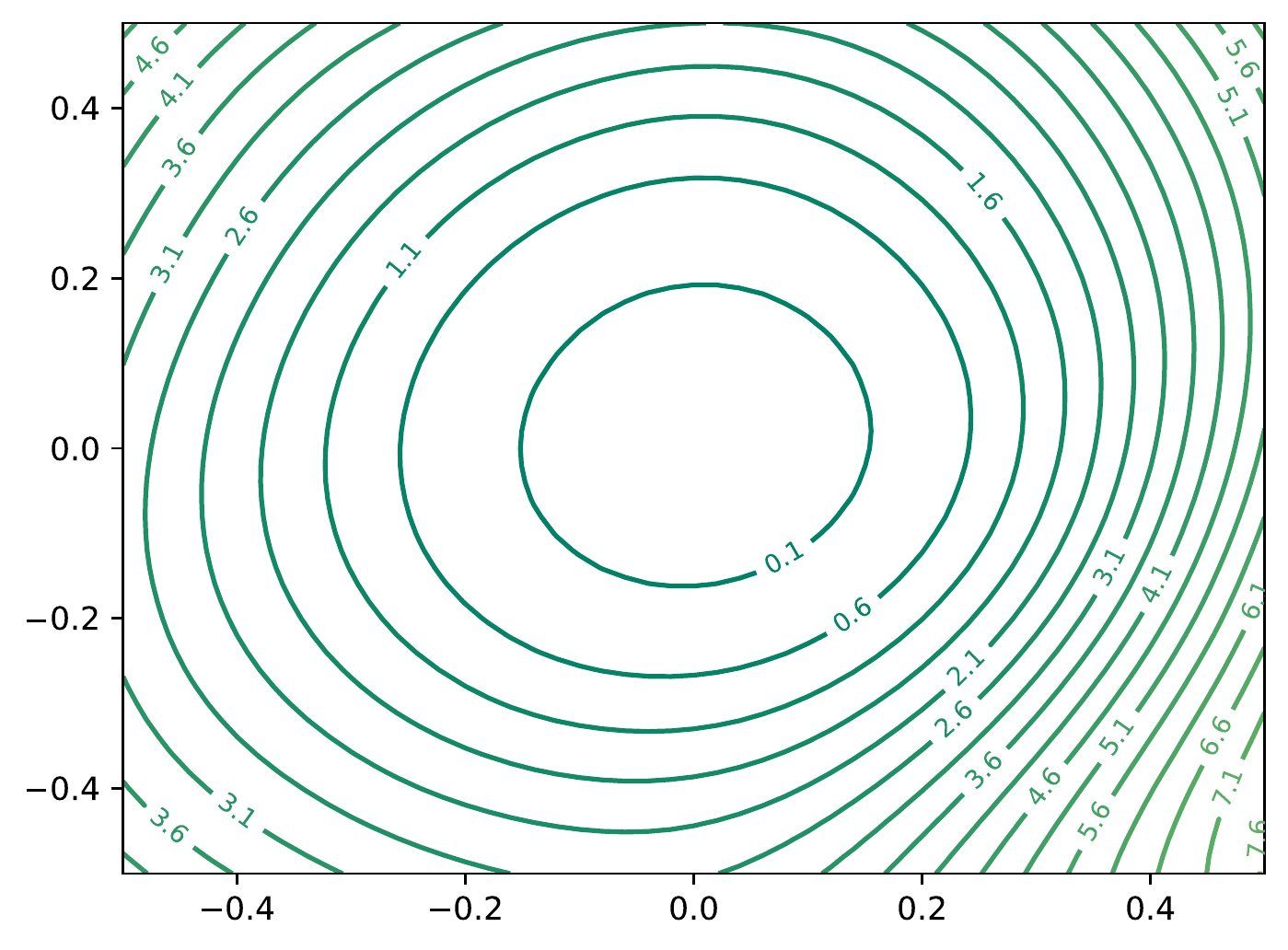}  
  \caption{D-SGD, 1024 total batch size}
\end{subfigure}
\begin{subfigure}[ResNet-18 on CIFAR-10 (D-SGD), 8196 total batch size]{.26\textwidth}
  \centering
  % include fourth image
  \includegraphics[width=.83\linewidth]{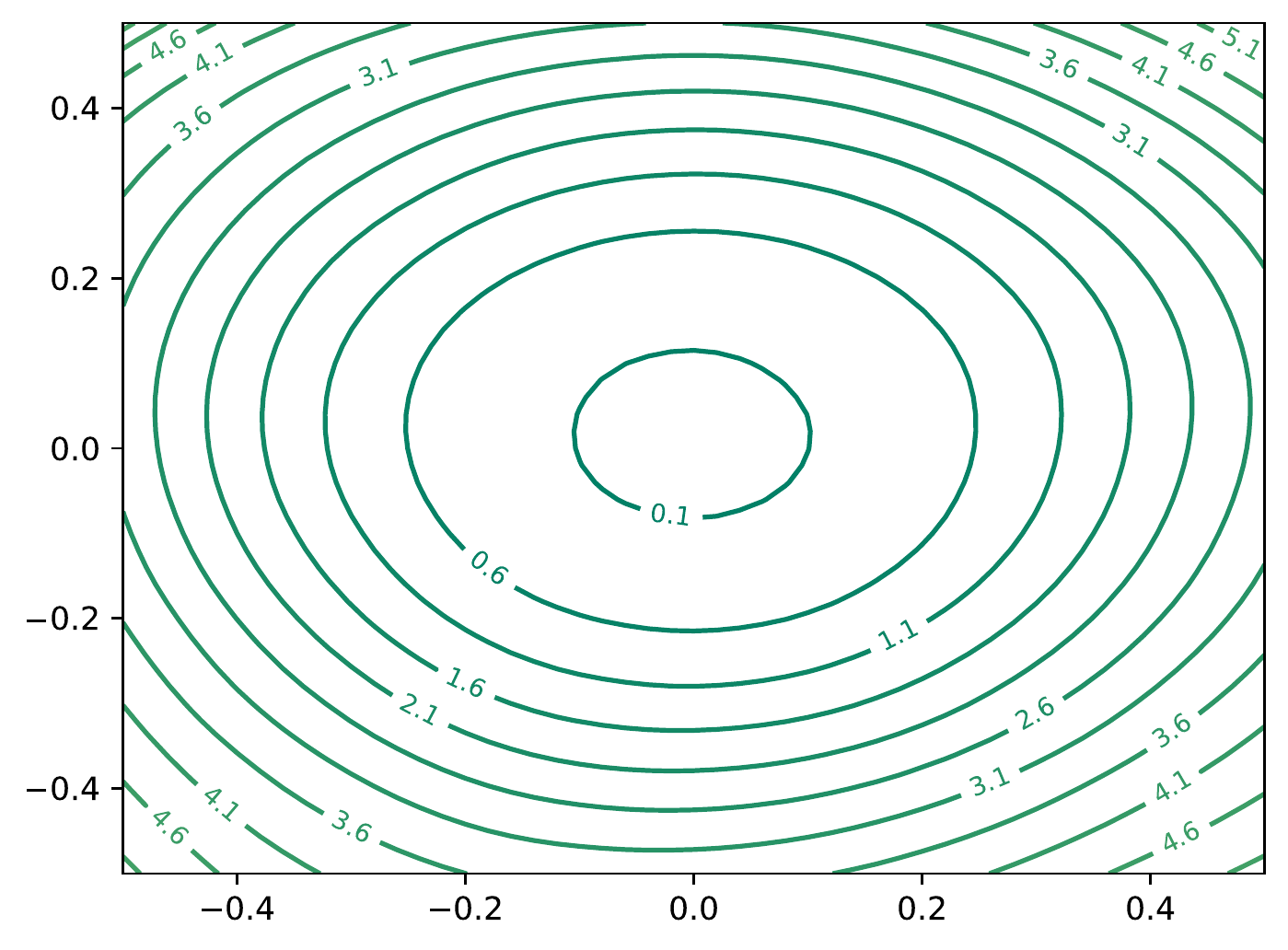}  
  \caption{D-SGD, 8196 total batch size}
\end{subfigure}
\caption{Minima 2D visualization of ResNet-18 trained on CIFAR-10 using C-SGD and D-SGD (Ring).}
\label{fig: 2d-minima-cifar10}
\end{figure*}

\begin{figure*}[ht!]
\centering
\begin{subfigure}[D-SGD (Ring), 1024 total batch size]{.23\textwidth}
  \centering
  % include fourth image
  \includegraphics[width=0.91\linewidth]{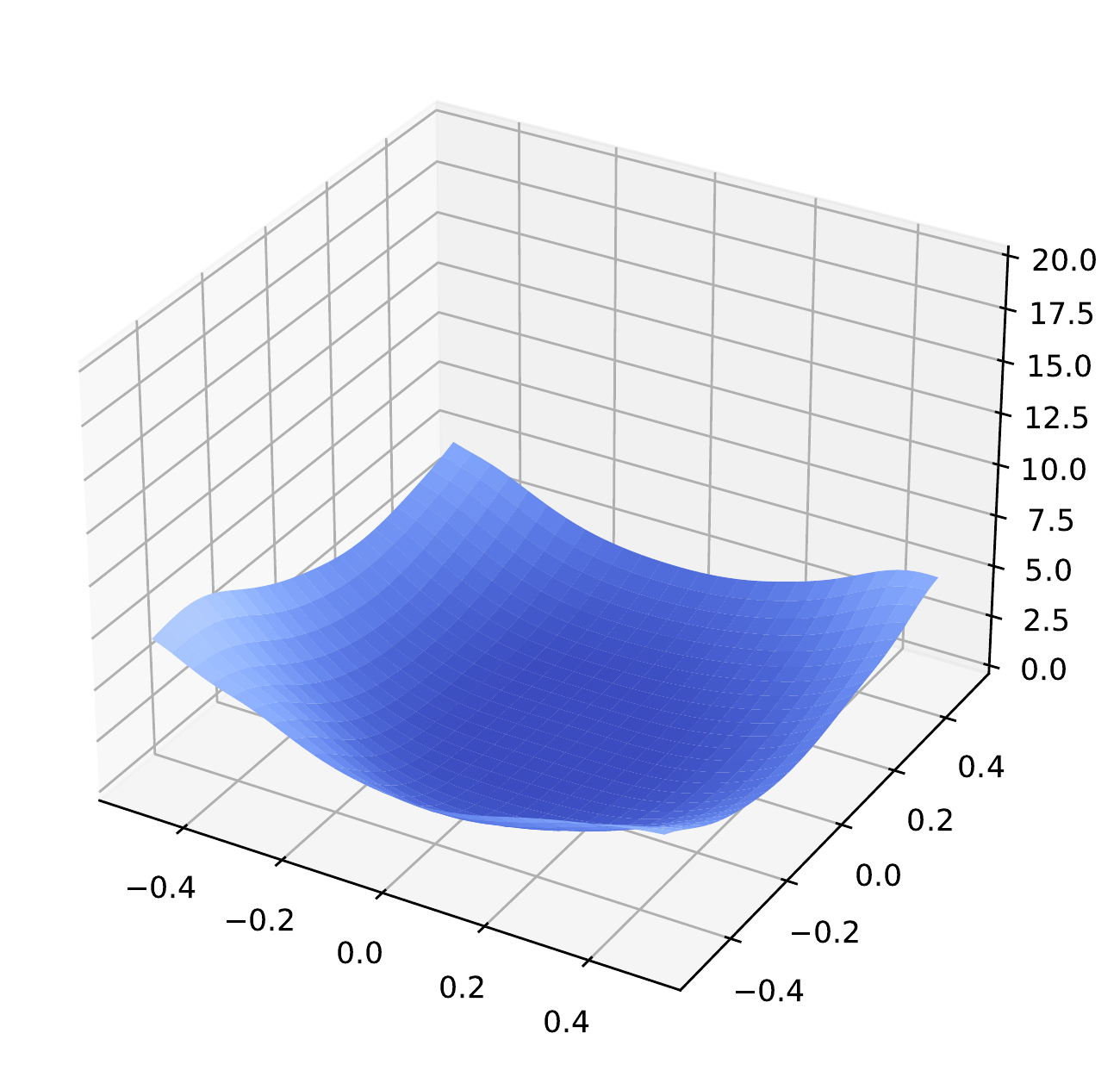}  
  \centering
  \captionsetup{justification=centering}
  \caption{D-SGD (Ring),\\ 1024 total batch size}
\end{subfigure}
\begin{subfigure}[D-SGD (Grid), 1024 total batch size]{.23\textwidth}
  \centering
  % include fourth image
  \includegraphics[width=0.91\linewidth]{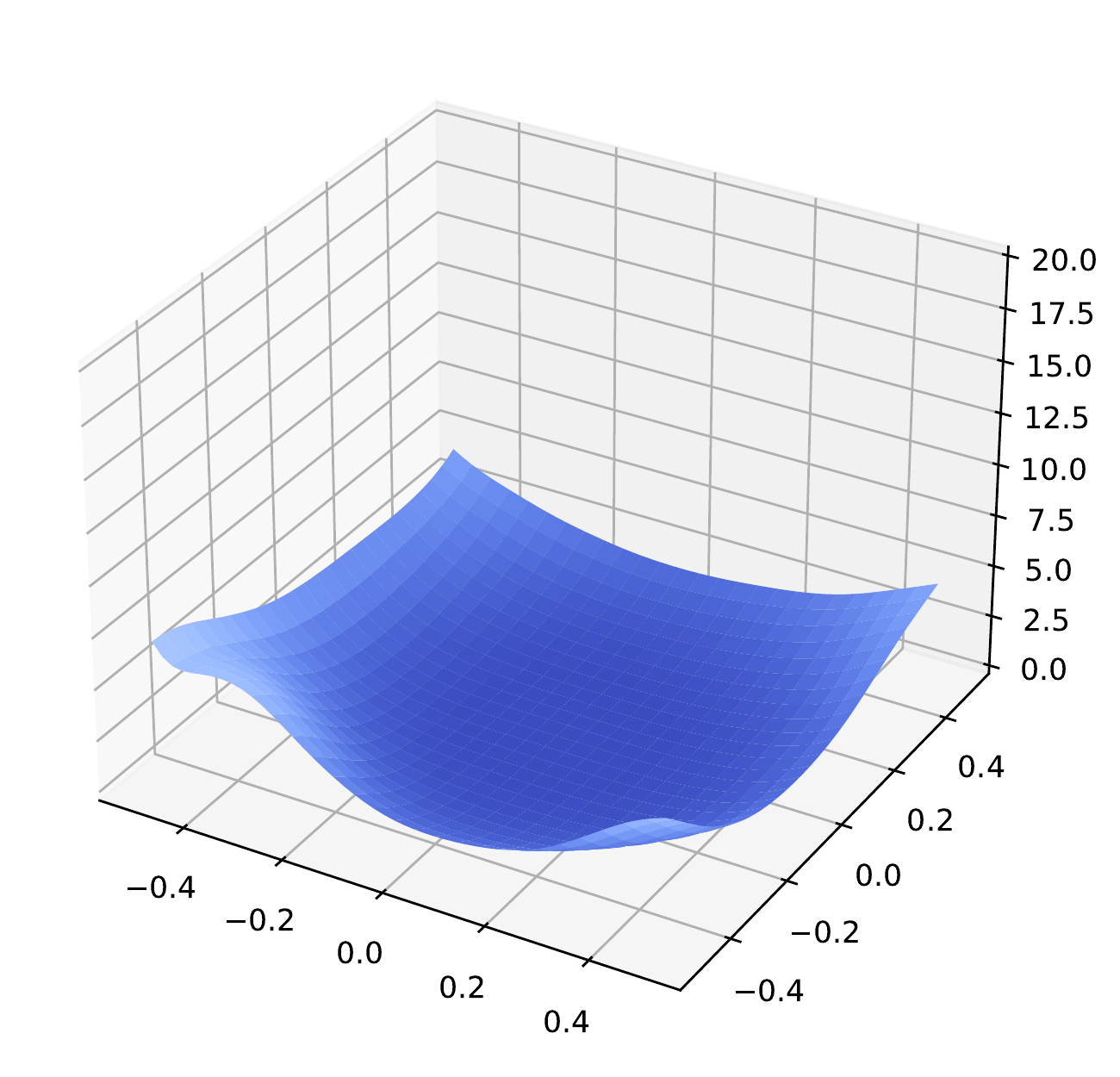}  
  \centering
  \captionsetup{justification=centering}  
  \caption{D-SGD (Grid),\\ 1024 total batch size}
\end{subfigure}
\begin{subfigure}[D-SGD (Exponential), 1024 total batch size]{.23\textwidth}
  \centering
  % include fourth image
  \includegraphics[width=0.91\linewidth]{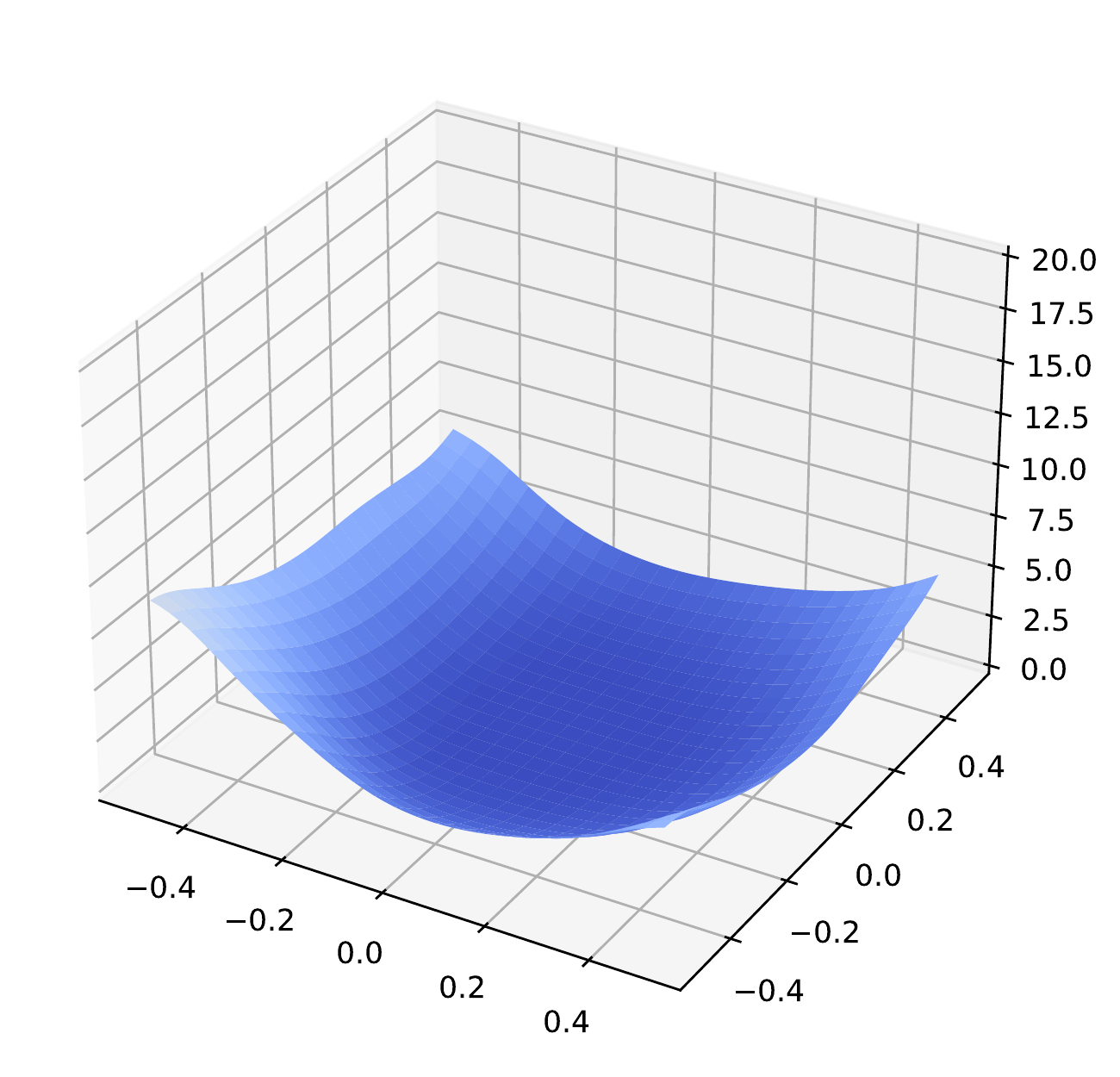}  
  \centering
  \captionsetup{justification=centering}  
  \caption{D-SGD (Exponential),\\ 1024 total batch size}
\end{subfigure}
\begin{subfigure}[D-SGD (Exponential), 1024 total batch size]{.23\textwidth}
  \centering
  % include fourth image
  \includegraphics[width=0.91\linewidth]{section/Figures/3D/CIFAR10-ResNet18-csgd-16-64.pdf}  
  \centering
  \captionsetup{justification=centering}  
  \caption{C-SGD,\\ 1024 total batch size}
\end{subfigure}
\medskip

\begin{subfigure}[D-SGD (Ring), 8192 total batch size]{.23\textwidth}
  \centering
  % include fourth image
  \includegraphics[width=0.91\linewidth]{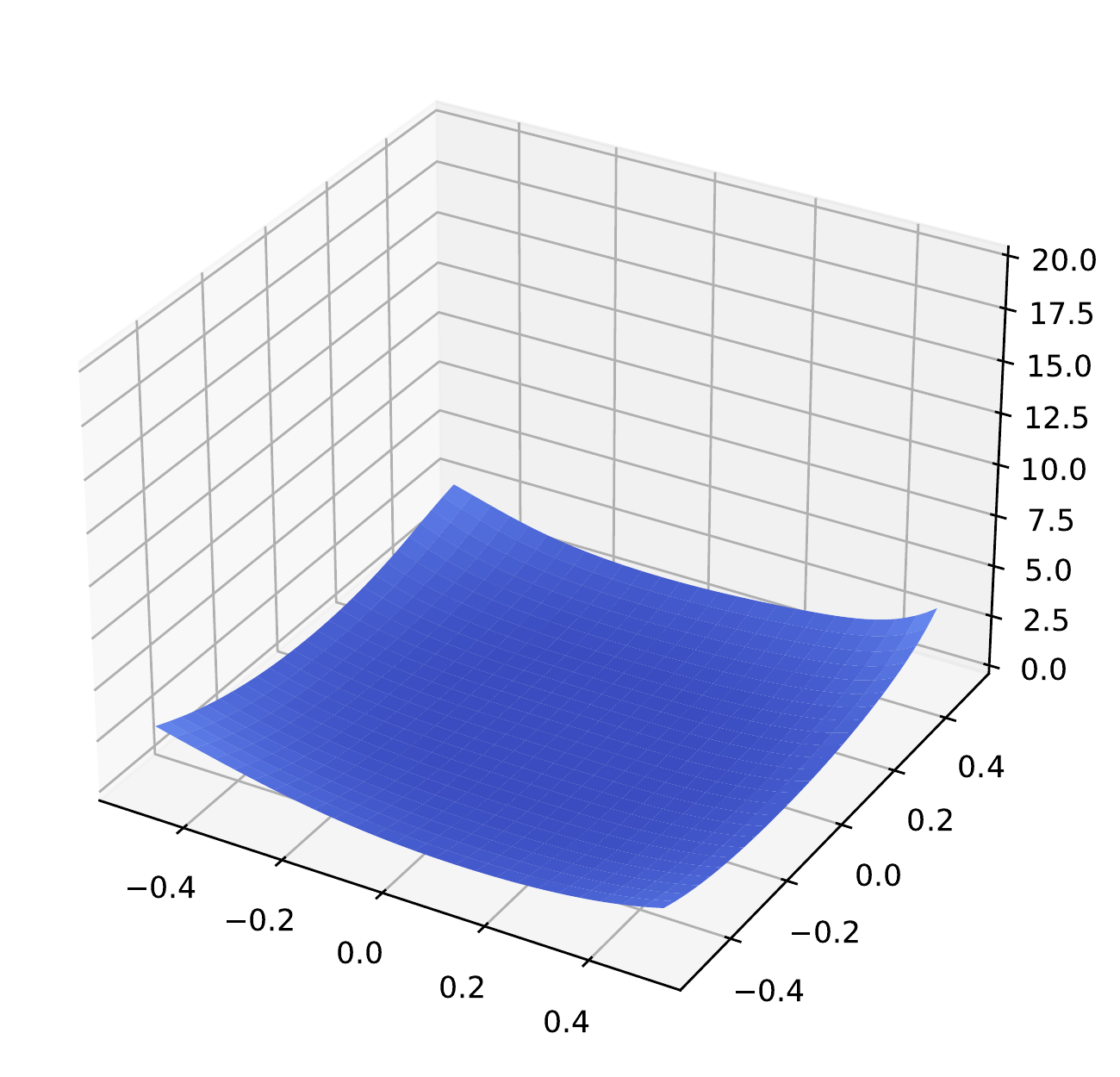}  
  \centering
  \captionsetup{justification=centering}  
  \caption{D-SGD (Ring),\\ 8192 total batch size}
\end{subfigure}
\begin{subfigure}[D-SGD (Grid), 8192 total batch size]{.23\textwidth}
  \centering
  % include fourth image
  \includegraphics[width=0.91\linewidth]{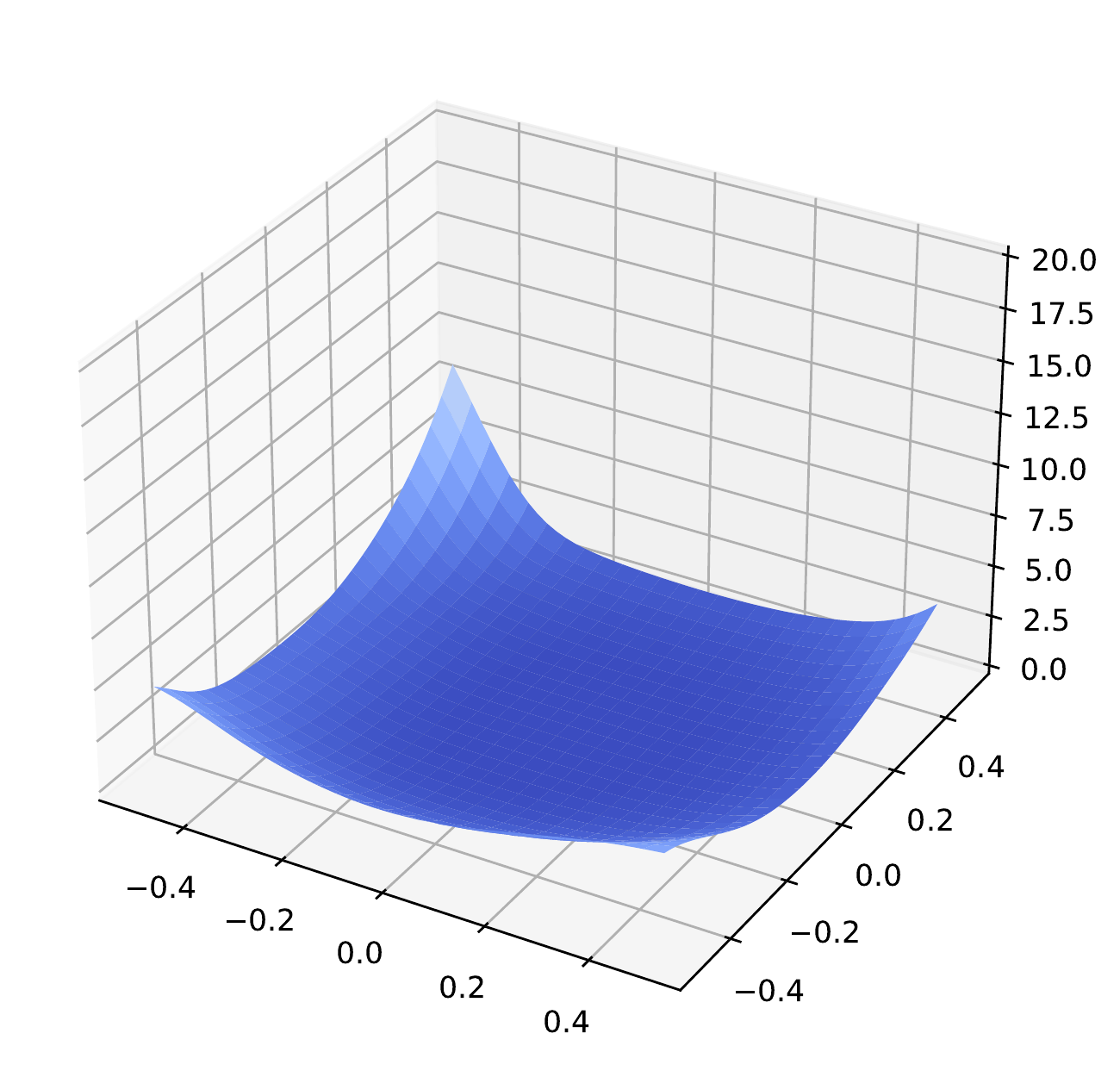}  
  \centering
  \captionsetup{justification=centering}  
  \caption{D-SGD (Grid),\\ 8192 total batch size}
\end{subfigure}
\begin{subfigure}[D-SGD (Exponential), 8196 total batch size]{.23\textwidth}
  \centering
  % include fourth image
  \includegraphics[width=0.91\linewidth]{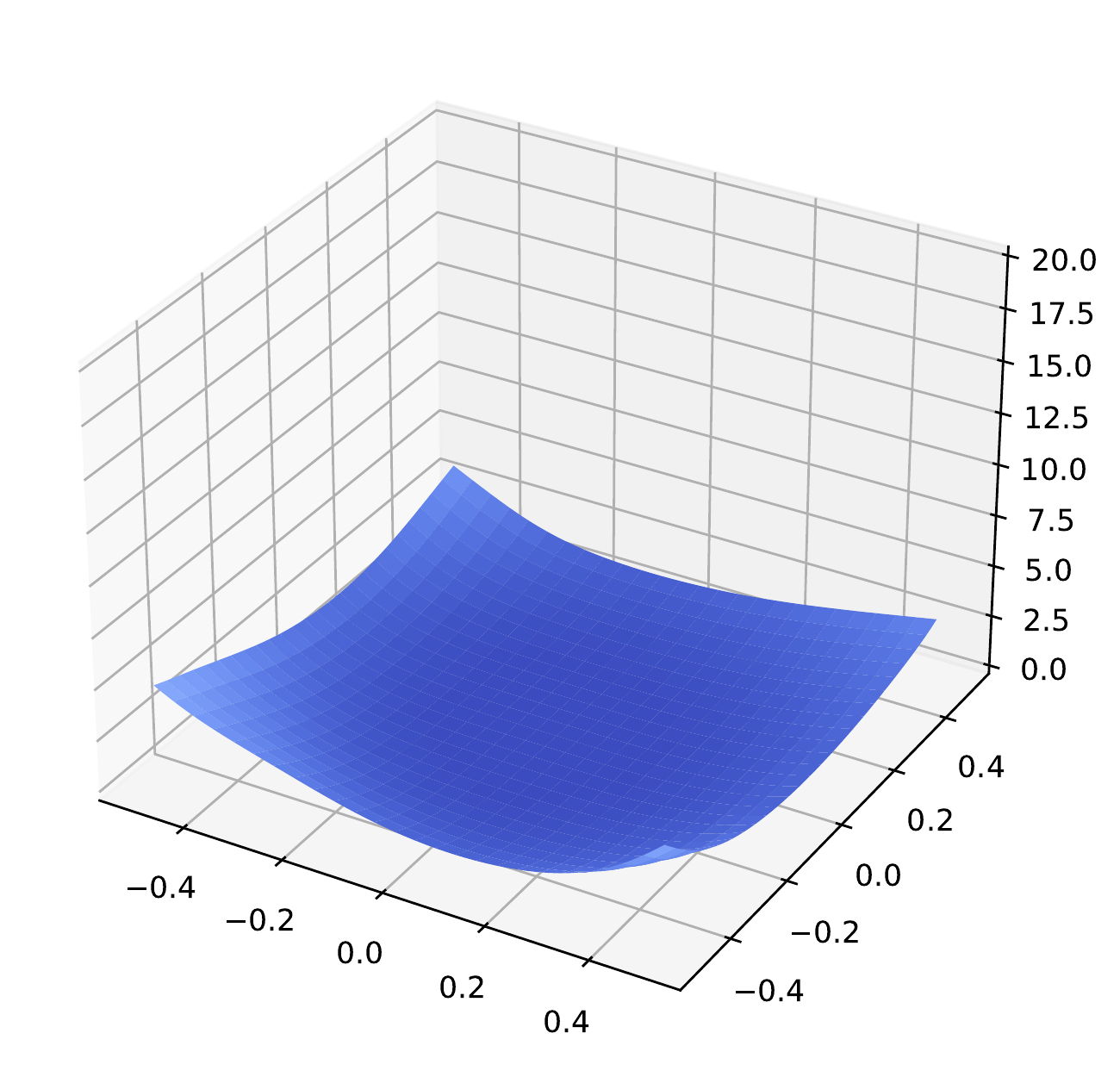}  
  \centering
  \captionsetup{justification=centering}  
  \caption{D-SGD (Exponential),\\ 8196 total batch size}
\end{subfigure}
\begin{subfigure}[D-SGD (exponential), 1024 total batch size]{.23\textwidth}
  \centering
  % include fourth image
  \includegraphics[width=0.91\linewidth]{section/Figures/3D/CIFAR10-ResNet18-csgd-16-512.pdf}  
  \centering
  \captionsetup{justification=centering}  
  \caption{C-SGD,\\ 8196 total batch size}
\end{subfigure}
\caption{Minima 3D visualization of ResNet-18 trained on CIFAR-10 using D-SGD with ring, grid-like and exponential topologies.}
\label{fig: 3d-minima-cifar10-complete}
\end{figure*}

We also compare the minima learned by C-SGD and D-SGD with multiple topologies in \cref{fig: 3d-minima-cifar10-complete}.

From \cref{fig: 2d-minima-cifar10} and \cref{fig: 3d-minima-cifar10-complete}, one can observe that (1) the minima of D-SGD with multiple commonly-used topologies could be flatter than those of C-SGD; and that (2) the gap in flatness increases as the total batch size increases. The findings support the claims made by \cref{th: dsgd-sam} and \cref{th: large-batch}.

\newpage

\section{Proof\label{sec:proof}}
\subsection{Propositions on the consensus distance\label{sec: lemma}}

We will introduce some useful propositions about the consensus distance, as shown below.
\begin{tcolorbox}[notitle, rounded corners, colframe=middlegrey, colback=lightblue, 
       boxrule=2pt, boxsep=0pt, left=0.15cm, right=0.17cm, enhanced, 
       toprule=2pt,
    ]
\begin{proposition}[\citep{kong2021consensus}\label{prop: consensus-distance}]
Suppose that the averaged gradient norm satisfies $\frac{1}{m} \sum_{j=1}^{m}\left\|\nabla \mL\left(\bw_{j}{\scriptstyle (t)}\right)\right\|^{2}\leq (1+\frac{1-\lambda}{4})\frac{1}{m} \sum_{j=1}^{m}\left\|\nabla \mL\left(\bw_{j}{\scriptstyle (t+1)}\right)\right\|^{2}$, then the consensus distance of D-SGD satisfies
\begin{align*}
&\Tr(\bm{\Xi}{\scriptstyle (t)})
=\frac{1}{m}\sum_{j=1}^{m}\|\bw_{j}{\scriptstyle (t)}-\bw_{a}{\scriptstyle (t)}\|_2^2\\
&=\lambda 
\cdot\mathcal{O}{\left(\frac{\frac{1}{m} \sum_{j=1}^{m}\left\|\nabla \mL\left(\bw_{j}{\scriptstyle (t)}\right)\right\|^{2}}{{(1-\lambda)}^{2}}\unaryplus\frac{\frac{1}{m} \sum_{j=1}^{m} \mathbb{E}_{\mu_j{\scriptstyle (t)}\sim \mathcal{D}}\left\|\nabla \mL^{\mu_j{\scriptstyle (t)}}\left(\bw_{j}{\scriptstyle (t)}\right)\unaryminus\nabla \mL\left(\bw_{j}{\scriptstyle (t)}\right)\right\|_{2}^{2}}{1-\lambda}\right)},
\end{align*}
where $\lambda$ equals $1-\textnormal{spectral gap}$ (see \cref{def:spectral-gap}). 
\end{proposition}
\end{tcolorbox}

\begin{tcolorbox}[notitle, rounded corners, colframe=middlegrey, colback=lightblue, 
       boxrule=2pt, boxsep=0pt, left=0.15cm, right=0.17cm, enhanced, 
       toprule=2pt,
    ]
\begin{proposition}[Descent condition of $\Tr(\bm{\Xi}{\scriptstyle (t)})$\label{prop: consensus-distance-decrease}]
    If the learning rate $\eta$ satisfies
    \begin{align*}
        \eta\leq \frac{{\Tr(\bm{\Xi}{\scriptstyle (t)})}{(1-\lambda)}}{\sqrt{6}\lambda^{\frac{1}{2}}}{\left[\frac{1}{m} \sum_{j=1}^{m}\left\|\nabla \mL\left(\bw_{j}{\scriptstyle (t)}\right)\right\|^{2}+(1-\lambda)\cdot\frac{1}{m} \sum_{j=1}^{m} \mathbb{E}_{\mu_j{\scriptstyle (t)}\sim \mathcal{D}}\left\|\nabla \mL^{\mu_j{\scriptstyle (t)}}\left(\bw_{j}{\scriptstyle (t)}\right)\unaryminus\nabla \mL\left(\bw_{j}{\scriptstyle (t)}\right)\right\|_{2}^{2}\right]}^{-\frac{1}{2}},
    \end{align*}
    it holds that,
    \begin{align*}
        \Tr(\bm{\Xi}{\scriptstyle (t+1)})\leq\Tr(\bm{\Xi}{\scriptstyle (t)}).
    \end{align*}
\end{proposition}
\end{tcolorbox}
As the terms in brackets are of order $\mathcal{O}{\left(\frac{{(1-\lambda)}^2}{{\lambda}}\cdot{\Tr(\bm{\Xi}{\scriptstyle (t)})}\right)}$ according to \cref{prop: consensus-distance}, the assumption on $\eta$ becomes $\eta\leq \mathcal{O}{({{\Tr(\bm{\Xi}{\scriptstyle (t)})}}^{\frac{1}{2}})}$. The proof follows directly from Lemma C.2 in \citep{kong2021consensus}.

\begin{tcolorbox}[notitle, rounded corners, colframe=middlegrey, colback=lightblue, 
       boxrule=2pt, boxsep=0pt, left=0.15cm, right=0.17cm, enhanced, 
       toprule=2pt,
    ]
\begin{proposition}\label{prop: order-of-residuals}
Denote $\bm{\Xi}{\scriptstyle (t)}= \frac{1}{m}\sum_{j=1}^{m}(\bw_{j}{\scriptstyle (t)}\unaryminus\bw_{a}{\scriptstyle (t)}){(\bw_{j}{\scriptstyle (t)}\unaryminus\bw_{a}{\scriptstyle (t)})}^{\top} \in \rbb^{d\times d}$ as the weight diversity matrix at the step $t$.
It holds that,
\begin{align*}
\ebb_{\epsilon \sim \mathcal{N}(0, \bm{\Xi}{\scriptstyle (t)})}\|\epsilon\|_2^3
={(\Tr(\bm{\Xi}{\scriptstyle (t)}))}^{\frac{3}{2}}
={(\frac{1}{m}\sum_{j=1}^{m}\|\bw_{j}{\scriptstyle (t)}-\bw_{a}{\scriptstyle (t)}\|_2^2)}^{\frac{3}{2}}
\leq {(\frac{1}{m}\sum_{j=1}^{m}\|\bw_{j}{\scriptstyle (t)}-\bw_{a}{\scriptstyle (t)}\|_2^3)}.
\end{align*}
\end{proposition}
\end{tcolorbox}
The inequality is derived from the generalized mean inequality \citep{marshall1979inequalities}.

\subsection{Proof of \cref{th: dsgd-sam}\label{sec:proof-dsgd-sam}}

\begin{tcolorbox}[notitle, rounded corners, colframe=middlegrey, colback=lightblue, 
       boxrule=2pt, boxsep=0pt, left=0.15cm, right=0.17cm, enhanced, 
       toprule=2pt,
    ]
\begin{proposition}\label{prop: gradient-diversity-quadratic}
The gradient diversity in \cref{eq: graident-decomposition} equals to zero in the following cases: 

\vspace{0.5em}

(1) the loss is quadratic, i.e., $\mL=\bw^{\top}\mH\bw +A\bw+b$, where $A\in\rbb^{d\times d}$ and $b\in\rbb^d$; 

(2) the optimization algorithm is distributed centralized SGD (see \cref{def: c-sgd}).
\end{proposition}
\end{tcolorbox}

\textit{Proof of \cref{prop: gradient-diversity-quadratic}}.

On quadratic loss, we have
\begin{align*}
\frac{1}{m}\sum_{j=1}^{m}[\nabla \mL^{\mu_j{\scriptscriptstyle (t)}}\left(\bw_{j}{\scriptstyle (t)}\right)\unaryminus\nabla \mL^{\mu_j{\scriptscriptstyle (t)}}\left(\bw_{a}{\scriptstyle (t)}\right)]
= \frac{1}{m}\sum_{j=1}^{m} [\mH\bw_{j}{\scriptstyle (t)}\unaryplus A\unaryminus\mH\bw_{a}{\scriptstyle (t)}\unaryminus A]
= \mH\frac{1}{m}\sum_{j=1}^{m} [\bw_{j}{\scriptstyle (t)}\unaryminus\bw_{a}{\scriptstyle (t)}]=0.
\end{align*}
In distributed centralized SGD, the gradient diversity satisfies
\begin{align*}
\frac{1}{m}\sum_{j=1}^{m}[\nabla \mL^{\mu_j{\scriptscriptstyle (t)}}\underbrace{\left(\bw_{j}{\scriptstyle (t)}\right)}_{=\bw_{a}{\scriptstyle (t)}}\unaryminus\nabla \mL^{\mu_j{\scriptscriptstyle (t)}}\left(\bw_{a}{\scriptstyle (t)}\right)]
=0.
\end{align*}
{\color{magenta}\qed}

It can be deduced directly from \cref{prop: gradient-diversity-quadratic} that distributed centralized SGD, which has constant zero gradient diversity, is equivalent to standard single-worker mini-batch SGD with equivalently large batch size.

\begin{tcolorbox}[notitle, rounded corners, colframe=middlegrey, colback=lightblue, 
       boxrule=2pt, boxsep=0pt, left=0.15cm, right=0.17cm, enhanced, 
       toprule=2pt,
    ]
\textbf{Theorem 1} (D-SGD as SAM).
\textit{Suppose $\mL \in C^4\left(\mathbb{R}^d\right)$, i.e., $\mL$ is four times continuously differentiable. The mean iterate of the global averaged model $\bw_{a}{\scriptstyle (t)}=\frac{1}{m}\sum_{j=1}^m\bw_{j}{\scriptstyle (t)}$ of D-SGD can be written as follows:
\begin{align*}
\ebb_{\mu{\scriptscriptstyle (t)}}&[\bw_{a}{\scriptstyle (t+1)}]
=\bw_{a}{\scriptstyle (t)}-\eta\color{DarkBlue}{\underbrace{\color{black}{\tikzmarknode{descent_direction}{\hlmath{LightBlue}{\ebb_{\epsilon \sim \mathcal{N}(0, \bm{\Xi}{\scriptstyle (t)})}[\nabla\mL_{\bw_{a}{\scriptstyle (t)}+\epsilon}]}}}}_{\text{asymptotic descent direction}}}
+\color{black}{\underbrace{\mathcal{O}(\eta\ \ebb_{\epsilon \sim \mathcal{N}(0, \bm{\Xi}{\scriptstyle (t)})}\|\epsilon\|_2^3+\frac{\eta}{m}\sum_{j=1}^{m}\|\bw_{j}{\scriptstyle (t)}-\bw_{a}{\scriptstyle (t)}\|_2^3)}_{\text{higher-order residual terms}}},
\end{align*}
 where $\bm{\Xi}{\scriptstyle (t)}= \frac{1}{m}\sum_{j=1}^{m}(\bw_{j}{\scriptstyle (t)}-\bw_{a}{\scriptstyle (t)}){(\bw_{j}{\scriptstyle (t)}-\bw_{a}{\scriptstyle (t)})}^T$ denotes the weight diversity matrix. }
\end{tcolorbox}

\textit{Proof of \cref{th: dsgd-sam}.}

\textbf{(1) Derive the iterate of the averaged model $\bw_{a}{\scriptstyle (t)}$.}

Directly analyzing the dynamics of the diffusion-like decentralized systems where information is gradually spread across the network is non-trivial.
Instead, we focus on $\bw_{a}{\scriptstyle (t)}=\frac{1}{m}\sum_{j=1}^m\bw_{j}{\scriptstyle (t)}$, the global averaged model of D-SGD, whose update can be written as follows,
\begin{align}
\bw_{a}{\scriptstyle (t+1)}
\unaryequal\bw_{a}{\scriptstyle (t)}\unaryminus\eta\big[\nabla \mL^{\mu{\scriptscriptstyle (t)}}_{\bw_{a}{\scriptstyle (t)}}
\unaryplus\underbrace{\frac{1}{m}\sum_{j=1}^{m}(\nabla \mL^{\mu_j{\scriptscriptstyle (t)}}_{\bw_{j}{\scriptstyle (t)}}\unaryminus\nabla \mL^{\mu_j{\scriptscriptstyle (t)}}_{\bw_{a}{\scriptstyle (t)}})}_{\text {\textit{gradient diversity among local workers}}}\big].\label{eq: graident-decomposition-appendix}
\end{align} 

\textbf{Remark.}
\cref{eq: graident-decomposition-appendix}  shows that decentralization introduces an additional noise, which characterizes the gradient diversity between the global averaged model $\bw_{a}{\scriptstyle (t)}$ and the local models $\bw_{j}{\scriptstyle (t)}$ for $j\unaryequal1, \dots, m$, compared with its centralized counterpart. 
Therefore, we note that 
\begin{center}
    \hltext{VeryLightBlue}{\textit{analyzing the gradient diversity is the major challenge of decentralized (gradient-based) learning}}.
\end{center}
\textbf{Insight.}
We also note that the gradient diversity equals to zero on quadratic objective $\mL$ (see \cref{prop: gradient-diversity-quadratic}). Therefore, the quadratic approximation of loss functions $\mL$ \citep{pmlr-v97-zhu19e, ibayashi2021quasi,liu2021noise,ziyin2022strength} might be insufficient to characterize how decentralization impacts the training dynamics of D-SGD, especially on neural network loss landscapes where quadratic approximation may not be accurate even around minima \citep{JML-1-247}. To better understand the dynamics of D-SGD on complex landscapes, it is crucial to consider \hltext{VeryLightBlue}{\textit{higher-order geometric information}} of objective $\mL$. In the following, we approximate the gradient diversity using Taylor expansion, instead of analyzing it on non-convex non-$\beta$-smooth loss $\mL$ directly, which is highly non-trivial.

\textbf{(2) Perform Taylor expansion on the gradient diversity.}
 Technically, we perform a second-order Taylor expansion on the gradient diversity around $\bw_{a}{\scriptstyle (t)}$:
\begin{align*}
\frac{1}{m}&\sum_{j=1}^{m}(\nabla \mL^{\mu_j{\scriptscriptstyle (t)}}_{\bw_{j}{\scriptstyle (t)}}\unaryminus\nabla \mL^{\mu_j{\scriptscriptstyle (t)}}_{\bw_{a}{\scriptstyle (t)}})
= \frac{1}{m}\sum_{j=1}^{m} \mH^{\mu_j{\scriptscriptstyle (t)}}_{\bw_{a}{\scriptstyle (t)}}\cdot(\bw_{j}{\scriptstyle (t)}\unaryminus\bw_{a}{\scriptstyle (t)})
\unaryplus \frac{1}{2m}\sum_{j=1}^{m} \mT^{\mu_j{\scriptscriptstyle (t)}}_{\bw_{a}{\scriptstyle (t)}}\otimes[(\bw_{j}{\scriptstyle (t)}\unaryminus\bw_{a}{\scriptstyle (t)})(\bw_{j}{\scriptstyle (t)}\unaryminus\bw_{a}{\scriptstyle (t)})^{\top}],
\end{align*} 
plus residual terms $\mathcal{O}(\frac{1}{m}\sum_{j=1}^{m}\|\bw_{j}{\scriptstyle (t)}\unaryminus\bw_{a}{\scriptstyle (t)}\|_2^3)$.
Here $\mH^{\mu_j{\scriptscriptstyle (t)}}_{\bw_{a}{\scriptstyle (t)}}\triangleq \frac{1}{|\mu_j{\scriptstyle (t)}|}\sum_{\zeta{\scriptstyle (t)}=1}^{|\mu_j{\scriptstyle (t)}|}\mH({\bw_{a}{\scriptstyle (t)}};z_{j, \zeta{\scriptstyle (t)}})$ denotes the empirical Hessian matrix evaluated at $\bw_{a}{\scriptstyle (t)}$ and $\mT^{\mu_j{\scriptscriptstyle (t)}}_{\bw_{a}{\scriptstyle (t)}} \triangleq \frac{1}{|\mu_j{\scriptstyle (t)}|}\sum_{\zeta{\scriptstyle (t)}=1}^{|\mu_j{\scriptstyle (t)}|}\mT({\bw_{a}{\scriptstyle (t)}};z_{j, \zeta{\scriptstyle (t)}})$ stacks for the empirical third-order partial derivative tensor at $\bw_{a}{\scriptstyle (t)}$, where $\mu_j{\scriptstyle (t)}$ and $z_{j, \zeta{\scriptstyle (t)}}$ follows the notation in \cref{def: c-sgd}.

% Analogous to the works investigating the SGD dynamics \citep{JMLR:v18:17-214, pmlr-v97-zhu19e, ziyin2022strength}, we then calculate the expectation of the gradient diversity.
% The expectation of gradient diversity is calculated first as follows.
As $\bw_{a}{\scriptstyle (t)}$ and local models $\bw_{j}{\scriptstyle (t)}\ (j\unaryequal1,\dots, m)$ are only correlated with the super batch before the $t$-th iteration (see \cref{def: dec-sgd}), taking expectation over $\mu{\scriptscriptstyle (t)}$ provides
\begin{align*}
\ebb_{\mu{\scriptscriptstyle (t)}}&\big[\frac{1}{m}\sum_{j=1}^{m}(\nabla \mL^{\mu_j{\scriptscriptstyle (t)}}_{\bw_{j}{\scriptstyle (t)}}\unaryminus\nabla \mL^{\mu_j{\scriptscriptstyle (t)}}_{\bw_{a}{\scriptstyle (t)}})\big]
= \mH_{\bw_{a}{\scriptstyle (t)}} \cdot \underbrace{\frac{1}{m}\sum_{j=1}^{m} (\bw_{j}{\scriptstyle (t)}\unaryminus\bw_{a}{\scriptstyle (t)})}_{= 0}
\unaryplus  \frac{1}{2}\mT_{\bw_{a}{\scriptstyle (t)}}\otimes\big[\frac{1}{m}\sum_{j=1}^{m} (\bw_{j}{\scriptstyle (t)}\unaryminus\bw_{a}{\scriptstyle (t)})(\bw_{j}{\scriptstyle (t)}\unaryminus\bw_{a}{\scriptstyle (t)})^{\top}\big],
\end{align*} 
plus residual terms $\mathcal{O}(\frac{1}{m}\sum_{j=1}^{m}\|\bw_{j}{\scriptstyle (t)}\unaryminus\bw_{a}{\scriptstyle (t)}\|_2^3)$, where $\mH_{\bw_{a}{\scriptstyle (t)}}\unaryequal\ebb_{\mu_j{\scriptscriptstyle (t)}} [\mH^{\mu_j{\scriptscriptstyle (t)}}_{\bw_{a}{\scriptstyle (t)}}]$ and $\mT_{\bw_{a}{\scriptstyle (t)}}\unaryequal\ebb_{\mu_j{\scriptscriptstyle (t)}} [\mT^{\mu_j{\scriptscriptstyle (t)}}_{\bw_{a}{\scriptstyle (t)}}]$.

The $i$-th entry of the above equation reads
\begin{align}
&\ebb_{\mu{\scriptscriptstyle (t)}}\big[\frac{1}{m}\sum_{j=1}^{m}(\partial_i \mL^{\mu_j{\scriptscriptstyle (t)}}_{\bw_{j}{\scriptstyle (t)}}\unaryminus\partial_i \mL^{\mu_j{\scriptscriptstyle (t)}}_{\bw_{a}{\scriptstyle (t)}})\big]\nonumber\\
&= \frac{1}{2}\underbrace{\sum_{l, s} \partial_{i l s }^3 \mL_{\bw_{a}{\scriptstyle (t)}} \frac{1}{m}\sum_{j=1}^{m}{(\bw_{j}{\scriptstyle (t)}\unaryminus\bw_{a}{\scriptstyle (t)})}_{l} {(\bw_{j}{\scriptstyle (t)}\unaryminus\bw_{a}{\scriptstyle (t)})}_s}_{\unaryequal \left[\partial_i \sum_{l s} \partial_{l s}^2 \mL_{\bw} \left(\frac{1}{m}\sum_{j\unaryequal1}^{m}{(\bw_{j}{\scriptscriptstyle (t)}\unaryminus\bw_{a}{\scriptscriptstyle (t)})}_{l} {(\bw_{j}{\scriptscriptstyle (t)}\unaryminus\bw_{a}{\scriptscriptstyle (t)})}_s\right)\right]|_{\bw\unaryequal\bw_{a}{\scriptscriptstyle (t)}}}
+ \mathcal{O}(\frac{1}{m}\sum_{j=1}^{m}\|\bw_{j}{\scriptstyle (t)}\unaryminus\bw_{a}{\scriptstyle (t)}\|_2^3 ),
\label{eq: taylor-expansion-appendix}
\end{align} 
where $(\bw_{j}{\scriptstyle (t)}\unaryminus\bw_{a}{\scriptstyle (t)})_l$ denotes the $l$-th entry of the vector $\bw_{j}{\scriptstyle (t)}\unaryminus\bw_{a}{\scriptstyle (t)}$. Details regarding the derivations are provided below.
\begin{align}
    &\sum_{l, s} \partial_{i l s }^3 \mL_{\bw_{a}{\scriptstyle (t)}} \frac{1}{m}\sum_{j=1}^{m}{(\bw_{j}{\scriptstyle (t)}\unaryminus\bw_{a}{\scriptstyle (t)})}_{l} {(\bw_{j}{\scriptstyle (t)}\unaryminus\bw_{a}{\scriptstyle (t)})}_s\nonumber\\
    &=\left(\sum_{l, s} {\frac{\partial^3 \mL({\bw})}{\partial (\bw)_i \partial (\bw)_l \partial (\bw)_s}}\Big|_{\bw={\bw_{a}{\scriptstyle (t)}}}\right)\cdot\frac{1}{m}\sum_{j=1}^{m}{(\bw_{j}{\scriptstyle (t)}\unaryminus\bw_{a}{\scriptstyle (t)})}_{l} {(\bw_{j}{\scriptstyle (t)}\unaryminus\bw_{a}{\scriptstyle (t)})}_s\nonumber\\
    &=\left(\sum_{l, s} {\frac{\partial^3 \mL({\bw})}{\partial (\bw)_i \partial (\bw)_l \partial (\bw)_s}}\cdot \frac{1}{m}\sum_{j=1}^{m}{(\bw_{j}{\scriptstyle (t)}\unaryminus\bw_{a}{\scriptstyle (t)})}_{l} {(\bw_{j}{\scriptstyle (t)}\unaryminus\bw_{a}{\scriptstyle (t)})}_s\right)\Big|_{\bw={\bw_{a}{\scriptstyle (t)}}}\nonumber\\
    &=\frac{\partial \sum_{l, s} {\frac{\partial^2 \mL({\bw})}{ \partial (\bw)_l \partial (\bw)_s}}\cdot \frac{1}{m}\sum_{j=1}^{m}{(\bw_{j}{\scriptstyle (t)}\unaryminus\bw_{a}{\scriptstyle (t)})}_{l} {(\bw_{j}{\scriptstyle (t)}\unaryminus\bw_{a}{\scriptstyle (t)})}_s }{\partial (\bw)_i}\Big|_{\bw={\bw_{a}{\scriptstyle (t)}}},
    \label{eq: Clairaut}
\end{align}
where $(\bw)_{i}, (\bw)_{l}$ and $(\bw)_{s}$ represent the $i$-th, $l$-th and $s$-th entry of the vector $\bw$, respectively. The third equality above is due to Clairaut’s theorem \citep{rudin1976principles}.

The RHS of the \cref{eq: Clairaut} without taking value $\bw_{a}{\scriptstyle (t)}$ is the $i$-th partial derivative of a quantity we term \textit{Hessian-consensus alignment},
\begin{align*}
    \Tr(\nabla^2 \mL_{\bw}\cdot\frac{1}{m}\sum_{j\unaryequal1}^{m}{(\bw_{j}{\scriptstyle (t)}\unaryminus\bw_{a}{\scriptstyle (t)})}_{l} {(\bw_{j}{\scriptstyle (t)}\unaryminus\bw_{a}{\scriptstyle (t)})}_s)
    = \sum_{l s} \partial_{l s}^2 \mL_{\bw} \cdot\frac{1}{m}\sum_{j\unaryequal1}^{m}{(\bw_{j}{\scriptstyle (t)}\unaryminus\bw_{a}{\scriptstyle (t)})}_{l} {(\bw_{j}{\scriptstyle (t)}\unaryminus\bw_{a}{\scriptstyle (t)})}_s.
    % \label{eq: Hessian-consensus-appendix}
\end{align*}
Denote $\bm{\Xi}{\scriptstyle (t)}\unaryequal\frac{1}{m}\sum_{j=1}^{m}(\bw_{j}{\scriptstyle (t)}\unaryminus\bw_{a}{\scriptstyle (t)}){(\bw_{j}{\scriptstyle (t)}\unaryminus\bw_{a}{\scriptstyle (t)})}^T$. The positive definiteness and symmetry of $\bm{\Xi}{\scriptstyle (t)}$ guarantees
\begin{align*}
    \Tr(\nabla^2 \mL_{\bw}\bm{\Xi}{\scriptscriptstyle (t)})
    &=\Tr(\nabla^2 \mL_{\bw}\ebb_{\epsilon \sim \mathcal{N}(0, \bm{\Xi}{\scriptscriptstyle (t)})}[\epsilon\epsilon^T])\nonumber\\
    &=\ebb_{\epsilon \sim \mathcal{N}(0, \bm{\Xi}{\scriptscriptstyle (t)})}[\epsilon^T\nabla^2 \mL_{\bw}\epsilon]\nonumber\\
    &= \underbrace{\ebb_{\epsilon \sim \mathcal{N}(0, \bm{\Xi}{\scriptscriptstyle (t)})}[2(\mL_{\bw+\epsilon}\unaryminus\mL_{\bw})}_{\text{\textit{average-direction sharpness at} $\bw$}}\unaryplus\mathcal{O}{(\|\epsilon\|_2^3)}],
\end{align*}
where the last equality is due to a second-order Taylor expansion of $\mL_{\bw+\epsilon}$ around $\bw$.

Therefore, we arrive at
\begin{align}
    \left(\nabla\Tr(\nabla^2 \mL_{\bw}\bm{\Xi}{\scriptscriptstyle (t)})\right)|_{\bw=\bw_{a}{\scriptstyle (t)}}
    = \underbrace{\ebb_{\epsilon \sim \mathcal{N}(0, \bm{\Xi}{\scriptscriptstyle (t)})}[2(\nabla\mL_{\bw_{a}{\scriptstyle (t)}+\epsilon}\unaryminus\nabla\mL_{\bw_{a}{\scriptstyle (t)}})}_{\text{\textit{average-direction sharpness at} $\bw_{a}{\scriptstyle (t)}$}}\unaryplus\mathcal{O}{(\|\epsilon\|_2^3)}].
\label{eq: dsgd-sam-appendix}
\end{align}

Combining \cref{eq: dsgd-sam-appendix} and \cref{eq: graident-decomposition-appendix} gives
\begin{align*}
\ebb_{\mu{\scriptscriptstyle (t)}}&[\bw_{a}{\scriptstyle (t+1)}]
=\bw_{a}{\scriptstyle (t)}-\eta\nabla\underbrace{\ebb_{\epsilon \sim \mathcal{N}(0, \bm{\Xi}{\scriptstyle (t)})}[\mL_{\bw_{a}{\scriptstyle (t)}+\epsilon}]}_{\text{\textit{asymptotic descent direction}}}
\unaryplus\underbrace{\mathcal{O}(\eta\ \ebb_{\epsilon \sim \mathcal{N}(0, \bm{\Xi}{\scriptstyle (t)})}\|\epsilon\|_2^3\unaryplus\frac{\eta}{m}\sum_{j=1}^{m}\|\bw_{j}{\scriptstyle (t)}\unaryminus\bw_{a}{\scriptstyle (t)}\|_2^3 )}_{\text{\textit{higher-order residual terms}}},
\end{align*} 
which completes the proof.

{\color{magenta}\qed}

\textbf{Asymptotic equivalence.} Asymptotic equivalence is a fundamental concept in mathematics which describes the equivalent behavior of functions as their inputs approach a limit point \citep{erdelyi1956asymptotic, de1981asymptotic}. Two functions are said to be asymptotically equivalent if their ratio approaches $1$ as the input approaches the limit point. The asymptotic equivalence allows us to simplify complex functions and identify functions that exhibit the same limiting behavior. In the following, we define a new asymptotic equivalence called relative asymptotic equivalence, which characterizes the limiting equivalent behavior of two functions with respect to certain conditions.

\begin{definition}[Conditional asymptotic equivalence\label{def: asymptotic-equivalence}] 
Let $f(u_{1}, \cdots, u_{m})$ and $g(u_{1}, \cdots, u_{m})$ be two multivariate functions of $u_{1}, \cdots, u_{m}$, where $m$ is an arbitrary positive integer. $f(u_{1}, \cdots, u_{m})$ and $g(u_{1}, \cdots, u_{m})$ are said to be conditional asymptotically equivalent with respect to the limiting condition(s) $C$ if and only if
\begin{align}
    \lim_{C} \frac{f(u_{1}, \cdots, u_{m})}{g(u_{1}, \cdots, u_{m})} = 1.
\end{align}
\end{definition}
The conditional asymptotic equivalence is a direct extension of the original asymptotic equivalence, where the limiting conditions are only on the original variables $u_{1}, \cdots, u_{m}$.

According to \cref{def: asymptotic-equivalence}, $\eta\cdot \ebb_{\epsilon \sim \mathcal{N}(0, \bm{\Xi}{\scriptstyle (t)})}[\nabla\mL_{\bw_{a}{\scriptstyle (t)}+\epsilon}]$ and $\mathcal{O}(\eta\ \ebb_{\epsilon \sim \mathcal{N}(0, \bm{\Xi}{\scriptstyle (t)})}\|\epsilon\|_2^3+\frac{\eta}{m}\sum_{j=1}^{m}\|\bw_{j}{\scriptstyle (t)}-\bw_{a}{\scriptstyle (t)}\|_2^3)$ in \cref{th: dsgd-sam} are conditional asymptotically equivalent with respect to the condition $(\bw_{j}{\scriptstyle (t)}-\bw_{a}{\scriptstyle (t)})\to 0, \forall j=1,\dots, m$, as
\begin{align}\label{eq: asymptotic-equivalence}
    &\lim_{\substack{(\bw_{j}{\scriptscriptstyle (t)}\unaryminus\bw_{a}{\scriptscriptstyle (t)})\unaryrightarrow 0 \\ \forall j=1,\dots, m}} \frac{\eta\left[\ebb_{\epsilon \sim \mathcal{N}(0, \bm{\Xi}{\scriptstyle (t)})}[\nabla\mL_{\bw_{a}{\scriptstyle (t)}\unaryplus\epsilon}]\unaryminus\nabla\mL_{\bw_{a}{\scriptstyle (t)}}\unaryplus\nabla\mL_{\bw_{a}{\scriptstyle (t)}}\right]\unaryplus\mathcal{O}(\eta\ \ebb_{\epsilon \sim \mathcal{N}(0, \bm{\Xi}{\scriptstyle (t)})}\|\epsilon\|_2^3\unaryplus\frac{\eta}{m}\sum_{j=1}^{m}\|\bw_{j}{\scriptstyle (t)}\unaryminus\bw_{a}{\scriptstyle (t)}\|_2^3)}{\eta\left[\ebb_{\epsilon \sim \mathcal{N}(0, \bm{\Xi}{\scriptstyle (t)})}[\nabla\mL_{\bw_{a}{\scriptstyle (t)}\unaryplus\epsilon}]\unaryminus\nabla\mL_{\bw_{a}{\scriptstyle (t)}}\unaryplus\nabla\mL_{\bw_{a}{\scriptstyle (t)}}\right]}\nonumber\\
    & = \lim_{\substack{(\bw_{j}{\scriptscriptstyle (t)}\unaryminus\bw_{a}{\scriptscriptstyle (t)})\unaryrightarrow 0 \\ \forall j=1,\dots, m}}\frac{\frac{\mu_1}{m}\sum_{j=1}^{m}\|\bw_{j}{\scriptstyle (t)}-\bw_{a}{\scriptstyle (t)}\|_2^2+\nabla\mL_{\bw_{a}{\scriptstyle (t)}}+\frac{\mu_2}{m}\sum_{j=1}^{m}\|\bw_{j}{\scriptstyle (t)}-\bw_{a}{\scriptstyle (t)}\|_2^3}{\frac{\mu_1}{m}\sum_{j=1}^{m}\|\bw_{j}{\scriptstyle (t)}-\bw_{a}{\scriptstyle (t)}\|_2^2+\nabla\mL_{\bw_{a}{\scriptstyle (t)}}}\nonumber\\
    & = 1.
\end{align}
The first equality above is obtained by equivalent infinitesimal substitutions, where
\begin{align}
    \mu_1=\lim_{\substack{(\bw_{j}{\scriptscriptstyle (t)}\unaryminus\bw_{a}{\scriptscriptstyle (t)})\unaryrightarrow 0 \\ \forall j=1,\dots, m}} \frac{\ebb_{\epsilon \sim \mathcal{N}(0, \bm{\Xi}{\scriptstyle (t)})}[\nabla\mL_{\bw_{a}{\scriptstyle (t)}+\epsilon}]-\nabla\mL_{\bw_{a}{\scriptstyle (t)}}}{\frac{1}{m}\sum_{j=1}^{m}\|\bw_{j}{\scriptstyle (t)}-\bw_{a}{\scriptstyle (t)}\|_2^2}\in\rbb \text{ and}
\end{align}
\begin{align}
    \mu_2=\lim_{\substack{(\bw_{j}{\scriptscriptstyle (t)}\unaryminus\bw_{a}{\scriptscriptstyle (t)})\unaryrightarrow 0 \\ \forall j=1,\dots, m}} \frac{\mathcal{O}(\eta\ \ebb_{\epsilon \sim \mathcal{N}(0, \bm{\Xi}{\scriptstyle (t)})}\|\epsilon\|_2^3+\frac{\eta}{m}\sum_{j=1}^{m}\|\bw_{j}{\scriptstyle (t)}-\bw_{a}{\scriptstyle (t)}\|_2^3)}{\frac{1}{m}\sum_{j=1}^{m}\|\bw_{j}{\scriptstyle (t)}-\bw_{a}{\scriptstyle (t)}\|_2^3}\in\rbb
\end{align}
depend only on the higher-order geometric information of $\mL$ and are independent of $\bw_{j}{\scriptstyle (t)}-\bw_{a}{\scriptstyle (t)}, \forall j=1,\dots, m$.

According to \cref{eq: asymptotic-equivalence}, the expected gradient of D-SGD is asymptotically equivalent to $\ebb_{\epsilon \sim \mathcal{N}(0, \bm{\Xi}{\scriptstyle (t)})}[\nabla\mL_{\bw_{a}{\scriptstyle (t)}+\epsilon}]$, the gradient direction of an average-direction SAM. Note that in the limit when $\bw_{j}{\scriptstyle (t)}-\bw_{a}{\scriptstyle (t)}=0, \forall j=1,\dots, m$, D-SGD reduces to the standard SGD, which can be viewed as a specific instance of the average-direction SAM with $\epsilon=0$.

\subsection{Proof of \cref{coro: smoothing}\label{sec: smoothing}}
\begin{tcolorbox}[notitle, rounded corners, colframe=middlegrey, colback=lightblue, 
       boxrule=2pt, boxsep=0pt, left=0.15cm, right=0.17cm, enhanced, 
       toprule=2pt,
    ]
\textbf{Corollary 2} \textit{(Gradient smoothing effect of D-SGD). Given that vanilla loss function $\mL_{\bw}$ is $\alpha$-Lipschitz continuous and the gradient $\nabla\mL_{\bw}$ is $\beta$-Lipschitz continuous. We conclude that the gradient $\nabla\mL_{\bw+\epsilon}$ where $\epsilon\sim\mathcal{N}(0,\bm{\Xi}{\scriptstyle (t)})$ is $\min \left\{\frac{\sqrt{2}\alpha}{\sigma_{\textnormal{min}}}, \beta\right\}$-Lipschitz continuous, where $\sigma_{\textnormal{min}}$ denotes the smallest eigenvalue of $\bm{\Xi}{\scriptstyle (t)}$.}
\end{tcolorbox}

\cref{coro: smoothing} is an extension of Theorem 1 in \citep{bisla2022low}. To prove \cref{coro: smoothing}, we recall  Lemma 1 in \citep{bisla2022low}.

\begin{lemma}[\citep{bisla2022low}\label{lemma: bound-loss-perturb}]
Given that vanilla loss function $\mL_{\bw}$ is $\alpha$-Lipschitz continuous, $\forall \bw_1, \bw_2 \in \mathbb{R}^d$, we have
\begin{align}
 \|\mathbb{E}_{\epsilon\sim\mathcal{N}(0,\bm{\Xi}{\scriptstyle (t)})}\left[\nabla \mL_{\bw_1+\epsilon}-\nabla \mL_{\bw_2+\epsilon}\right]\|_2
\leq \alpha \int|p(\epsilon-\bw_1)-p(\epsilon-\bw_2)| d \epsilon,\label{eq: bound-loss-perturb}
\end{align}  
\end{lemma}
where $p$ denotes the density function of $\mathcal{N}(0,\bm{\Xi}{\scriptstyle (t)})$.

\textit{Proof of \cref{coro: smoothing}.}

(1) Proving the gradient $\nabla\mL_{\bw+\epsilon}$ is $ \frac{\sqrt{2}\alpha}{\sigma_{\textnormal{min}}}$-Lipschitz continuous.

We start by writing the RHS of \cref{eq: bound-loss-perturb} as
\begin{align*}
    &\int|p(\epsilon-\bw_1)-p(\epsilon-\bw_2)| d \epsilon\leq\\
 & =\int_{\epsilon:\|\epsilon-\bw_1\|_2 \geq\|\epsilon-\bw_2\|_2}[p(\epsilon-\bw_1)-p(\epsilon-\bw_2)] d \epsilon+\int_{\epsilon:\|\epsilon-\bw_1\|_2 \leq\|\epsilon-\bw_2\|_2}[p(\epsilon-\bw_1)-p(\epsilon-\bw_2)] d \epsilon \\
& =2 \int_{\epsilon:\|\epsilon-x\|_2 \leq\|\epsilon-\bw_2\|_2}[p(\epsilon-\bw_1)-p(\epsilon-\bw_2)] d \epsilon.\\  
\end{align*}
The change of variables $\hat{\epsilon}=\epsilon-\bw_1$ for $p(\epsilon-\bw_1)$ and $\hat{\epsilon}=\epsilon-\bw_2$ for $p(\epsilon-\bw_2)$ gives
\begin{align*}
     \int_{\epsilon:\|\epsilon-x\|_2 \leq\|\epsilon-\bw_2\|_2}[p(\epsilon-\bw_1)-p(\epsilon-\bw_2)] d \epsilon
     &= w\int_{\epsilon:\|\hat{\epsilon} \mid \leq\| \hat{\epsilon}+(x-y) \|} p(\hat{\epsilon}) d \hat{\epsilon}
     -2 \int_{\epsilon:\|\hat{\epsilon}\| \geq\|\hat{\epsilon}-(x-y)\|} p(\hat{\epsilon}) d \hat{\epsilon}\\
     &=2 \mathbb{P}_{\hat{\epsilon} \sim p}(\|\hat{\epsilon}\| \leq\|\hat{\epsilon}+(\bw_1-\bw_2)\|)-2 \mathbb{P}_{\hat{\epsilon} \sim p}(\|\hat{\epsilon}\| \geq\|\hat{\epsilon}-(\bw_1-\bw_2)\|). 
\end{align*}
The first part in the RHS of the above equality reads,
\begin{align*}
 \mathbb{P}_{\hat{\epsilon} \sim p}(\|\hat{\epsilon}\| \leq\|\hat{\epsilon}+(\bw_1-\bw_2)\|) 
&=  \mathbb{P}_{\hat{\epsilon} \sim p}\left(\|\hat{\epsilon}\|^2 \leq\|\hat{\epsilon}+(\bw_1-\bw_2)\|^2\right) \\
&=  \mathbb{P}_{\hat{\epsilon} \sim p}\left(2\langle\hat{\epsilon}, \bw_1-\bw_2\rangle \geq-\|\bw_1-\bw_2\|^2\right) \\
&=  \mathbb{P}_{\hat{\epsilon} \sim p}\left(2\left\langle\hat{\epsilon}, \frac{\bw_1-\bw_2}{\|\bw_1-\bw_2\|}\right\rangle \geq-\|\bw_1-\bw_2\|\right) .
\end{align*}
According to the fact that $\|\frac{\bw_1-\bw_2}{\|\bw_1-\bw_2\|_2}\|_2=1$, which implies $\left\langle\hat{\epsilon}, \frac{\bw_1-\bw_2}{\|\bw_1-\bw_2\|_2}\right\rangle \sim \mathcal{N}\left(0, \frac{{(\bw_1-\bw_2)}^{\top}\bm{\Xi}{\scriptstyle (t)}{(\bw_1-\bw_2)}}{\|\bw_1-\bw_2\|^2_2}\right)$, we obtain,
\begin{align*}
& \mathbb{P}_{\hat{\epsilon} \sim p}(\|\hat{\epsilon}\| \leq\|\hat{\epsilon}+(\bw_1-\bw_2)\|) \\
= & \mathbb{P}_{\hat{\epsilon} \sim p}\left(\left\langle\hat{\epsilon}, \frac{\bw_1-\bw_2}{\|\bw_1-\bw_2\|_2}\right\rangle \geq-\frac{\|\bw_1-\bw_2\|_2}{2}\right) \\
= & \int_{-\frac{\|\bw_1-\bw_2\|_2}{2}}^{+\infty} \frac{1}{\sqrt{2 \pi }}\frac{\|\bw_1-\bw_2\|_2}{{(\bw_1-\bw_2^2)}^{\top}\bm{\Xi}{\scriptstyle (t)}{(\bw_1-\bw_2)}} \exp \left(-\frac{\hat{\epsilon}^2\|\bw_1-\bw_2^2\|_2}{2{(\bw_1-\bw_2)}^{\top}\bm{\Xi}{\scriptstyle (t)}{(\bw_1-\bw_2)}}\right) d \hat{\hat{\epsilon}}.
\end{align*}
By some token, we have,
\begin{align*}
& \mathbb{P}_{\hat{\epsilon} \sim p}(\|\hat{\epsilon}\| \leq\|\hat{\epsilon}-(\bw_1-\bw_2)\|) \\
= & \int_{\frac{\|\bw_1-\bw_2\|_2}{2}}^{+\infty} \frac{1}{\sqrt{2 \pi }}\frac{\|\bw_1-\bw_2\|_2^2}{{(\bw_1-\bw_2)}^{\top}\bm{\Xi}{\scriptstyle (t)}{(\bw_1-\bw_2)}} \exp \left(-\frac{\hat{\epsilon}^2\|\bw_1-\bw_2\|_2^2}{2{(\bw_1-\bw_2)}^{\top}\bm{\Xi}{\scriptstyle (t)}{(\bw_1-\bw_2)}}\right) d \hat{\hat{\epsilon}}.    
\end{align*}
Combining the two inequalities gives
\begin{align*}
    \mathbb{P}_{\hat{\epsilon} \sim p}&(\|\hat{\epsilon}\| \leq\|\hat{\epsilon}+(\bw_1-\bw_2)\|) \\
    & =\int_{-\frac{\|\bw_1-\bw_2\|_2}{2}}^{\frac{\|\bw_1-\bw_2\|_2}{2}} \frac{\sqrt{2}}{\sqrt{\pi }}\frac{\|\bw_1-\bw_2\|^2_2}{{(\bw_1-\bw_2)}^{\top}\bm{\Xi}{\scriptstyle (t)}{(\bw_1-\bw_2)}} \exp \left(-\frac{\hat{\epsilon}^2\|\bw_1-\bw_2\|^2_2}{2{(\bw_1-\bw_2)}^{\top}\bm{\Xi}{\scriptstyle (t)}{(\bw_1-\bw_2)}}\right) d \hat{\hat{\epsilon}}.
\end{align*}
Since $ \exp \left(-\frac{\hat{\epsilon}^2\|\bw_1-\bw_2\|_2}{2{(\bw_1-\bw_2)}^{\top}\bm{\Xi}{\scriptstyle (t)}{(\bw_1-\bw_2)}}\right) \leq 1$, we can write
\begin{align*}
    \mathbb{P}_{\hat{\epsilon} \sim p}&(\|\hat{\epsilon}\| \leq\|\hat{\epsilon}+(\bw_1-\bw_2)\|) 
     \leq \frac{\sqrt{2}}{\sqrt{\pi }} \frac{\|\bw_1-\bw_2\|^2_2}{{(\bw_1-\bw_2)}^{\top}\bm{\Xi}{\scriptstyle (t)}{(\bw_1-\bw_2)}} \|\bw_1-\bw_2\|_2.
\end{align*}
The goal then becomes upper bounding $\frac{\|\bw_1-\bw_2\|^2_2}{{(\bw_1-\bw_2)}^{\top}\bm{\Xi}{\scriptstyle (t)}{(\bw_1-\bw_2)}}$ with a constant, as shown below:
\begin{align*}
    \frac{\|\bw_1-\bw_2\|^2_2}{{(\bw_1-\bw_2)}^{\top}\bm{\Xi}{\scriptstyle (t)}{(\bw_1-\bw_2)}}
    &=\frac{\|\bw_1-\bw_2\|^2_2}{\Tr{\left({(\bw_1-\bw_2)}^{\top}\bm{\Xi}{\scriptstyle (t)}{(\bw_1-\bw_2)}\right)}}\\
    &=\frac{\|\bw_1-\bw_2\|^2_2}{\Tr{\left(\bm{\Xi}{\scriptstyle (t)}{(\bw_1-\bw_2)}{(\bw_1-\bw_2)}^{\top}\right)}}\\
    &\leq \frac{\|\bw_1-\bw_2\|^2_2}{\sigma_{\textnormal{min}}\Tr{({(\bw_1-\bw_2)}{(\bw_1-\bw_2)}^{\top})}}\\
    &=\frac{1}{\sigma_{\textnormal{min}}},
\end{align*}
where the second equality uses the cyclic property of trace and the inequality is due to Von Neumann's trace inequality \citep{von1937some}.

Therefore, we can prove that the gradient $\nabla\mL_{\bw+\epsilon}$ is $ \frac{\sqrt{2}\alpha}{\sigma_{\textnormal{min}}}$-Lipschitz continuous by \cref{lemma: bound-loss-perturb}.

(2) Proving the gradient $\nabla\mL_{\bw+\epsilon}$ is $\beta$-Lipschitz continuous.
\begin{align*}
    \|\mathbb{E}_{\epsilon\sim\mathcal{N}(0,\bm{\Xi}{\scriptstyle (t)})}\left[\nabla \mL_{\bw_1+\epsilon}-\nabla \mL_{\bw_2+\epsilon}\right]\|_2
    &\leq \|\int(\nabla \mL_{\bw_1+\epsilon}-\nabla \mL_{\bw_2+\epsilon})\ p(\epsilon)\ d \epsilon\|_2\\
    &\leq \int \|\nabla \mL_{\bw_1+\epsilon}-\nabla \mL_{\bw_2+\epsilon}\|_2\ p(\epsilon)\ d \epsilon\\
    &\leq \beta \|\bw_1-\bw_2\|_2 \int p(\epsilon) d \epsilon\\
    &= \beta \|\bw_1-\bw_2\|_2.
\end{align*}

The proof is now complete.

{\color{magenta}\qed}

\newpage
\subsection{Proof of \cref{th: large-batch}\label{sec:proof-large-batch}}

\begin{tcolorbox}[notitle, rounded corners, colframe=middlegrey, colback=lightblue, 
       boxrule=2pt, boxsep=0pt, left=0.15cm, right=0.17cm, enhanced, 
       toprule=2pt,
    ]
\textbf{Theorem 4.} \textit{Let us denote $B=|\mu|$ as the total batch size. With a probability greater than $1-\mathcal{O}(\frac{B}{(N-B)\eta^2})$, D-SGD implicit minimizes the following objective function:
\begin{align*}
    &\mL_{\bw}^{\text{D-SGD}}\nonumber\\
    &\unaryapprox\mL_{\bw}^\mu \unaryplus\underbrace{\Tr(\mH^\mu_\bw \bm{\Xi}{\scriptstyle (t)})\unaryplus\frac{\eta}{4}\Tr({(\mH^\mu_\bw)}^2 \bm{\Xi}{\scriptstyle (t)})}_{\text{batch size independent sharpness regularizer}}
    \unaryplus \frac{\eta}{4}\|\nabla\mL_{\bw}^\mu\|_2^2 
    \unaryplus\underbrace{\frac{\kappa}{N}\sum_{j=1}^{N}\left[\|\nabla\mL_{\bw}^j\unaryminus \nabla\mL_{\bw}^\mu\|_2^2 \unaryplus
    \Tr({(\mH_\bw^j\unaryminus\mH_\bw^\mu)}^2\bm{\Xi}{\scriptstyle (t)})\right]}_{\text{batch size dependent variance regularizer}}
    \unaryplus\mathcal{R}^{A}\unaryplus\mathcal{O}{(\eta^2)},
\end{align*}
where $\kappa=\frac{\eta}{B}\cdot\frac{N\unaryminus B}{(N\unaryminus1)}$, and $\mathcal{R}^{A}$ absorbs all higher-order residuals. The empirical loss and the gradient on the super-batch $\mu$, denoted by $\mL^{\mu}_{\bw}$ and $ \nabla\mL^{\mu}_{\bw}$, respectively, are calculated as averages over the one-sample gradients $\nabla \mL_{\bw}^j$. Similarly, the empirical Hessian $\mH_\bw^\mu$ is an averages of $\mH_{\bw}^j=\mH(\bw;z_j)$.}
\end{tcolorbox}

To prove \cref{th: large-batch}, we start by showing how the gradient variance and the total batch size affect the training dynamics of the distributed central SGD.

\begin{tcolorbox}[notitle, rounded corners, colframe=middlegrey, colback=lightblue, 
       boxrule=2pt, boxsep=0pt, left=0.15cm, right=0.17cm, enhanced, 
       toprule=2pt,
    ]
\begin{lemma}\label{lemma: implicit-reg-sgd}
    Recall that $N$ denotes the total training sample size and $\eta$ denotes the learning rate. Let us denote $B=|\mu|$ as the total batch size. With a probability greater than $1-\mathcal{O}(\frac{B}{(N-B)\eta^2})$, distributed centralized SGD (see \cref{def: c-sgd}) implicitly minimizes the following objective function:
\begin{align}
    \mL_{\bw}^{\text{C-SGD}}\approx\underbrace{\ \mL_{\bw}^{\mu}\ }_{\text{original loss}}+\frac{\eta}{4}\cdot\underbrace{\ \|\nabla \mL_{\bw}^{\mu}\|_2^2\ }_{\text{magnitude of gradient}}
    + \frac{\eta}{B}\cdot\frac{N-B}{(N-1)}\cdot \underbrace{\frac{1}{N}\sum_{j=1}^{N}\|\nabla \mL_{\bw}^j- \nabla\mL^{\mu}_{\bw}\|_2^2}_{\text{variance of one-sample gradients}} +\mathcal{O}{(\eta^2)},\label{eq: implicit-reg-sgd}
\end{align}
where the empirical loss and gradient on the super-batch $\mu$, denoted by $ \mL^{\mu}_{\bw}=\frac{1}{N}\sum_{\zeta=1}^{N} \mL(\bw;z_\zeta)$ and $ \nabla\mL^{\mu}_{\bw}=\frac{1}{N}\sum_{\zeta=1}^{N} \nabla\mL(\bw;z_\zeta)$, respectively, are calculated as averages over the one-sample gradients, represented by $\nabla \mL_{\bw}^j=\nabla\mL(\bw;z_j)$, at each sample $z_j$.
\end{lemma}
\end{tcolorbox}

\textbf{Remark.} \cref{lemma: implicit-reg-sgd} demonstrates that the ``true" loss function which C-SGD (or standard SGD) optimizes is closely tracked by the original loss plus the magnitude of averaged gradient and a constant times the variance of one-sample gradients. The last term, i.e., the total variance of one-sample gradient, serves as a measure of generalizability.

\textbf{Intuition of the generalization advantage of low gradient variance.}
To intuitively explain the regularization effect of $\frac{1}{N}\sum_{j=1}^{N}\|\nabla \mL_{\bw}^j- \mL^{\mu}_{\bw}\|_2^2$, the empirical variance of gradient, we make a cartoon illustration of the loss function near two minima: the left one has low gradient variance and the right one has high gradient variance\footnote{The illustration is inspired by a blog named ``Notes on the origin of implicit regularization in SGD" written by Ferenc Huszár.}.

\begin{figure*}[h!]
\centering
\begin{subfigure}[ResNet-18 on CIFAR-10 (C-SGD), 128 total batch size]{.39\textwidth}
  \centering
  % include fourth image
  \includegraphics[width=1.0\linewidth]{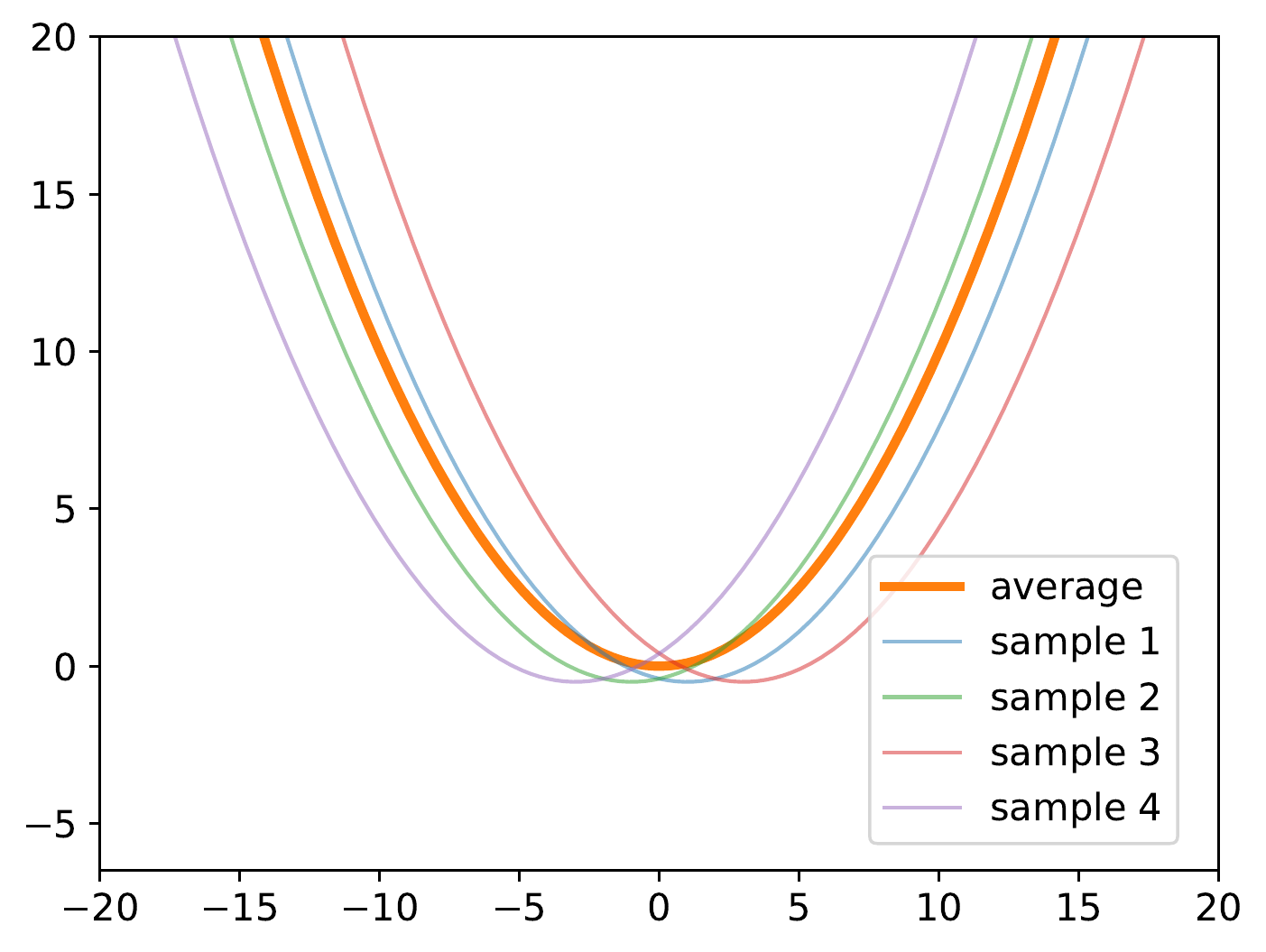}  
\end{subfigure}
\begin{subfigure}[ResNet-18 on CIFAR-10 (C-SGD), 1024 total batch size]{.39\textwidth}
  \centering
  % include fourth image
  \includegraphics[width=1.0\linewidth]{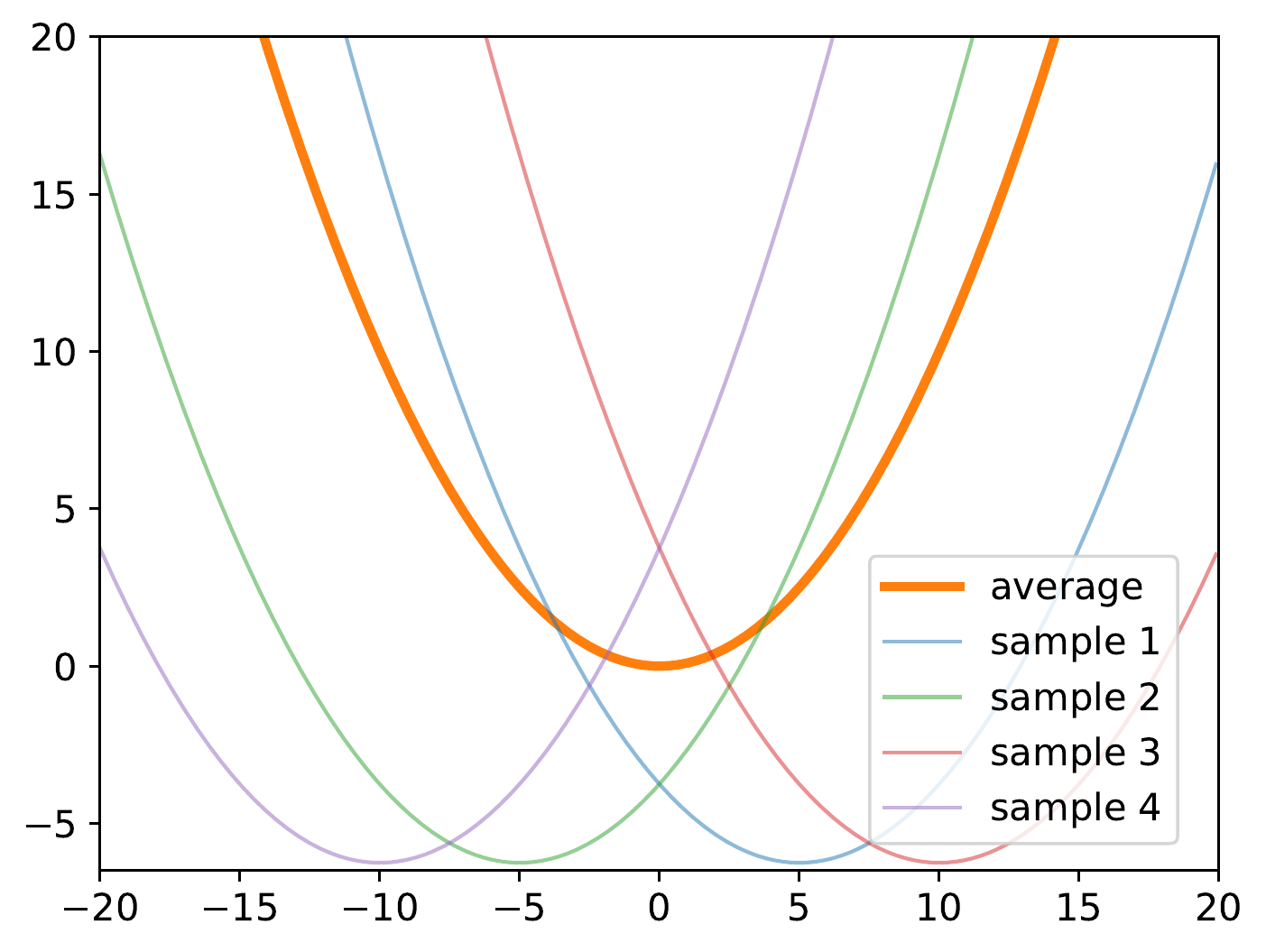}  
\end{subfigure}
\caption{An illustration of minima with low gradient variance (left) and high gradient variance (right)}
\label{fig: gradient-variance}
\end{figure*}

The figures depict loss functions with identical average empirical loss, yet the one-sample losses are more tightly grouped in the left figure, and dispersed in the right figure.
\cref{fig: gradient-variance} illustrates that the loss function with lower gradient variance exhibits a lower sensitivity with respect to the specific sample used for evaluation. Therefore, the minima characterized by lower gradient variance may exhibit consistency in their performance when evaluated on unseen validation samples, which guarantees good generalization performance.

\cref{lemma: implicit-reg-sgd} proves that there exists an implicit regularization effect on the variance of the one-sample gradient in SGD to improve generalization performance. It is noteworthy that as the total batch size $B$ approaches the total training sample size $N$, the regularization term diminishes rapidly, as the ratio $\frac{N-B}{N-1}$ approaches $0$ asymptotically.

\textit{Proof of \cref{lemma: implicit-reg-sgd}}.

We recall the true loss function approximation of single worker SGD proposed by \citet{smith2021on}:
\begin{align}
    \mL_{\bw}^{\text{SGD}}=\ \mL_{\bw}^{\mu}\ +\frac{\eta}{4}\cdot\|\nabla \mL_{\bw}^{\mu}\|_2^2\ 
    + \eta\cdot\underbrace{\frac{1}{mB^2}\sum_{i=0}^{m-1}\|\sum_{j=iB+1}^{iB+B}\nabla \mL_{\bw}^j- \mL^{\mu}_{\bw}\|_2^2}_{\text{variance of mini-batch gradients}}.\label{eq: sgd-approx}
\end{align}

Denote $V_j=\nabla\mL_{\bw}^j-\nabla\mL_{\bw}^{\mu}$. The total variance of the mini-batch gradient in \cref{eq: sgd-approx} can be  written as follows:
\begin{align*}
    \frac{1}{mB^2}\sum_{i=0}^{m-1}\|\sum_{j=iB+1}^{iB+B}V_j\|_2^2
    &=\frac{1}{mB^2}\sum_{i=0}^{m-1}\sum_{j_1=iB+1}^{iB+B}\sum_{j_1=iB+1}^{iB+B}V_{j_1}^{\top} V_{j_2}\nonumber\\
    & = \frac{1}{mB^2}\sum_{i=0}^{m-1}\sum_{j_1=iB+1}^{iB+B}(V_{j_1}^{\top} V_{j_1}
    + \sum_{\substack{j_2=iB+1\\j_2\neq j_1}}^{iB+B}V_{j_1}^{\top} V_{j_2}).
\end{align*}
The first part of the equation can be trivially written as $\frac{1}{mB^2}\sum_{i=0}^{m-1}\sum_{j_1=iB+1}^{iB+B}V_{j_1}^{\top} V_{j_1}=\frac{1}{NB}\sum_{j_1=1}^{N}V_{j_1}^{\top} V_{j_1}$. To  tackle the remaining tricky part, we start by showing that
\begin{align}
     \frac{1}{mB^2}\sum_{i=0}^{m-1}\sum_{j_1=iB+1}^{iB+B}\sum_{\substack{j_2=iB+1\\j_2\neq j_1}}^{iB+B}V_{j_1}^{\top} V_{j_2}
     &= \frac{1}{B^2} \sum_{j_1=1}^{B}\sum_{\substack{j_2=1\\j_2\neq j_1}}^{B}(\frac{1}{m}\sum_{i=0}^{m-1} V_{iB+j_1}^{\top} V_{iB+j_2}). \label{eq: off-diagonal-v}
\end{align}
We then approximate $\frac{1}{m}\sum_{i=0}^{m-1} V_{iB+j_1}^{\top} V_{iB+j_2}$ using its expectation:
\begin{align*}
    \frac{1}{m}\sum_{i=0}^{m-1} V_{iB+j_1}^{\top} V_{iB+j_2}
    = \ebb_i[V_{iB+j_1}^{\top} V_{iB+j_2}] + \underbrace{\left(\frac{1}{m}\sum_{i=0}^{m-1} V_{iB+j_1}^{\top} V_{iB+j_2}-\ebb_i[V_{iB+j_1}^{\top} V_{iB+j_2}]\right)}_{\text{bias estimator}},
\end{align*}
where $\ebb_i$ eliminates the randomness of the sample $z_i$.

According to the Chebyshev’s inequality \citep{chebyshev1874valeurs,marshall1960multivariate}, with a probability greater than $1-\frac{\sigma^2}{(m-1)\eta^2}=1-\frac{B\sigma^2}{(N-B)\eta^2}$\footnote{For sufficient large total sample size $N=\Omega(B(1+\eta^{-2}))$, the probability $1-\frac{B\sigma^2}{N\eta^2}\approx 1$. For large total batch size $B$, if we apply the linear scaling rule when increases $B$ (i.e., $\frac{\eta}{B}=s$ where $s$ is a constant) then the probability becomes $1-\frac{\sigma^2}{(N-B)Bs^2}$.}, it holds that,
\begin{align*}
    \frac{1}{m}\sum_{i=0}^{m-1} V_{iB+j_1}^{\top} V_{iB+j_2}
    = \ebb_i[V_{iB+j_1}^{\top} V_{iB+j_2}] + \mathcal{O}(\eta), \label{eq: mean-approximation}
\end{align*}
where $\sigma^2$ is the variance of random variables $V_{iB+j_1}^{\top} V_{iB+j_2}\ (i= 1,\dots,m-1)$. 

Substituting the sample mean approximation into \cref{eq: off-diagonal-v}, we have that, with a probability greater than $1-\frac{B\sigma^2}{(N-B)\eta^2}$,
\begin{align}
    \frac{1}{mB^2}\sum_{i=0}^{m-1}\sum_{j_1=iB+1}^{iB+B}\sum_{\substack{j_2=iB+1\\j_2\neq j_1}}^{iB+B}V_{j_1}^{\top} V_{j_2}
    &= \frac{B(B-1)}{B^2} \left(\ebb_i[V_{iB+j_1}^{\top} V_{iB+j_2}(1-\delta_{j_1, j_2})]+\mathcal{O}{(\eta)}\right)\nonumber\\
    &= \frac{B(B-1)}{B^2} \left(\frac{1}{N(N-1)}\sum_{i_1=0}^{N}\sum_{\substack{i_2=0\\i_1\neq i_2}}^{N} V_{i_1}^{\top} V_{i_2}+\mathcal{O}{(\eta)}\right),
\end{align}
where $\delta_{j_1, j_2}$ denotes the Kronecker delta function. The derivation of the second equality is reminiscent of \cref{eq: mean-approximation},  with a sight change of the indices.

Due to the fact that
\begin{align*}
    \frac{1}{N(N-1)}\sum_{i_1=0}^{N}\sum_{i_2=0}^{N} V_{i_1}^{\top} V_{i_2}&=\frac{1}{N-1}\|\frac{1}{N}\sum_{i_1=1}^N\nabla\mL_{\bw}^{i_1}-\nabla\mL_{\bw}^{\mu}\|_2^2\nonumber\\
    &=\frac{1}{N-1}\|\nabla\mL_{\bw}^{\mu}-\nabla\mL_{\bw}^{\mu}\|_2^2\nonumber\\
    &=0,
\end{align*}
with a probability greater than $1-\frac{B\sigma^2}{(N-B)\eta^2}$, the total variance of mini-batch gradient in \cref{eq: sgd-approx} reads,
\begin{align*}
    \frac{1}{mB^2}\sum_{i=0}^{m-1}\|\sum_{j=iB+1}^{iB+B}V_j\|_2^2
    &=\frac{1}{NB}\sum_{j_1=1}^{N}V_{j_1}^{\top} V_{j_1}
    - \frac{(B-1)}{NB(N-1)} \left(\sum_{i_1=0}^{N} V_{i_1}^{\top} V_{i_1}\right)+\mathcal{O}{(\eta)}\nonumber\\
    &=\frac{N-B}{(N-1)B}\cdot \underbrace{\frac{1}{N}\sum_{j=1}^{N}\|\nabla \mL_{\bw}^j- \nabla\mL^{\mu}_{\bw}\|_2^2}_{\text{variance of one-sample gradients}} +\mathcal{O}{(\eta)}.
\end{align*}
According to \cref{prop: gradient-diversity-quadratic}, the above results also applies for distributed centralized SGD with equivalently large batch size $B$. The proof is now complete.

{\color{magenta}\qed}

\cref{th: large-batch} is then established based on \cref{lemma: implicit-reg-sgd}.

\textit{Proof of \cref{th: large-batch}}.

According to \cref{th: dsgd-sam},  the gradient of D-SGD can be written as
\begin{align}
    \ebb_{\mu}[\nabla\mL_\bw^{\textnormal{D-SGD}}]= \ebb_{\epsilon \sim \mathcal{N}(0, \bm{\Xi}{\scriptstyle (t)})}[\nabla\mL_{\bw+\epsilon}]
    +\mathcal{R},\label{eq: dsgd-mean-iterate}
\end{align}
where the higher-order residual terms are absorbed in $\mathcal{R}$ for simplicity\footnote{In the following, $\mathcal{R}^{\mu}$ and $\mathcal{R}^i$ denote the residual terms evaluated on super batch $\mu$ and sample $j$, respectively}.

The mini-batch form of \cref{eq: dsgd-mean-iterate} is
\begin{align*}
    \hat{\nabla\mL_\bw^{\mu}}=\ebb_{\epsilon \sim \mathcal{N}(0, \bm{\Xi}{\scriptstyle (t)})}[\nabla\mL_{\bw+\epsilon}^\mu]+\mathcal{R}^{\mu}.
\end{align*}
One can also derive the corresponding gradient evaluated on the $j$-th sample as
\begin{align*}
    \hat{\nabla\mL_\bw^{j}}=\ebb_{\epsilon \sim \mathcal{N}(0, \bm{\Xi}{\scriptstyle (t)})}[\nabla\mL_{\bw+\epsilon}^j]+\mathcal{R}^j.
\end{align*}
Replacing the original loss, the mini-batch gradient $\nabla\mL_\bw^{\mu}$ and the $j$-th sample gradient $\nabla\mL_\bw^{j}$ in \cref{eq: implicit-reg-sgd} with $\ebb_{\epsilon \sim \mathcal{N}(0, \bm{\Xi}{\scriptstyle (t)})}[\mL_{\bw+\epsilon}^\mu]$, $\hat{\nabla\mL_\bw^{\mu}}$ and  $\hat{\nabla\mL_\bw^{j}}$, respectively, we obtain,
\begin{align*}
    \mL_{\bw}^{\text{D-SGD}}=
    &\ebb_{\epsilon \sim \mathcal{N}(0, \bm{\Xi}{\scriptstyle (t)})}[\mL_{\bw+\epsilon}^\mu]+\frac{\eta}{4}\|\ebb_{\epsilon \sim \mathcal{N}(0, \bm{\Xi}{\scriptstyle (t)})}[\nabla\mL_{\bw+\epsilon}^\mu]+\nabla\mathcal{R}^{\mu}\|_2^2 +\mathcal{R}^{\mu} \nonumber\\
    &+ \frac{\eta}{B}\cdot\frac{N-B}{(N-1)}\cdot \frac{1}{N}\sum_{j=1}^{N}\|\ebb_{\epsilon \sim \mathcal{N}(0, \bm{\Xi}{\scriptstyle (t)})}[\nabla\mL_{\bw+\epsilon}^j]- \ebb_{\epsilon \sim \mathcal{N}(0, \bm{\Xi}{\scriptstyle (t)})}[\nabla\mL_{\bw+\epsilon}^\mu]+\mathcal{R}^{j}-\mathcal{R}^{\mu}\|_2^2 +\mathcal{O}{(\eta^2)}\nonumber\\
    =&\mL_{\bw}^\mu +\Tr(\mH^\mu_\bw \bm{\Xi}{\scriptstyle (t)})
    + \frac{\eta}{4}\|\nabla\mL_{\bw}^\mu\|_2^2 +\frac{\eta}{4}\Tr({(\mH^\mu_\bw)}^2 \bm{\Xi}{\scriptstyle (t)})\nonumber\\
    &+\frac{\eta}{B}\cdot\frac{N-B}{(N-1)}\cdot \frac{1}{N}\sum_{j=1}^{N}\left[\|\nabla\mL_{\bw}^j- \nabla\mL_{\bw}^\mu\|_2^2 +
    \Tr({(\mH_\bw^j-\mH_\bw^\mu)}^2\bm{\Xi}{\scriptstyle (t)})\right]
    +\mathcal{R}^{A}+\mathcal{O}{(\eta^2)},
\end{align*}
with a probability greater than $1-\mathcal{O}(\frac{B}{(N-B)\eta^2})$, where $\mathcal{R}^{A}$ absorb all higher-order residuals. The second equality is due to first-order Taylor expansions.

The proof is now complete.

{\color{magenta}\qed}

%% file: example_paper.bbl
\begin{thebibliography}{133}
\providecommand{\natexlab}[1]{#1}
\providecommand{\url}[1]{\texttt{#1}}
\expandafter\ifx\csname urlstyle\endcsname\relax
  \providecommand{\doi}[1]{doi: #1}\else
  \providecommand{\doi}{doi: \begingroup \urlstyle{rm}\Url}\fi

\bibitem[Andriushchenko \& Flammarion(2022)Andriushchenko and
  Flammarion]{pmlr-v162-andriushchenko22a}
Andriushchenko, M. and Flammarion, N.
\newblock Towards understanding sharpness-aware minimization.
\newblock In \emph{Proceedings of the 39th International Conference on Machine
  Learning}, volume 162, pp.\  639--668. PMLR, 2022.

\bibitem[Andriushchenko et~al.(2023)Andriushchenko, Bahri, Mobahi, and
  Flammarion]{andriushchenko2023sharpness}
Andriushchenko, M., Bahri, D., Mobahi, H., and Flammarion, N.
\newblock Sharpness-aware minimization leads to low-rank features.
\newblock \emph{arXiv preprint arXiv:2305.16292}, 2023.

\bibitem[Assran et~al.(2019)Assran, Loizou, Ballas, and
  Rabbat]{assran2019stochastic}
Assran, M., Loizou, N., Ballas, N., and Rabbat, M.
\newblock Stochastic gradient push for distributed deep learning.
\newblock In \emph{International Conference on Machine Learning}, 2019.

\bibitem[Bahri et~al.(2022)Bahri, Mobahi, and Tay]{bahri-etal-2022-sharpness}
Bahri, D., Mobahi, H., and Tay, Y.
\newblock Sharpness-aware minimization improves language model generalization.
\newblock In \emph{Proceedings of the 60th Annual Meeting of the Association
  for Computational Linguistics (Volume 1: Long Papers)}, pp.\  7360--7371.
  Association for Computational Linguistics, 2022.

\bibitem[Bassily et~al.(2020)Bassily, Feldman, Guzm\'{a}n, and
  Talwar]{NEURIPS2020_2e2c4bf7}
Bassily, R., Feldman, V., Guzm\'{a}n, C., and Talwar, K.
\newblock Stability of stochastic gradient descent on nonsmooth convex losses.
\newblock In \emph{Advances in Neural Information Processing Systems}, 2020.

\bibitem[Beltr{\'a}n et~al.(2022)Beltr{\'a}n, P{\'e}rez, S{\'a}nchez, Bernal,
  Bovet, P{\'e}rez, P{\'e}rez, and Celdr{\'a}n]{beltran2022decentralized}
Beltr{\'a}n, E. T.~M., P{\'e}rez, M.~Q., S{\'a}nchez, P. M.~S., Bernal, S.~L.,
  Bovet, G., P{\'e}rez, M.~G., P{\'e}rez, G.~M., and Celdr{\'a}n, A.~H.
\newblock Decentralized federated learning: Fundamentals, state-of-the-art,
  frameworks, trends, and challenges.
\newblock \emph{arXiv preprint arXiv:2211.08413}, 2022.

\bibitem[Bisla et~al.(2022)Bisla, Wang, and Choromanska]{bisla2022low}
Bisla, D., Wang, J., and Choromanska, A.
\newblock Low-pass filtering {SGD} for recovering flat optima in the deep
  learning optimization landscape.
\newblock In \emph{International Conference on Artificial Intelligence and
  Statistics}, pp.\  8299--8339. PMLR, 2022.

\bibitem[Bornstein et~al.(2023)Bornstein, Rabbani, Wang, Bedi, and
  Huang]{bornstein2023swift}
Bornstein, M., Rabbani, T., Wang, E.~Z., Bedi, A., and Huang, F.
\newblock {SWIFT}: Rapid decentralized federated learning via wait-free model
  communication.
\newblock In \emph{The Eleventh International Conference on Learning
  Representations}, 2023.

\bibitem[Borzunov et~al.(2022)Borzunov, Baranchuk, Dettmers, Ryabinin, Belkada,
  Chumachenko, Samygin, and Raffel]{borzunov2022petals}
Borzunov, A., Baranchuk, D., Dettmers, T., Ryabinin, M., Belkada, Y.,
  Chumachenko, A., Samygin, P., and Raffel, C.
\newblock Petals: Collaborative inference and fine-tuning of large models.
\newblock \emph{arXiv preprint arXiv:2209.01188}, 2022.

\bibitem[Bottou et~al.(2018)Bottou, Curtis, and
  Nocedal]{bottou2018optimization}
Bottou, L., Curtis, F.~E., and Nocedal, J.
\newblock Optimization methods for large-scale machine learning.
\newblock \emph{Siam Review}, 60\penalty0 (2):\penalty0 223--311, 2018.

\bibitem[Boyd et~al.(2004)Boyd, Boyd, and Vandenberghe]{boyd2004convex}
Boyd, S., Boyd, S.~P., and Vandenberghe, L.
\newblock \emph{Convex optimization}.
\newblock Cambridge university press, 2004.

\bibitem[Cao et~al.(2023)Cao, Rizk, Vlaski, and Sayed]{cao2023decentralized}
Cao, Y., Rizk, E., Vlaski, S., and Sayed, A.~H.
\newblock Decentralized adversarial training over graphs.
\newblock \emph{arXiv preprint arXiv:2303.13326}, 2023.

\bibitem[Cauchy et~al.(1847)]{cauchy1847methode}
Cauchy, A. et~al.
\newblock M{\'e}thode g{\'e}n{\'e}rale pour la r{\'e}solution des systemes
  d’{\'e}quations simultan{\'e}es.
\newblock \emph{Comp. Rend. Sci. Paris}, 25\penalty0 (1847):\penalty0 536--538,
  1847.

\bibitem[Chaudhari et~al.(2019)Chaudhari, Choromanska, Soatto, LeCun, Baldassi,
  Borgs, Chayes, Sagun, and Zecchina]{chaudhari2019entropy}
Chaudhari, P., Choromanska, A., Soatto, S., LeCun, Y., Baldassi, C., Borgs, C.,
  Chayes, J., Sagun, L., and Zecchina, R.
\newblock Entropy-{SGD}: Biasing gradient descent into wide valleys.
\newblock \emph{Journal of Statistical Mechanics: Theory and Experiment},
  2019\penalty0 (12):\penalty0 124018, 2019.

\bibitem[Chebyshev(1874)]{chebyshev1874valeurs}
Chebyshev, P.~L.
\newblock \emph{Sur les valeurs limites des intégrales}.
\newblock Imprimerie de Gauthier-Villars Paris, 1874.

\bibitem[Chen \& Huo(2016)Chen and Huo]{chen2016scalable}
Chen, K. and Huo, Q.
\newblock Scalable training of deep learning machines by incremental block
  training with intra-block parallel optimization and blockwise model-update
  filtering.
\newblock In \emph{2016 IEEE International Conference on Acoustics, Speech and
  Signal Processing (ICASSP)}, pp.\  5880--5884. IEEE Press, 2016.

\bibitem[Chen et~al.(2022)Chen, Hsieh, and Gong]{chen2022when}
Chen, X., Hsieh, C.-J., and Gong, B.
\newblock When vision transformers outperform resnets without pre-training or
  strong data augmentations.
\newblock In \emph{International Conference on Learning Representations}, 2022.

\bibitem[Chor et~al.(2023)Chor, Sefidgaran, and Zaidi]{chor2023more}
Chor, R., Sefidgaran, M., and Zaidi, A.
\newblock More communication does not result in smaller generalization error in
  federated learning.
\newblock \emph{arXiv preprint arXiv:2304.12216}, 2023.

\bibitem[Davis(1962)]{davis1962norm}
Davis, C.
\newblock The norm of the schur product operation.
\newblock \emph{Numerische Mathematik}, 4:\penalty0 343--344, 1962.

\bibitem[De~Bruijn(1981)]{de1981asymptotic}
De~Bruijn, N.~G.
\newblock \emph{Asymptotic methods in analysis}, volume~4.
\newblock Courier Corporation, 1981.

\bibitem[Dean et~al.(2012)Dean, Corrado, Monga, Chen, Devin, Mao, Ranzato,
  Senior, Tucker, Yang, et~al.]{dean2012large}
Dean, J., Corrado, G., Monga, R., Chen, K., Devin, M., Mao, M., Ranzato, M.,
  Senior, A., Tucker, P., Yang, K., et~al.
\newblock Large scale distributed deep networks.
\newblock \emph{Advances in neural information processing systems}, 2012.

\bibitem[Deng et~al.(2023)Deng, Sun, Li, and Li]{deng2023stability}
Deng, X., Sun, T., Li, S., and Li, D.
\newblock Stability-based generalization analysis of the asynchronous
  decentralized sgd.
\newblock In \emph{Proceedings of the AAAI Conference on Artificial
  Intelligence}, 2023.

\bibitem[Dosovitskiy et~al.(2021)Dosovitskiy, Beyer, Kolesnikov, Weissenborn,
  Zhai, Unterthiner, Dehghani, Minderer, Heigold, Gelly, Uszkoreit, and
  Houlsby]{dosovitskiy2021an}
Dosovitskiy, A., Beyer, L., Kolesnikov, A., Weissenborn, D., Zhai, X.,
  Unterthiner, T., Dehghani, M., Minderer, M., Heigold, G., Gelly, S.,
  Uszkoreit, J., and Houlsby, N.
\newblock An image is worth 16x16 words: Transformers for image recognition at
  scale.
\newblock In \emph{International Conference on Learning Representations}, 2021.

\bibitem[Du et~al.(2022)Du, Yan, Feng, Zhou, Zhen, Goh, and
  Tan]{du2022efficient}
Du, J., Yan, H., Feng, J., Zhou, J.~T., Zhen, L., Goh, R. S.~M., and Tan, V.
\newblock Efficient sharpness-aware minimization for improved training of
  neural networks.
\newblock In \emph{International Conference on Learning Representations}, 2022.

\bibitem[Erd{\'e}lyi(1956)]{erdelyi1956asymptotic}
Erd{\'e}lyi, A.
\newblock \emph{Asymptotic expansions}.
\newblock Number~3. Courier Corporation, 1956.

\bibitem[Farhadkhani et~al.(2023)Farhadkhani, Guerraoui, Gupta, Hoang, Pinot,
  and Stephan]{Farhadkhani2023}
Farhadkhani, S., Guerraoui, R., Gupta, N., Hoang, L.~N., Pinot, R., and
  Stephan, J.
\newblock Robust collaborative learning with linear gradient overhead.
\newblock In \emph{International Conference on Machine Learning}, 2023.

\bibitem[Foret et~al.(2021)Foret, Kleiner, Mobahi, and
  Neyshabur]{foret2021sharpnessaware}
Foret, P., Kleiner, A., Mobahi, H., and Neyshabur, B.
\newblock Sharpness-aware minimization for efficiently improving
  generalization.
\newblock In \emph{International Conference on Learning Representations}, 2021.

\bibitem[Geiping et~al.(2020)Geiping, Bauermeister, Dr\"{o}ge, and
  Moeller]{NEURIPS2020_c4ede56b}
Geiping, J., Bauermeister, H., Dr\"{o}ge, H., and Moeller, M.
\newblock Inverting gradients - how easy is it to break privacy in federated
  learning?
\newblock In \emph{Advances in Neural Information Processing Systems}, 2020.

\bibitem[Goyal et~al.(2017)Goyal, Doll{\'a}r, Girshick, Noordhuis, Wesolowski,
  Kyrola, Tulloch, Jia, and He]{goyal2017accurate}
Goyal, P., Doll{\'a}r, P., Girshick, R., Noordhuis, P., Wesolowski, L., Kyrola,
  A., Tulloch, A., Jia, Y., and He, K.
\newblock Accurate, large minibatch sgd: Training imagenet in 1 hour.
\newblock \emph{arXiv preprint arXiv:1706.02677}, 2017.

\bibitem[Gu et~al.(2023)Gu, Lyu, Huang, and Arora]{localsgdgeneralizes2023}
Gu, X., Lyu, K., Huang, L., and Arora, S.
\newblock Why (and when) does local {SGD} generalize better than {SGD}?
\newblock In \emph{International Conference on Learning Representations}, 2023.

\bibitem[Gurbuzbalaban et~al.(2022)Gurbuzbalaban, Hu, Simsekli, Yuan, and
  Zhu]{gurbuzbalaban2022heavy}
Gurbuzbalaban, M., Hu, Y., Simsekli, U., Yuan, K., and Zhu, L.
\newblock Heavy-tail phenomenon in decentralized sgd.
\newblock \emph{arXiv preprint arXiv:2205.06689}, 2022.

\bibitem[He et~al.(2019)He, Liu, and Tao]{he2019control}
He, F., Liu, T., and Tao, D.
\newblock Control batch size and learning rate to generalize well: Theoretical
  and empirical evidence.
\newblock In \emph{Advances in Neural Information Processing Systems}, pp.\
  1143--1152, 2019.

\bibitem[He et~al.(2016{\natexlab{a}})He, Zhang, Ren, and Sun]{he2016deep}
He, K., Zhang, X., Ren, S., and Sun, J.
\newblock Deep residual learning for image recognition.
\newblock In \emph{Proceedings of the IEEE conference on computer vision and
  pattern recognition}, 2016{\natexlab{a}}.

\bibitem[He et~al.(2016{\natexlab{b}})He, Zhang, Ren, and Sun]{he2016identity}
He, K., Zhang, X., Ren, S., and Sun, J.
\newblock Identity mappings in deep residual networks.
\newblock In \emph{European conference on computer vision}. Springer,
  2016{\natexlab{b}}.

\bibitem[Hochreiter \& Schmidhuber(1997)Hochreiter and
  Schmidhuber]{hochreiter1997flat}
Hochreiter, S. and Schmidhuber, J.
\newblock Flat minima.
\newblock \emph{Neural computation}, 9\penalty0 (1):\penalty0 1--42, 1997.

\bibitem[Hoffer et~al.(2017)Hoffer, Hubara, and Soudry]{hoffer2017train}
Hoffer, E., Hubara, I., and Soudry, D.
\newblock Train longer, generalize better: closing the generalization gap in
  large batch training of neural networks.
\newblock \emph{Advances in neural information processing systems}, 2017.

\bibitem[Huang et~al.(2017)Huang, Liu, van~der Maaten, and
  Weinberger]{Huang_2017_CVPR}
Huang, G., Liu, Z., van~der Maaten, L., and Weinberger, K.~Q.
\newblock Densely connected convolutional networks.
\newblock In \emph{Proceedings of the IEEE Conference on Computer Vision and
  Pattern Recognition (CVPR)}, July 2017.

\bibitem[Ibayashi \& Imaizumi(2021)Ibayashi and Imaizumi]{ibayashi2021quasi}
Ibayashi, H. and Imaizumi, M.
\newblock Exponential escape efficiency of sgd from sharp minima in
  non-stationary regime.
\newblock \emph{arXiv preprint arXiv:2111.04004}, 2021.

\bibitem[Ioffe \& Szegedy(2015)Ioffe and Szegedy]{ioffe2015batch}
Ioffe, S. and Szegedy, C.
\newblock Batch normalization: Accelerating deep network training by reducing
  internal covariate shift.
\newblock In \emph{International conference on machine learning}, 2015.

\bibitem[Izmailov et~al.(2018)Izmailov, Podoprikhin, Garipov, Vetrov, and
  Wilson]{izmailov2018averaging}
Izmailov, P., Podoprikhin, D., Garipov, T., Vetrov, D., and Wilson, A.~G.
\newblock Averaging weights leads to wider optima and better generalization.
\newblock \emph{arXiv preprint arXiv:1803.05407}, 2018.

\bibitem[Jiang et~al.(2020)Jiang, Neyshabur*, Mobahi, Krishnan, and
  Bengio]{Jiang*2020Fantastic}
Jiang, Y., Neyshabur*, B., Mobahi, H., Krishnan, D., and Bengio, S.
\newblock Fantastic generalization measures and where to find them.
\newblock In \emph{International Conference on Learning Representations}, 2020.

\bibitem[Keskar et~al.(2017)Keskar, Mudigere, Nocedal, Smelyanskiy, and
  Tang]{keskar2017on}
Keskar, N.~S., Mudigere, D., Nocedal, J., Smelyanskiy, M., and Tang, P. T.~P.
\newblock On large-batch training for deep learning: Generalization gap and
  sharp minima.
\newblock In \emph{International Conference on Learning Representations}. PMLR,
  2017.

\bibitem[Kim et~al.(2023)Kim, Park, Choi, and Lee]{kim2023stability}
Kim, H., Park, J., Choi, Y., and Lee, J.
\newblock Stability analysis of sharpness-aware minimization.
\newblock \emph{arXiv preprint arXiv:2301.06308}, 2023.

\bibitem[Kim et~al.(2022)Kim, Li, Hu, and Hospedales]{pmlr-v162-kim22f}
Kim, M., Li, D., Hu, S.~X., and Hospedales, T.
\newblock {F}isher {SAM}: Information geometry and sharpness aware
  minimisation.
\newblock In \emph{Proceedings of the 39th International Conference on Machine
  Learning}, volume 162 of \emph{Proceedings of Machine Learning Research},
  pp.\  11148--11161. PMLR, 2022.

\bibitem[Koloskova et~al.(2020)Koloskova, Loizou, Boreiri, Jaggi, and
  Stich]{pmlr-v119-koloskova20a}
Koloskova, A., Loizou, N., Boreiri, S., Jaggi, M., and Stich, S.
\newblock A unified theory of decentralized {SGD} with changing topology and
  local updates.
\newblock In \emph{International Conference on Machine Learning}, 2020.

\bibitem[Kong et~al.(2021)Kong, Lin, Koloskova, Jaggi, and
  Stich]{kong2021consensus}
Kong, L., Lin, T., Koloskova, A., Jaggi, M., and Stich, S.
\newblock Consensus control for decentralized deep learning.
\newblock In \emph{International Conference on Machine Learning}. PMLR, 2021.

\bibitem[Krizhevsky et~al.(2009)Krizhevsky, Hinton,
  et~al.]{krizhevsky2009learning}
Krizhevsky, A., Hinton, G., et~al.
\newblock Learning multiple layers of features from tiny images (tech. rep.).
\newblock \emph{University of Toronto}, 2009.

\bibitem[Krizhevsky et~al.(2017)Krizhevsky, Sutskever, and
  Hinton]{krizhevsky2017imagenet}
Krizhevsky, A., Sutskever, I., and Hinton, G.~E.
\newblock Imagenet classification with deep convolutional neural networks.
\newblock \emph{Communications of the ACM}, 60\penalty0 (6):\penalty0 84--90,
  2017.

\bibitem[Kwon et~al.(2021)Kwon, Kim, Park, and Choi]{kwon2021asam}
Kwon, J., Kim, J., Park, H., and Choi, I.~K.
\newblock Asam: Adaptive sharpness-aware minimization for scale-invariant
  learning of deep neural networks.
\newblock In \emph{International Conference on Machine Learning}, pp.\
  5905--5914. PMLR, 2021.

\bibitem[Le \& Yang(2015)Le and Yang]{le2015tiny}
Le, Y. and Yang, X.
\newblock Tiny imagenet visual recognition challenge.
\newblock \emph{CS 231N}, 2015.

\bibitem[Le~Bars et~al.(2023)Le~Bars, Bellet, Tommasi, Lavoie, and
  Kermarrec]{pmlr-v206-le-bars23a}
Le~Bars, B., Bellet, A., Tommasi, M., Lavoie, E., and Kermarrec, A.-M.
\newblock Refined convergence and topology learning for decentralized sgd with
  heterogeneous data.
\newblock In \emph{Proceedings of The 26th International Conference on
  Artificial Intelligence and Statistics}, volume 206, pp.\  1672--1702. PMLR,
  25--27 Apr 2023.

\bibitem[Li et~al.(2018)Li, Xu, Taylor, Studer, and
  Goldstein]{li2018visualizing}
Li, H., Xu, Z., Taylor, G., Studer, C., and Goldstein, T.
\newblock Visualizing the loss landscape of neural nets.
\newblock \emph{Advances in neural information processing systems}, 2018.

\bibitem[Li et~al.(2014)Li, Andersen, Smola, and Yu]{li2014communication}
Li, M., Andersen, D.~G., Smola, A.~J., and Yu, K.
\newblock Communication efficient distributed machine learning with the
  parameter server.
\newblock \emph{Advances in Neural Information Processing Systems}, 2014.

\bibitem[Li et~al.(2022)Li, Zhou, Tian, and Tao]{li2022learning}
Li, S., Zhou, T., Tian, X., and Tao, D.
\newblock Learning to collaborate in decentralized learning of personalized
  models.
\newblock In \emph{Proceedings of the IEEE/CVF Conference on Computer Vision
  and Pattern Recognition}, pp.\  9766--9775, 2022.

\bibitem[Li et~al.(2021)Li, Malladi, and Arora]{li2021validity}
Li, Z., Malladi, S., and Arora, S.
\newblock On the validity of modeling sgd with stochastic differential
  equations (sdes).
\newblock \emph{Advances in Neural Information Processing Systems}, 2021.

\bibitem[Lian et~al.(2017)Lian, Zhang, Zhang, Hsieh, Zhang, and
  Liu]{NIPS2017_f7552665}
Lian, X., Zhang, C., Zhang, H., Hsieh, C.-J., Zhang, W., and Liu, J.
\newblock Can decentralized algorithms outperform centralized algorithms? a
  case study for decentralized parallel stochastic gradient descent.
\newblock In \emph{Advances in Neural Information Processing Systems}, 2017.

\bibitem[Lian et~al.(2018)Lian, Zhang, Zhang, and Liu]{lian2018asynchronous}
Lian, X., Zhang, W., Zhang, C., and Liu, J.
\newblock Asynchronous decentralized parallel stochastic gradient descent.
\newblock In \emph{International Conference on Machine Learning}, 2018.

\bibitem[Liu et~al.(2021)Liu, Ziyin, and Ueda]{liu2021noise}
Liu, K., Ziyin, L., and Ueda, M.
\newblock Noise and fluctuation of finite learning rate stochastic gradient
  descent.
\newblock In \emph{International Conference on Machine Learning}. PMLR, 2021.

\bibitem[Liu et~al.(2022{\natexlab{a}})Liu, Mai, Chen, Hsieh, and
  You]{liu2022towards}
Liu, Y., Mai, S., Chen, X., Hsieh, C.-J., and You, Y.
\newblock Towards efficient and scalable sharpness-aware minimization.
\newblock In \emph{Proceedings of the IEEE/CVF Conference on Computer Vision
  and Pattern Recognition}, pp.\  12360--12370, 2022{\natexlab{a}}.

\bibitem[Liu et~al.(2022{\natexlab{b}})Liu, Mai, Cheng, Chen, Hsieh, and
  You]{liu2022random}
Liu, Y., Mai, S., Cheng, M., Chen, X., Hsieh, C.-J., and You, Y.
\newblock Random sharpness-aware minimization.
\newblock In Oh, A.~H., Agarwal, A., Belgrave, D., and Cho, K. (eds.),
  \emph{Advances in Neural Information Processing Systems}, 2022{\natexlab{b}}.

\bibitem[Liu et~al.(2022{\natexlab{c}})Liu, Kangqiao, Takashi, and
  Masahito]{ziyin2022strength}
Liu, Z., Kangqiao, L., Takashi, M., and Masahito, U.
\newblock Strength of minibatch noise in {SGD}.
\newblock In \emph{International Conference on Learning Representations},
  2022{\natexlab{c}}.

\bibitem[Lopes \& Sayed(2008)Lopes and Sayed]{lopes2008diffusion}
Lopes, C.~G. and Sayed, A.~H.
\newblock Diffusion least-mean squares over adaptive networks: Formulation and
  performance analysis.
\newblock \emph{IEEE Transactions on Signal Processing}, 2008.

\bibitem[Lu et~al.(2011)Lu, Tang, Regier, and Bow]{lu2011gossip}
Lu, J., Tang, C.~Y., Regier, P.~R., and Bow, T.~D.
\newblock Gossip algorithms for convex consensus optimization over networks.
\newblock \emph{IEEE Transactions on Automatic Control}, 2011.

\bibitem[Lu \& Wu(2020)Lu and Wu]{lu2020decentralized}
Lu, S. and Wu, C.~W.
\newblock Decentralized stochastic non-convex optimization over weakly
  connected time-varying digraphs.
\newblock In \emph{ICASSP 2020-2020 IEEE International Conference on Acoustics,
  Speech and Signal Processing (ICASSP)}, 2020.

\bibitem[Lu \& Sa(2023)Lu and Sa]{JMLR:v24:22-0044}
Lu, Y. and Sa, C.~D.
\newblock Decentralized learning: Theoretical optimality and practical
  improvements.
\newblock \emph{Journal of Machine Learning Research}, 24\penalty0
  (93):\penalty0 1--62, 2023.

\bibitem[Ma et~al.(2022)Ma, Kunin, Wu, and Ying]{JML-1-247}
Ma, C., Kunin, D., Wu, L., and Ying, L.
\newblock Beyond the quadratic approximation: The multiscale structure of
  neural network loss landscapes.
\newblock \emph{Journal of Machine Learning}, 1\penalty0 (3):\penalty0
  247--267, 2022.

\bibitem[MacKay(1992)]{mackay1992practical}
MacKay, D.~J.
\newblock A practical bayesian framework for backpropagation networks.
\newblock \emph{Neural computation}, 4\penalty0 (3):\penalty0 448--472, 1992.

\bibitem[Marshall \& Olkin(1960)Marshall and Olkin]{marshall1960multivariate}
Marshall, A.~W. and Olkin, I.
\newblock Multivariate chebyshev inequalities.
\newblock \emph{The Annals of Mathematical Statistics}, pp.\  1001--1014, 1960.

\bibitem[Marshall et~al.(1979)Marshall, Olkin, and
  Arnold]{marshall1979inequalities}
Marshall, A.~W., Olkin, I., and Arnold, B.~C.
\newblock \emph{Inequalities: theory of majorization and its applications}.
\newblock Springer, 1979.

\bibitem[McMahan et~al.(2017)McMahan, Moore, Ramage, Hampson, and
  Arcas]{pmlr-v54-mcmahan17a}
McMahan, B., Moore, E., Ramage, D., Hampson, S., and Arcas, B. A.~y.
\newblock {Communication-Efficient Learning of Deep Networks from Decentralized
  Data}.
\newblock In \emph{Proceedings of the 20th International Conference on
  Artificial Intelligence and Statistics}, volume~54, pp.\  1273--1282. PMLR,
  2017.

\bibitem[Merikoski(1984)]{merikoski1984trace}
Merikoski, J.~K.
\newblock On the trace and the sum of elements of a matrix.
\newblock \emph{Linear algebra and its applications}, 60:\penalty0 177--185,
  1984.

\bibitem[Mi et~al.(2022)Mi, Shen, Ren, Zhou, Sun, Ji, and Tao]{mi2022make}
Mi, P., Shen, L., Ren, T., Zhou, Y., Sun, X., Ji, R., and Tao, D.
\newblock Make sharpness-aware minimization stronger: A sparsified perturbation
  approach.
\newblock In Oh, A.~H., Agarwal, A., Belgrave, D., and Cho, K. (eds.),
  \emph{Advances in Neural Information Processing Systems}, 2022.

\bibitem[Mueller \& Hein(2022)Mueller and Hein]{mueller2022perturbing}
Mueller, M. and Hein, M.
\newblock Perturbing batchnorm and only batchnorm benefits sharpness-aware
  minimization.
\newblock In \emph{Has it Trained Yet? NeurIPS 2022 Workshop}, 2022.

\bibitem[Möllenhoff \& Khan(2023)Möllenhoff and Khan]{möllenhoff2023sam}
Möllenhoff, T. and Khan, M.~E.
\newblock {SAM} as an optimal relaxation of bayes.
\newblock In \emph{The Eleventh International Conference on Learning
  Representations}, 2023.

\bibitem[Nadiradze et~al.(2021)Nadiradze, Sabour, Davies, Li, and
  Alistarh]{nadiradze2021asynchronous}
Nadiradze, G., Sabour, A., Davies, P., Li, S., and Alistarh, D.
\newblock Asynchronous decentralized sgd with quantized and local updates.
\newblock \emph{Advances in Neural Information Processing Systems}, 2021.

\bibitem[Narayanan et~al.(2021)Narayanan, Shoeybi, Casper, LeGresley, Patwary,
  Korthikanti, Vainbrand, Kashinkunti, Bernauer, Catanzaro,
  et~al.]{narayanan2021efficient}
Narayanan, D., Shoeybi, M., Casper, J., LeGresley, P., Patwary, M.,
  Korthikanti, V., Vainbrand, D., Kashinkunti, P., Bernauer, J., Catanzaro, B.,
  et~al.
\newblock Efficient large-scale language model training on gpu clusters using
  megatron-lm.
\newblock In \emph{Proceedings of the International Conference for High
  Performance Computing, Networking, Storage and Analysis}, pp.\  1--15, 2021.

\bibitem[Nedic(2020)]{nedic2020distributed}
Nedic, A.
\newblock Distributed gradient methods for convex machine learning problems in
  networks: Distributed optimization.
\newblock \emph{IEEE Signal Processing Magazine}, 2020.

\bibitem[Nedi{\'c} \& Olshevsky(2014)Nedi{\'c} and
  Olshevsky]{nedic2014distributed}
Nedi{\'c}, A. and Olshevsky, A.
\newblock Distributed optimization over time-varying directed graphs.
\newblock \emph{IEEE Transactions on Automatic Control}, 60\penalty0
  (3):\penalty0 601--615, 2014.

\bibitem[Nedic \& Ozdaglar(2009)Nedic and Ozdaglar]{nedic2009distributed}
Nedic, A. and Ozdaglar, A.
\newblock Distributed subgradient methods for multi-agent optimization.
\newblock \emph{IEEE Transactions on Automatic Control}, 54\penalty0
  (1):\penalty0 48--61, 2009.

\bibitem[Neyshabur et~al.(2017)Neyshabur, Bhojanapalli, Mcallester, and
  Srebro]{NIPS2017_10ce03a1}
Neyshabur, B., Bhojanapalli, S., Mcallester, D., and Srebro, N.
\newblock Exploring generalization in deep learning.
\newblock In \emph{Advances in Neural Information Processing Systems}, 2017.

\bibitem[Paszke et~al.(2019)Paszke, Gross, Massa, Lerer, Bradbury, Chanan,
  Killeen, Lin, Gimelshein, Antiga, et~al.]{paszke2019pytorch}
Paszke, A., Gross, S., Massa, F., Lerer, A., Bradbury, J., Chanan, G., Killeen,
  T., Lin, Z., Gimelshein, N., Antiga, L., et~al.
\newblock Pytorch: An imperative style, high-performance deep learning library.
\newblock \emph{Advances in neural information processing systems}, 2019.

\bibitem[Reddi et~al.(2021)Reddi, Charles, Zaheer, Garrett, Rush,
  Kone{\v{c}}n{\'y}, Kumar, and McMahan]{reddi2021adaptive}
Reddi, S.~J., Charles, Z., Zaheer, M., Garrett, Z., Rush, K.,
  Kone{\v{c}}n{\'y}, J., Kumar, S., and McMahan, H.~B.
\newblock Adaptive federated optimization.
\newblock In \emph{International Conference on Learning Representations}, 2021.

\bibitem[Rissanen(1983)]{rissanen1983universal}
Rissanen, J.
\newblock A universal prior for integers and estimation by minimum description
  length.
\newblock \emph{The Annals of statistics}, 11\penalty0 (2):\penalty0 416--431,
  1983.

\bibitem[Robbins(1951)]{Robbins1951ASA}
Robbins, H.~E.
\newblock A stochastic approximation method.
\newblock \emph{Annals of Mathematical Statistics}, 22:\penalty0 400--407,
  1951.

\bibitem[Rockafellar \& Wets(2009)Rockafellar and
  Wets]{rockafellar2009variational}
Rockafellar, R.~T. and Wets, R. J.-B.
\newblock \emph{Variational analysis}, volume 317.
\newblock Springer Science \& Business Media, 2009.

\bibitem[Rudin et~al.(1976)]{rudin1976principles}
Rudin, W. et~al.
\newblock \emph{Principles of mathematical analysis}.
\newblock McGraw-hill New York, 1976.

\bibitem[Seneta(2006)]{seneta2006non}
Seneta, E.
\newblock \emph{Non-negative matrices and Markov chains}.
\newblock Springer Science \& Business Media, 2006.

\bibitem[Shallue et~al.(2019)Shallue, Lee, Antognini, Sohl-Dickstein, Frostig,
  and Dahl]{JMLR:v20:18-789}
Shallue, C.~J., Lee, J., Antognini, J., Sohl-Dickstein, J., Frostig, R., and
  Dahl, G.~E.
\newblock Measuring the effects of data parallelism on neural network training.
\newblock \emph{Journal of Machine Learning Research}, 20\penalty0
  (112):\penalty0 1--49, 2019.

\bibitem[Shamir \& Srebro(2014)Shamir and Srebro]{shamir2014distributed}
Shamir, O. and Srebro, N.
\newblock Distributed stochastic optimization and learning.
\newblock In \emph{2014 52nd Annual Allerton Conference on Communication,
  Control, and Computing (Allerton)}, 2014.

\bibitem[Shen et~al.(2023)Shen, Sun, Yu, Ding, Tian, and
  Tao]{shen2023efficient}
Shen, L., Sun, Y., Yu, Z., Ding, L., Tian, X., and Tao, D.
\newblock On efficient training of large-scale deep learning models: A
  literature review.
\newblock \emph{arXiv preprint arXiv:2304.03589}, 2023.

\bibitem[Shi et~al.(2015)Shi, Ling, Wu, and Yin]{shi2015extra}
Shi, W., Ling, Q., Wu, G., and Yin, W.
\newblock Extra: An exact first-order algorithm for decentralized consensus
  optimization.
\newblock \emph{SIAM Journal on Optimization}, 2015.

\bibitem[Shi et~al.(2023{\natexlab{a}})Shi, Liu, Sun, Lin, Shen, Wang, and
  Tao]{shi2023towards}
Shi, Y., Liu, Y., Sun, Y., Lin, Z., Shen, L., Wang, X., and Tao, D.
\newblock Towards more suitable personalization in federated learning via
  decentralized partial model training.
\newblock \emph{arXiv preprint arXiv:2305.15157}, 2023{\natexlab{a}}.

\bibitem[Shi et~al.(2023{\natexlab{b}})Shi, Shen, Wei, Sun, Yuan, Wang, and
  Tao]{pmlr-v202-shi23d}
Shi, Y., Shen, L., Wei, K., Sun, Y., Yuan, B., Wang, X., and Tao, D.
\newblock Improving the model consistency of decentralized federated learning.
\newblock In \emph{Proceedings of the 40th International Conference on Machine
  Learning}, pp.\  31269--31291. PMLR, 2023{\natexlab{b}}.

\bibitem[Smith(2017)]{smith2017cyclical}
Smith, L.~N.
\newblock Cyclical learning rates for training neural networks.
\newblock In \emph{2017 IEEE winter conference on applications of computer
  vision (WACV)}, pp.\  464--472. IEEE, 2017.

\bibitem[Smith et~al.(2020)Smith, Elsen, and De]{smith2020generalization}
Smith, S., Elsen, E., and De, S.
\newblock On the generalization benefit of noise in stochastic gradient
  descent.
\newblock In \emph{International Conference on Machine Learning}. PMLR, 2020.

\bibitem[Smith et~al.(2021)Smith, Dherin, Barrett, and De]{smith2021on}
Smith, S.~L., Dherin, B., Barrett, D., and De, S.
\newblock On the origin of implicit regularization in stochastic gradient
  descent.
\newblock In \emph{International Conference on Learning Representations}, 2021.

\bibitem[Song et~al.(2022)Song, Li, Jin, Shi, Yan, Yin, and
  Yuan]{song2022communicationefficient}
Song, Z., Li, W., Jin, K., Shi, L., Yan, M., Yin, W., and Yuan, K.
\newblock Communication-efficient topologies for decentralized learning with
  \$o(1)\$ consensus rate.
\newblock In Oh, A.~H., Agarwal, A., Belgrave, D., and Cho, K. (eds.),
  \emph{Advances in Neural Information Processing Systems}, 2022.

\bibitem[Sun et~al.(2021)Sun, Li, and Wang]{sun2021stability}
Sun, T., Li, D., and Wang, B.
\newblock Stability and generalization of decentralized stochastic gradient
  descent.
\newblock In \emph{Proceedings of the AAAI Conference on Artificial
  Intelligence}, volume~35, pp.\  9756--9764, 2021.

\bibitem[Taheri et~al.(2020)Taheri, Mokhtari, Hassani, and
  Pedarsani]{taheri2020quantized}
Taheri, H., Mokhtari, A., Hassani, H., and Pedarsani, R.
\newblock Quantized decentralized stochastic learning over directed graphs.
\newblock In \emph{International Conference on Machine Learning}. PMLR, 2020.

\bibitem[Tang et~al.(2018)Tang, Lian, Yan, Zhang, and Liu]{tang2018d}
Tang, H., Lian, X., Yan, M., Zhang, C., and Liu, J.
\newblock D2: Decentralized training over decentralized data.
\newblock In \emph{International Conference on Machine Learning}. PMLR, 2018.

\bibitem[Tsitsiklis et~al.(1986)Tsitsiklis, Bertsekas, and
  Athans]{tsitsiklis1986distributed}
Tsitsiklis, J., Bertsekas, D., and Athans, M.
\newblock Distributed asynchronous deterministic and stochastic gradient
  optimization algorithms.
\newblock \emph{IEEE transactions on automatic control}, 31\penalty0
  (9):\penalty0 803--812, 1986.

\bibitem[Tsitsiklis(1984)]{tsitsiklis1984problems}
Tsitsiklis, J.~N.
\newblock Problems in decentralized decision making and computation.
\newblock Technical report, Massachusetts Inst of Tech Cambridge Lab for
  Information and Decision Systems, 1984.

\bibitem[Ujv{\'a}ry et~al.(2022)Ujv{\'a}ry, Telek, Kerekes, M{\'e}sz{\'a}ros,
  and Husz{\'a}r]{ujvary2022rethinking}
Ujv{\'a}ry, S., Telek, Z., Kerekes, A., M{\'e}sz{\'a}ros, A., and Husz{\'a}r,
  F.
\newblock Rethinking sharpness-aware minimization as variational inference.
\newblock In \emph{OPT 2022: Optimization for Machine Learning (NeurIPS 2022
  Workshop)}, 2022.

\bibitem[Vershynin(2018)]{vershynin2018high}
Vershynin, R.
\newblock \emph{High-dimensional probability: An introduction with applications
  in data science}, volume~47.
\newblock Cambridge university press, 2018.

\bibitem[Vogels et~al.(2021)Vogels, He, Koloskova, Karimireddy, Lin, Stich, and
  Jaggi]{vogels2021relaysum}
Vogels, T., He, L., Koloskova, A., Karimireddy, S.~P., Lin, T., Stich, S.~U.,
  and Jaggi, M.
\newblock Relaysum for decentralized deep learning on heterogeneous data.
\newblock \emph{Advances in Neural Information Processing Systems},
  34:\penalty0 28004--28015, 2021.

\bibitem[Von~Neumann(1937)]{von1937some}
Von~Neumann, J.
\newblock \emph{Some matrix-inequalities and metrization of matric space}.
\newblock 1937.

\bibitem[Wang et~al.(2020)Wang, Liu, Liang, Joshi, and Poor]{wang2020tackling}
Wang, J., Liu, Q., Liang, H., Joshi, G., and Poor, H.~V.
\newblock Tackling the objective inconsistency problem in heterogeneous
  federated optimization.
\newblock \emph{Advances in neural information processing systems},
  33:\penalty0 7611--7623, 2020.

\bibitem[Wang et~al.(2022)Wang, Das, Joshi, Kale, Xu, and
  Zhang]{wang2022unreasonable}
Wang, J., Das, R., Joshi, G., Kale, S., Xu, Z., and Zhang, T.
\newblock On the unreasonable effectiveness of federated averaging with
  heterogeneous data.
\newblock \emph{arXiv preprint arXiv:2206.04723}, 2022.

\bibitem[Wang et~al.(2023)Wang, Lee, and Lei]{wang2023reconstructing}
Wang, Z., Lee, J., and Lei, Q.
\newblock Reconstructing training data from model gradient, provably.
\newblock In \emph{International Conference on Artificial Intelligence and
  Statistics}, pp.\  6595--6612. PMLR, 2023.

\bibitem[Warnat-Herresthal et~al.(2021)Warnat-Herresthal, Schultze, Shastry,
  Manamohan, Mukherjee, Garg, Sarveswara, H{\"a}ndler, Pickkers, Aziz,
  et~al.]{warnat2021swarm}
Warnat-Herresthal, S., Schultze, H., Shastry, K.~L., Manamohan, S., Mukherjee,
  S., Garg, V., Sarveswara, R., H{\"a}ndler, K., Pickkers, P., Aziz, N.~A.,
  et~al.
\newblock Swarm learning for decentralized and confidential clinical machine
  learning.
\newblock \emph{Nature}, 2021.

\bibitem[Wen et~al.(2023)Wen, Ma, and Li]{howsharpness2023}
Wen, K., Ma, T., and Li, Z.
\newblock How sharpness-aware minimization minimizes sharpness?
\newblock In \emph{International Conference on Learning Representations}, 2023.

\bibitem[Wu et~al.(2020)Wu, Xia, and Wang]{wu2020adversarial}
Wu, D., Xia, S.-T., and Wang, Y.
\newblock Adversarial weight perturbation helps robust generalization.
\newblock \emph{Advances in Neural Information Processing Systems},
  33:\penalty0 2958--2969, 2020.

\bibitem[Wu \& Su(2023)Wu and Su]{wu2023}
Wu, L. and Su, W.~J.
\newblock The implicit regularization of dynamical stability in stochastic
  gradient descent.
\newblock In \emph{International Conference on Machine Learning}. PMLR, 2023.

\bibitem[Xiao \& Boyd(2004)Xiao and Boyd]{xiao2004fast}
Xiao, L. and Boyd, S.
\newblock Fast linear iterations for distributed averaging.
\newblock \emph{Systems \& Control Letters}, 2004.

\bibitem[Xu et~al.(2021)Xu, Zhang, and Wang]{xu2021dp}
Xu, J., Zhang, W., and Wang, F.
\newblock A(dp)$^2$sgd: Asynchronous decentralized parallel stochastic gradient
  descent with differential privacy.
\newblock \emph{IEEE Transactions on Pattern Analysis and Machine
  Intelligence}, 2021.

\bibitem[Yang et~al.(2020)Yang, Gang, and Bajwa]{9084329}
Yang, Z., Gang, A., and Bajwa, W.~U.
\newblock Adversary-resilient distributed and decentralized statistical
  inference and machine learning: An overview of recent advances under the
  byzantine threat model.
\newblock \emph{IEEE Signal Processing Magazine}, 37\penalty0 (3):\penalty0
  146--159, 2020.

\bibitem[Yin et~al.(2018)Yin, Pananjady, Lam, Papailiopoulos, Ramchandran, and
  Bartlett]{pmlr-v84-yin18a}
Yin, D., Pananjady, A., Lam, M., Papailiopoulos, D., Ramchandran, K., and
  Bartlett, P.
\newblock Gradient diversity: a key ingredient for scalable distributed
  learning.
\newblock In \emph{Proceedings of the Twenty-First International Conference on
  Artificial Intelligence and Statistics}, volume~84, pp.\  1998--2007. PMLR,
  2018.

\bibitem[Yin et~al.(2021)Yin, Mallya, Vahdat, Alvarez, Kautz, and
  Molchanov]{Yin_2021_CVPR}
Yin, H., Mallya, A., Vahdat, A., Alvarez, J.~M., Kautz, J., and Molchanov, P.
\newblock See through gradients: Image batch recovery via gradinversion.
\newblock In \emph{Proceedings of the IEEE/CVF Conference on Computer Vision
  and Pattern Recognition (CVPR)}, 2021.

\bibitem[Ying et~al.(2021{\natexlab{a}})Ying, Yuan, Chen, Hu, Pan, and
  Yin]{ying2021exponential}
Ying, B., Yuan, K., Chen, Y., Hu, H., Pan, P., and Yin, W.
\newblock Exponential graph is provably efficient for decentralized deep
  training.
\newblock In \emph{Advances in Neural Information Processing Systems},
  2021{\natexlab{a}}.

\bibitem[Ying et~al.(2021{\natexlab{b}})Ying, Yuan, Hu, Chen, and
  Yin]{ying2021bluefog}
Ying, B., Yuan, K., Hu, H., Chen, Y., and Yin, W.
\newblock Bluefog: Make decentralized algorithms practical for optimization and
  deep learning.
\newblock \emph{arXiv preprint arXiv:2111.04287}, 2021{\natexlab{b}}.

\bibitem[You et~al.(2017)You, Gitman, and Ginsburg]{you2017large}
You, Y., Gitman, I., and Ginsburg, B.
\newblock Large batch training of convolutional networks.
\newblock \emph{arXiv preprint arXiv:1708.03888}, 2017.

\bibitem[You et~al.(2018)You, Zhang, Hsieh, Demmel, and
  Keutzer]{10.1145/3225058.3225069}
You, Y., Zhang, Z., Hsieh, C.-J., Demmel, J., and Keutzer, K.
\newblock Imagenet training in minutes.
\newblock In \emph{Proceedings of the 47th International Conference on Parallel
  Processing}. Association for Computing Machinery, 2018.

\bibitem[You et~al.(2020)You, Li, Reddi, Hseu, Kumar, Bhojanapalli, Song,
  Demmel, Keutzer, and Hsieh]{You2020Large}
You, Y., Li, J., Reddi, S., Hseu, J., Kumar, S., Bhojanapalli, S., Song, X.,
  Demmel, J., Keutzer, K., and Hsieh, C.-J.
\newblock Large batch optimization for deep learning: Training bert in 76
  minutes.
\newblock In \emph{International Conference on Learning Representations}, 2020.

\bibitem[Yuan et~al.(2022)Yuan, He, Davis, Zhang, Dao, Chen, Liang, Re, and
  Zhang]{yuan2022decentralized}
Yuan, B., He, Y., Davis, J.~Q., Zhang, T., Dao, T., Chen, B., Liang, P., Re,
  C., and Zhang, C.
\newblock Decentralized training of foundation models in heterogeneous
  environments.
\newblock In Oh, A.~H., Agarwal, A., Belgrave, D., and Cho, K. (eds.),
  \emph{Advances in Neural Information Processing Systems}, 2022.

\bibitem[Yuan et~al.(2021)Yuan, Chen, Huang, Zhang, Pan, Xu, and
  Yin]{yuan2021decentlam}
Yuan, K., Chen, Y., Huang, X., Zhang, Y., Pan, P., Xu, Y., and Yin, W.
\newblock Decentlam: Decentralized momentum sgd for large-batch deep training.
\newblock In \emph{Proceedings of the IEEE/CVF International Conference on
  Computer Vision}, pp.\  3029--3039, 2021.

\bibitem[Zellner(1988)]{zellner1988optimal}
Zellner, A.
\newblock Optimal information processing and bayes's theorem.
\newblock \emph{The American Statistician}, 42\penalty0 (4):\penalty0 278--280,
  1988.

\bibitem[Zhang et~al.(2020)Zhang, Cui, Kayi, Liu, Finkler, Kingsbury, Saon,
  Mroueh, Buyuktosunoglu, Das, Kung, and Picheny]{Zhang2020ImprovingEI}
Zhang, W., Cui, X., Kayi, A., Liu, M., Finkler, U., Kingsbury, B., Saon, G.,
  Mroueh, Y., Buyuktosunoglu, A., Das, P., Kung, D.~S., and Picheny, M.
\newblock Improving efficiency in large-scale decentralized distributed
  training.
\newblock \emph{ICASSP 2020 - 2020 IEEE International Conference on Acoustics,
  Speech and Signal Processing (ICASSP)}, pp.\  3022--3026, 2020.

\bibitem[Zhang et~al.(2021)Zhang, Liu, Feng, Cui, Kingsbury, and
  Tu]{zhang2021loss}
Zhang, W., Liu, M., Feng, Y., Cui, X., Kingsbury, B., and Tu, Y.
\newblock Loss landscape dependent self-adjusting learning rates in
  decentralized stochastic gradient descent.
\newblock \emph{arXiv preprint arXiv:2112.01433}, 2021.

\bibitem[Zheng et~al.(2021)Zheng, Zhang, and Mao]{zheng2021regularizing}
Zheng, Y., Zhang, R., and Mao, Y.
\newblock Regularizing neural networks via adversarial model perturbation.
\newblock In \emph{Proceedings of the IEEE/CVF Conference on Computer Vision
  and Pattern Recognition}, pp.\  8156--8165, 2021.

\bibitem[Zhu et~al.(2019{\natexlab{a}})Zhu, Liu, and Han]{NEURIPS2019_60a6c400}
Zhu, L., Liu, Z., and Han, S.
\newblock Deep leakage from gradients.
\newblock In \emph{Advances in Neural Information Processing Systems},
  2019{\natexlab{a}}.

\bibitem[Zhu et~al.(2022)Zhu, He, Zhang, Niu, Song, and Tao]{zhu2022topology}
Zhu, T., He, F., Zhang, L., Niu, Z., Song, M., and Tao, D.
\newblock Topology-aware generalization of decentralized sgd.
\newblock In \emph{International Conference on Machine Learning}. PMLR, 2022.

\bibitem[Zhu et~al.(2019{\natexlab{b}})Zhu, Wu, Yu, Wu, and
  Ma]{pmlr-v97-zhu19e}
Zhu, Z., Wu, J., Yu, B., Wu, L., and Ma, J.
\newblock The anisotropic noise in stochastic gradient descent: Its behavior of
  escaping from sharp minima and regularization effects.
\newblock In \emph{International Conference on Machine Learning}. PMLR,
  2019{\natexlab{b}}.

\bibitem[Zhuang et~al.(2022)Zhuang, Gong, Yuan, Cui, Adam, Dvornek, sekhar
  tatikonda, s~Duncan, and Liu]{zhuang2022surrogate}
Zhuang, J., Gong, B., Yuan, L., Cui, Y., Adam, H., Dvornek, N.~C., sekhar
  tatikonda, s~Duncan, J., and Liu, T.
\newblock Surrogate gap minimization improves sharpness-aware training.
\newblock In \emph{International Conference on Learning Representations}, 2022.

\end{thebibliography}
